\documentclass[10pt,twocolumn,letterpaper]{article}
\pdfoutput=1 
\usepackage{iccv}
\usepackage{times}
\usepackage{epsfig}
\usepackage{graphicx}
\usepackage{amsmath}
\usepackage{amssymb}
\usepackage{caption}
\usepackage{subcaption}
\usepackage{paralist}
\usepackage{array}
\usepackage{placeins}
\usepackage{tablefootnote}
\usepackage{epstopdf}

\usepackage[pagebackref=true,breaklinks=true,letterpaper=true,colorlinks,bookmarks=false]{hyperref}

\iccvfinalcopy

\begin{document}

\title{Object detection via a multi-region \& semantic segmentation-aware  CNN model}

\author{Spyros Gidaris\\
Universite Paris Est, Ecole des Ponts ParisTech\\
{\tt\small gidariss@imagine.enpc.fr}
\and
Nikos Komodakis\\
Universite Paris Est, Ecole des Ponts ParisTech\\
{\tt\small nikos.komodakis@enpc.fr}\\
}

\maketitle

\begin{abstract}
We propose an object detection system that relies on a multi-region deep convolutional neural network (CNN) that  also encodes semantic segmentation-aware features. The resulting  CNN-based representation aims at capturing a diverse set of  discriminative appearance factors and exhibits localization sensitivity that is essential for accurate object localization. We exploit the above properties of our recognition module by integrating it on an iterative localization mechanism that alternates between scoring a box proposal and refining its location with a deep CNN regression model. Thanks to the efficient use of our modules, we detect objects with very high localization accuracy. 
On the detection challenges of PASCAL VOC2007 and PASCAL VOC2012 we achieve mAP of $78.2\%$ and $73.9\%$ 
correspondingly, surpassing any other published work by a significant margin.
\let\thefootnote\relax\footnotetext{This work was supported by the ANR SEMAPOLIS project. Its code will become available on \url{<https://github.com/gidariss/mrcnn-object-detection>}.}
\end{abstract}

\section{Introduction}
One of the most studied problems of computer vision is that of object detection: given an image return all the instances of one or more type of objects in form of bounding boxes that tightly enclose them. 
The last two years, huge improvements have been observed on this task thanks to the recent advances of deep learning community~\cite{lecun1989backpropagation, bengio2007greedy, hinton2006reducing}.
Among them, most notable is the work of Sermanet et al.~\cite{sermanet2013overfeat} with the Overfeat framework and the work of Girshick et al.~\cite{girshick2014rich} with the R-CNN framework.

Overfeat~\cite{sermanet2013overfeat} uses two CNN models that applies on a sliding window fashion on multiple scales of an image.
The first is used to classify if a window contains an object and the second to predict the true bounding box location of the object. 
Finally, the dense class and location predictions are merged with a greedy algorithm in order to produce the final set of object detections.

R-CNN~\cite{girshick2014rich} uses Alex Krizhevsky's Net~\cite{krizhevsky2012imagenet} to extract features from box proposals provided by selective search~\cite{van2011segmentation} and then classifies them with class specific linear SVMs.
They manage to train networks with millions of parameters  by first pre-training on the auxiliary task of image classification and then fine-tuning on a small set of images annotated for the detection task. 
This simple pipeline surpasses by a large margin the detection performance of all the previously published systems, 
such as deformable parts models~\cite{felzenszwalb2010object} or non-linear multi-kernel approaches~\cite{vedaldi2009multiple}. Their success comes from the fact that they replaced the hand-engineered features like HOG~\cite{dalal2005histograms} or SIFT~\cite{lowe2004distinctive} with the high level object representations produced from the last layer of a CNN model. 
By employing an even deeper CNN model, such as the 16-layers VGG-Net~\cite{simonyan2014very}, they boosted the performance another $7$ points.
 
\begin{figure}[t!]
\center
\renewcommand{\figurename}{Figure}
\renewcommand{\captionlabelfont}{\bf}
\renewcommand{\captionfont}{\small} 
\begin{center}
        \begin{subfigure}[b]{0.155\textwidth}
                \includegraphics[width=\textwidth]{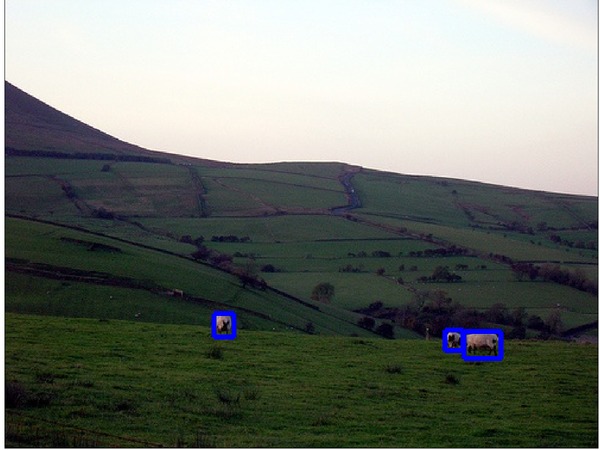}
        \end{subfigure}
        \begin{subfigure}[b]{0.155\textwidth}
                \includegraphics[width=\textwidth]{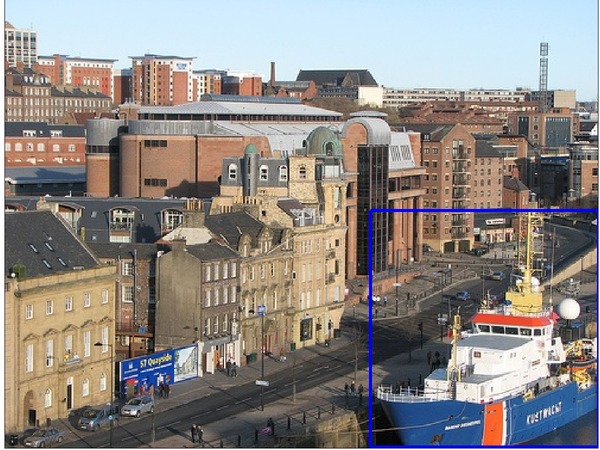}
        \end{subfigure}  
        \begin{subfigure}[b]{0.155\textwidth}
                \includegraphics[width=\textwidth]{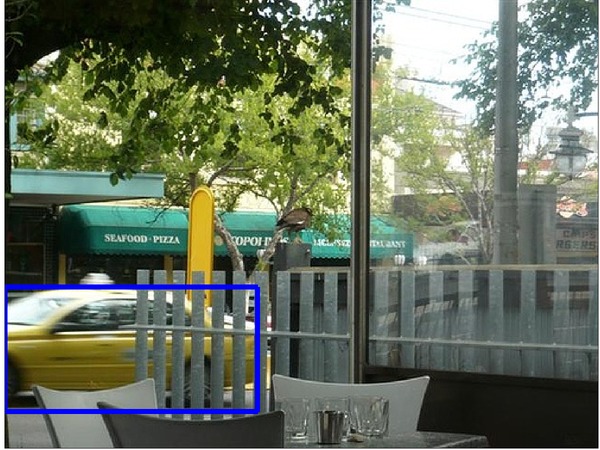}
        \end{subfigure}
\end{center}
\vspace{5pt}
   \caption{\textbf{Left:} detecting the sheep on this scene is very difficult without referring on the context, mountainish landscape. \textbf{Center:} In contrast, the context on the right image can only confuse the detection of the boat. The pure object characteristics is what a recognition model should focus on in this case. \textbf{Right:} This car instance is occluded on its right part and the recognition model should focus on the left part in order to confidently detect.}
\label{fig:Diffucult}
\vspace{-1.0em}
\end{figure}

In this paper we aim to further advance the state-of-the-art on object detection by improving on  two   key aspects that play a critical role in this task: object representation and object localization.

\textbf{\emph{Object representation.}} One of the lessons learned from the above-mentioned works  is that indeed features matter a lot on object detection and our work is partly motivated from this observation.
However, instead of  proposing  only a network architecture that is deeper, here we also opt  for an architecture of greater width, \ie, one whose last hidden layers provide features of increased dimensionality. 
In doing so, our goal is to build a richer candidate box representation. This is accomplished at two levels:

{\textbf{(1).}} At a first level, we want our object representation  to  capture several different aspects of an object such as its pure appearance characteristics,  the distinct appearance of its different regions (object parts), context appearance, the joint appearance on both sides of the object boundaries, and semantics. 
We believe that such a rich representation will further facilitate the problem of recognising (even difficult) object instances under a variety of circumstances (like, \eg, those depicted in figure \ref{fig:Diffucult}). 
In order to achieve our goal, we propose  a multi-component CNN model, called  \emph{multi-region CNN} hereafter,  each component of which is steered to focus on a different region of the object thus enforcing diversification of the discriminative appearance factors captured by it. 

Additionally, as we will explain shortly, by properly choosing and arranging some of these regions, we aim also to help our representation in being less invariant to inaccurate localization of an object. Note that this  property, which is highly desirable for detection, contradicts with  the built-in invariances of  CNN models, which stem from   the use of  max-pooling layers. 

\textbf{(2).} At a second level, inspired by the close connection that exists between segmentation and detection, we wish to enrich the above representation so that it also captures semantic segmentation information. To that end, we extend the above CNN model such that it also learns novel  CNN-based semantic segmentation-aware features. Importantly,   learning these features (\ie,  training  the extended unified CNN model) does not require having ground truth  object segmentations as training data. 

\textbf{\emph{Object localization.}} Besides object representation, our work is also motivated from the observation that, due to the tremendous classification capability of the recent CNN models~\cite{krizhevsky2012imagenet,zeiler2014visualizing, simonyan2014very,ioffe2015batch,he2015delving,szegedy2014going}, the bottleneck for good detection performance is now the accurate object localization. 
Indeed, it was noticed on R-CNN~\cite{girshick2014rich} that the most common type of false positives is the mis-localized detections. 
They fix some of them by employing a post processing step of bounding box regression that they apply on the final list of detections. 
However, their technique only helps on small localization errors.
We believe that there is much more space for improvement on this aspect. 
In order to prove it, we attempt to built a more powerful localization system that relies on combining our multi-region CNN model with a CNN-model for bounding box regression, which are used within an iterative scheme that alternates between scoring candidate boxes and refining their coordinates.

\textbf{Contributions.} To summarize, our contributions are as follows:

  \textbf{(1).} 
  We develop a multi-region CNN recognition model that yields an enriched object representation capable to  capture a diversity of discriminative appearance factors and to exhibit localization sensitivity that is desired for the task of accurate object localization.

  \textbf{(2).} We furthermore extend the above model by proposing a unified neural network architecture that also learns semantic segmentation-aware CNN features for the task of object detection. These features are jointly learnt in a weakly supervised manner, thus  requiring no additional  annotation.

  \textbf{(3).} We show how to significantly improve the localization capability by coupling the aforementioned  CNN recognition model with a CNN model for bounding box regression, adopting a scheme that alternates between scoring candidate boxes and refining their locations, as well as modifying the post-processing step of non-maximum-suppression. 

  \textbf{(4).} Our detection system achieves mAP of $78.2\%$ and $73.9\%$ on VOC2007~\cite{everingham2008pascal} and VOC2012~\cite{everingham2012pascal} detection challenges respectively, thus surpassing the previous state-of-art by a very significant margin.

The remainder of the paper is structured as follows:\ 
We discuss related work in \S\ref{sec:related_work}. 
We describe our multi-region CNN model in \S\ref{sec:multi_region_cnn}. 
We show how to extend it to also learn semantic segmentation-aware CNN features in \S\ref{sec:segmentation_features}. 
Our localization scheme is described in \S\ref{sec:object_localization} and implementation details are provided in \S\ref{sec:implementation_details}. 
We present experimental results in \S\ref{sec:experimental_results}, qualitative results in \S\ref{sec:qualitative_results} and conclude in \S\ref{sec:conclusions}.

\section{Related Work} \label{sec:related_work}
\begin{figure*}[t!]
\center
\renewcommand{\figurename}{Figure}
\renewcommand{\captionlabelfont}{\bf}
\renewcommand{\captionfont}{\small} 
\begin{center}
\includegraphics[width=0.9\textwidth]{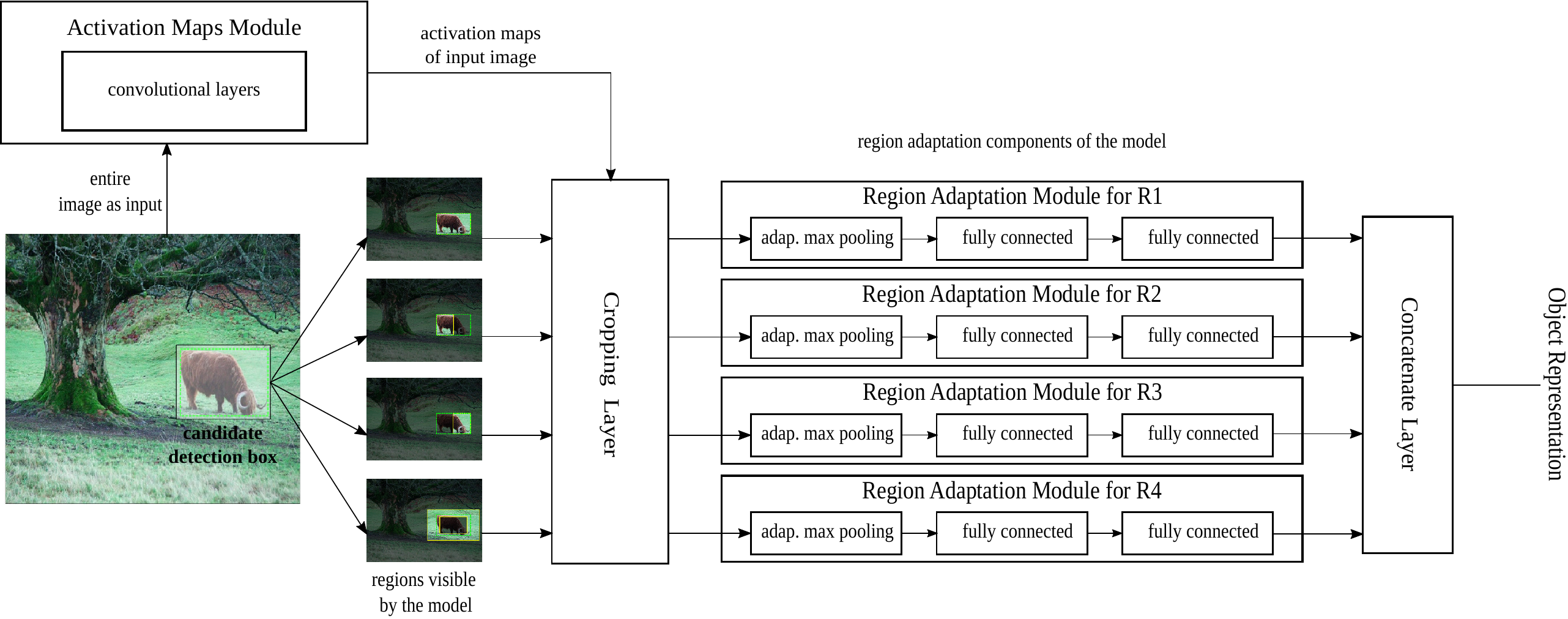}
\end{center}
\vspace{5pt}
   \caption{Multi Region CNN architecture. 
   For clarity we present only four of the regions that participate on it. An ``adaptive max pooling'' layer uses spatially adaptive pooling as in \cite{he2014spatial} (but with a one-level pyramid).
The above architecture can be extended to also learn  semantic segmentation-aware CNN features (see section \S\ref{sec:segmentation_features}) by including additional `activation-maps' and  `region-adaptation' modules  that are properly adapted for this task.}
\label{fig:Pipeline}
\vspace{-0.6em}
\end{figure*}

Apart from Overfeat~\cite{sermanet2013overfeat} and R-CNN~\cite{girshick2014rich}, several other recent papers are dealing with the object detection problem using deep neural networks. 
One is the very recent work of Zhu et al.~\cite{zhu2015segdeepm}, which shares some conceptual similarities with ours.
Specifically, they extract features from an additional region in order to capture the contextual appearance of a candidate box, they utilize a MRF inference framework to exploit object segmentation proposals (obtained through parametric min-cuts) in order to improve the object detection accuracy, and also use iterative box regression (based on ridge regression). 
More than them, we use multiple regions designed to diversify the appearance factors captured by our representation and to improve localization, we exploit CNN-based semantic segmentation-aware features (integrated in a unified neural network architecture), and make use of a deep CNN model for bounding box regression, as well as a box-voting scheme after non-max-suppression.
Feature extraction from multiple regions has also been exploited  for performing object recognition in videos by Leordeanu et al.~\cite{leordeanu2014features}.
As features they use the outputs of HOG~\cite{dalal2005histograms}+SVM classifiers trained on each region separately and the 1000-class predictions of a CNN pre-trained on ImageNet.
Instead, we fine-tune our deep networks on each region separately in order to  accomplish our goal of learning deep features that  will adequately capture their discriminative appearance characteristics. 
Furthermore, our regions exhibit more variety on their shape that, as we will see in section~\ref{sec:regions_role}, helps on boosting the detection performance.
On~\cite{szegedy2014scalable}, they designed a deep CNN model for object proposals generation and they use contextual features extracted from the last hidden layer of a CNN model trained on ImageNet classification task after they have applied it on large crops of the image. 
On~\cite{ouyang2014deepid}, they introduce a deep CNN with a novel deformation constrained pooling layer, a new strategy for pre-training that uses the bounding box annotations provided from ImageNet localization task, and contextual features derived by applying a pre-trained on ImageNet CNN on the whole image and treating the 1000-class probabilities for ImageNet objects as global contextual features. 
On SPP-Net~\cite{he2014spatial} detection framework, instead of applying their deep CNN on each candidate box separately as R-CNN does, they extract the convolutional feature maps from the whole image, project the candidate boxes on them, and then with an adaptive max-pooling layer, which consists of multiple pooling levels, they produce fixed length feature vectors that they pass through the fully connected layers of the CNN model. 
Thanks to those modifications, they manage to speed up computation by a considerable factor while maintaining high detection accuracy. Our work adopts this paradigm of processing.

Contemporary to our work are the approaches of~\cite{ren2015object,girshick2015fast,shaoqing2015faster} that are also based on the SPP-Net framework.
On~\cite{ren2015object}, they improve the SPP framework by replacing the sub-network component that is applied on the convolutional features extracted from the whole image with a deeper convolutional network. 
On~\cite{girshick2015fast}, they focus on simplifying the training phase of SPP-Net and R-CNN and speeding up both the testing and the training phases. 
Also, by fine-tuning the whole network and adopting a multi-task objective that has both box classification loss and box regression loss, they manage to improve the accuracy of their system.
Finally, on~\cite{shaoqing2015faster} they extend~\cite{girshick2015fast} by adding a new sub-network component for predicting class-independent proposals and thus making the system both faster and independent of object proposal algorithms.

\section{Multi-Region CNN Model} \label{sec:multi_region_cnn}
\begin{figure*}[t!]
\center
\renewcommand{\figurename}{Figure}
\renewcommand{\captionlabelfont}{\bf}
\renewcommand{\captionfont}{\small} 
        \begin{center}
        \begin{subfigure}[b]{0.19\textwidth}
        \center
                \renewcommand{\figurename}{Figure}
                \renewcommand{\captionlabelfont}{\bf}
                \renewcommand{\captionfont}{\footnotesize} 
                \includegraphics[width=\textwidth]{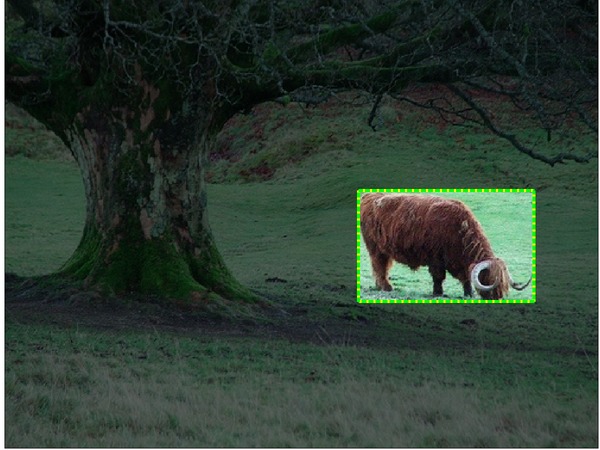}
                \caption{Original box}
                \label{fig:A}
        \end{subfigure}
        \hspace{0.03cm} 
        \begin{subfigure}[b]{0.19\textwidth}
        \center
                \renewcommand{\figurename}{Figure}
                \renewcommand{\captionlabelfont}{\bf}
                \renewcommand{\captionfont}{\footnotesize} 
                \includegraphics[width=\textwidth]{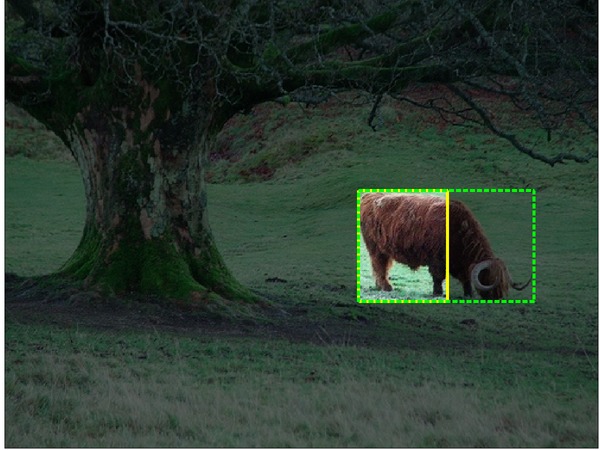}
                \caption{Half left}
                \label{fig:B}
        \end{subfigure}
        \hspace{0.03cm} 
        \begin{subfigure}[b]{0.19\textwidth}
        \center
                \renewcommand{\figurename}{Figure}
                \renewcommand{\captionlabelfont}{\bf}
                \renewcommand{\captionfont}{\footnotesize} 
                \includegraphics[width=\textwidth]{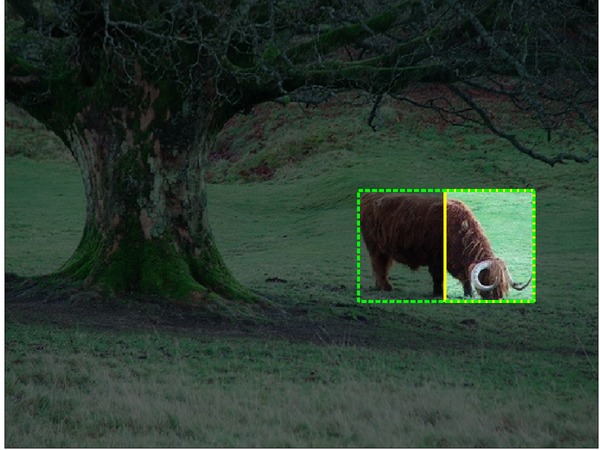}
                \caption{Half right}
                \label{fig:C}
        \end{subfigure}
        \hspace{0.03cm}
        \begin{subfigure}[b]{0.19\textwidth}
        \center
                \renewcommand{\figurename}{Figure}
                \renewcommand{\captionlabelfont}{\bf}
                \renewcommand{\captionfont}{\footnotesize} 
                \includegraphics[width=\textwidth]{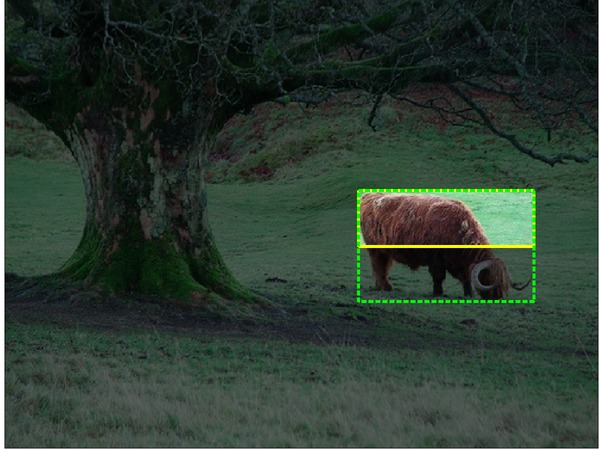}
                \caption{Half up}
                \label{fig:D}
        \end{subfigure}
        \hspace{0.03cm}
        \begin{subfigure}[b]{0.19\textwidth}
        \center
                \renewcommand{\figurename}{Figure}
                \renewcommand{\captionlabelfont}{\bf}
                \renewcommand{\captionfont}{\footnotesize} 
                \includegraphics[width=\textwidth]{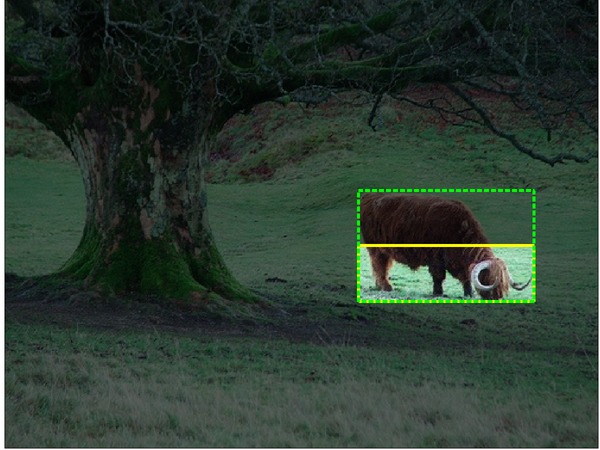}
                \caption{Half bottom}
                \label{fig:E}
        \end{subfigure}  
        
        \begin{subfigure}[b]{0.19\textwidth}
        \center
                \renewcommand{\figurename}{Figure}
                \renewcommand{\captionlabelfont}{\bf}
                \renewcommand{\captionfont}{\footnotesize} 
                \includegraphics[width=\textwidth]{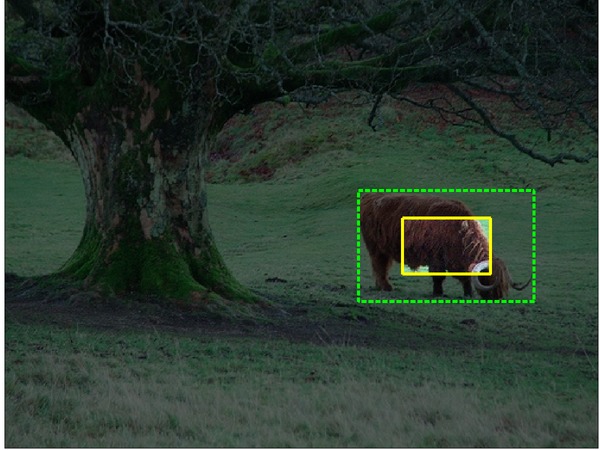}
                \caption{Central Region}
                \label{fig:F}
        \end{subfigure}
        \hspace{0.03cm}
        \begin{subfigure}[b]{0.19\textwidth}
        \center
                \renewcommand{\figurename}{Figure}
                \renewcommand{\captionlabelfont}{\bf}
                \renewcommand{\captionfont}{\footnotesize} 
                \includegraphics[width=\textwidth]{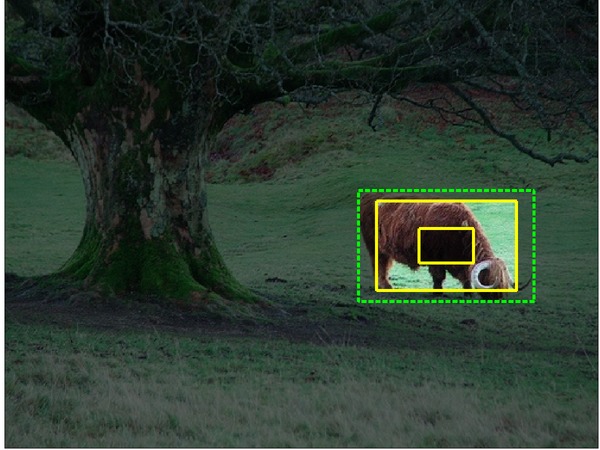}
                \caption{Central Region}
                \label{fig:G}
        \end{subfigure}
        \hspace{0.03cm} 
        \begin{subfigure}[b]{0.19\textwidth}
        \center
                \renewcommand{\figurename}{Figure}
                \renewcommand{\captionlabelfont}{\bf}
                \renewcommand{\captionfont}{\footnotesize} 
                \includegraphics[width=\textwidth]{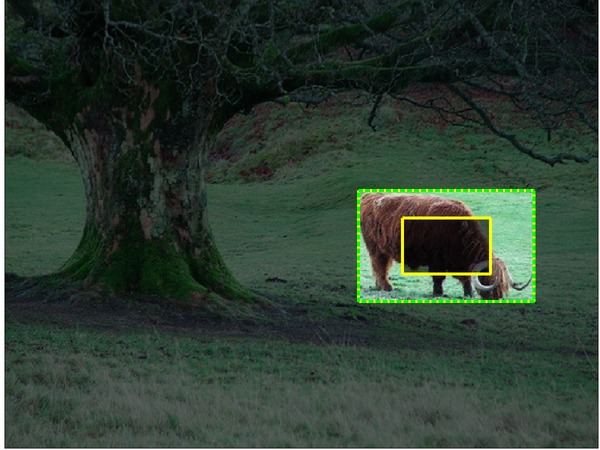}
                \caption{Border Region}
                \label{fig:H}
        \end{subfigure}
        \hspace{0.03cm}
        \begin{subfigure}[b]{0.19\textwidth}
        \center
                \renewcommand{\figurename}{Figure}
                \renewcommand{\captionlabelfont}{\bf}
                \renewcommand{\captionfont}{\footnotesize} 
                \includegraphics[width=\textwidth]{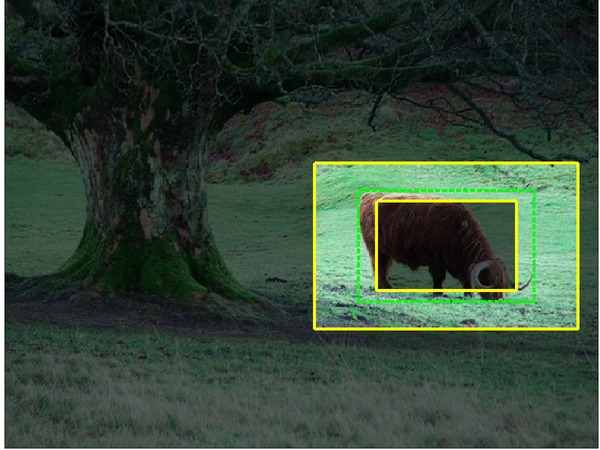}
                \caption{Border Region}
                \label{fig:I}
        \end{subfigure}
        \hspace{0.03cm}
        \begin{subfigure}[b]{0.19\textwidth}
        \center
                \renewcommand{\figurename}{Figure}
                \renewcommand{\captionlabelfont}{\bf}
                \renewcommand{\captionfont}{\footnotesize} 
                \includegraphics[width=\textwidth]{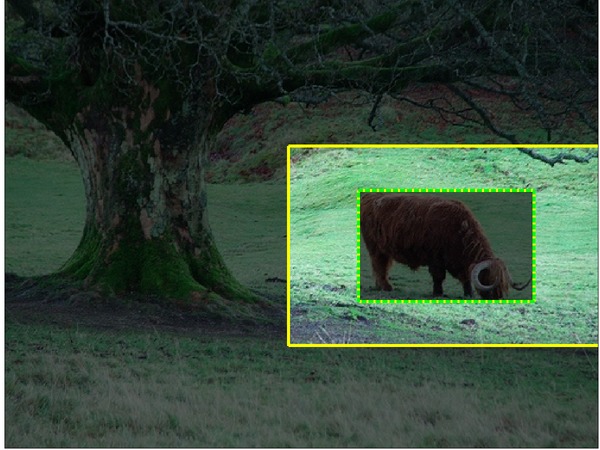}
                \caption{Context. Region}
                \label{fig:J}
        \end{subfigure}          
                      
        \end{center}
        \vspace{5pt}
        \caption{Illustration of the regions used on the Multi-Region CNN model. 
        With yellow solid lines are the borders of the regions and with green dashed lines are the borders of the candidate detection box. 
        \textbf{Region a:} it is the candidate box itself as being used on R-CNN~\cite{girshick2014rich}.
        \textbf{Region b, c, d, e:} they are the left/right/up/bottom half parts of the candidate box.
        \textbf{Region f:} it is obtained by scaling the candidate box by a factor of 0.5.
        \textbf{Region g:} the inner box is obtained by scaling the candidate box by a factor of 0.3 and the outer box by a factor of 0.8. 
        \textbf{Region h:} we obtain the inner box by scaling the candidate box by a factor of 0.5 and the outer box has the same size as the candidate box.
        \textbf{Region i:} the inner box is obtained by scaling the candidate box by a factor of 0.8 and the outer box by a factor of 1.5.  
        \textbf{Region j:} the inner box is the candidate box itself and the outer box is obtained by scaling the candidate box by a factor of 1.8.   
        }
        \label{fig:BoxAreas}
        \vspace{-0.6em}
\end{figure*}

The recognition model that we propose consists of a multi-component CNN network, each component of which is chosen so as to focus on a different region of an object. 
We call this a Multi-Region CNN model. We begin by  describing first its overall architecture. To that end, in order to facilitate the description of our model we introduce a general CNN architecture abstraction that decomposes the computation into two different modules:
\begin{description}
\item[\emph{\textbf{Activation maps module.}}] This part of the network gets as input the entire image and outputs activation maps (feature maps) by forwarding it through a sequence of convolutional layers.  

\item[\emph{\textbf{Region adaptation module.}}] Given a region $R$ on the image and the activation maps of the image, this module projects $R$ on the activation maps, crops the activations that lay inside it, pools them with a spatially adaptive (max-)pooling layer~\cite{he2014spatial}, and then forwards them through a multi-layer network.
\end{description}

Under this formalism, the architecture of the Multi-Region CNN model can be seen in figure~\ref{fig:Pipeline}.
Initially, the entire image is forwarded through the activation maps module. 
Then, a candidate detection box $B$ is analysed on a set of (possibly overlapping) regions $\{R_i\}_{i=1}^k$ each of which is assigned to a dedicated region adaptation module (note that these regions are always defined relatively to the bounding box $B$). 
As  mentioned previously, each of these region adaptation modules passes the activations pooled from its assigned region through a multilayer network that produces a high level feature.
Finally, the candidate box representation is obtained by concatenating the last hidden layer outputs of all the region adaptation modules. 

By steering the focus  on different regions of an object, our aim  is: (i) to force the network to capture various complementary aspects of the object’s appearance (\eg, context, object parts, \etc), thus  leading to a much richer and more robust object representation, and (ii) to also make the resulting representation more sensitive to inaccurate localization (\eg, by focusing on the border regions of an object), which is also crucial for object detection.

In the next section we describe how we choose the regions $\{R_i\}_{i=1}^k$ to achieve the above goals, and also discuss their role on object detection. 

\subsection{Region components and their role on detection} \label{sec:regions_role}
\begin{figure*}
\center
\renewcommand{\figurename}{Figure}
\renewcommand{\captionlabelfont}{\bf}
\renewcommand{\captionfont}{\small} 
\begin{center}
\includegraphics[width=0.9\textwidth]{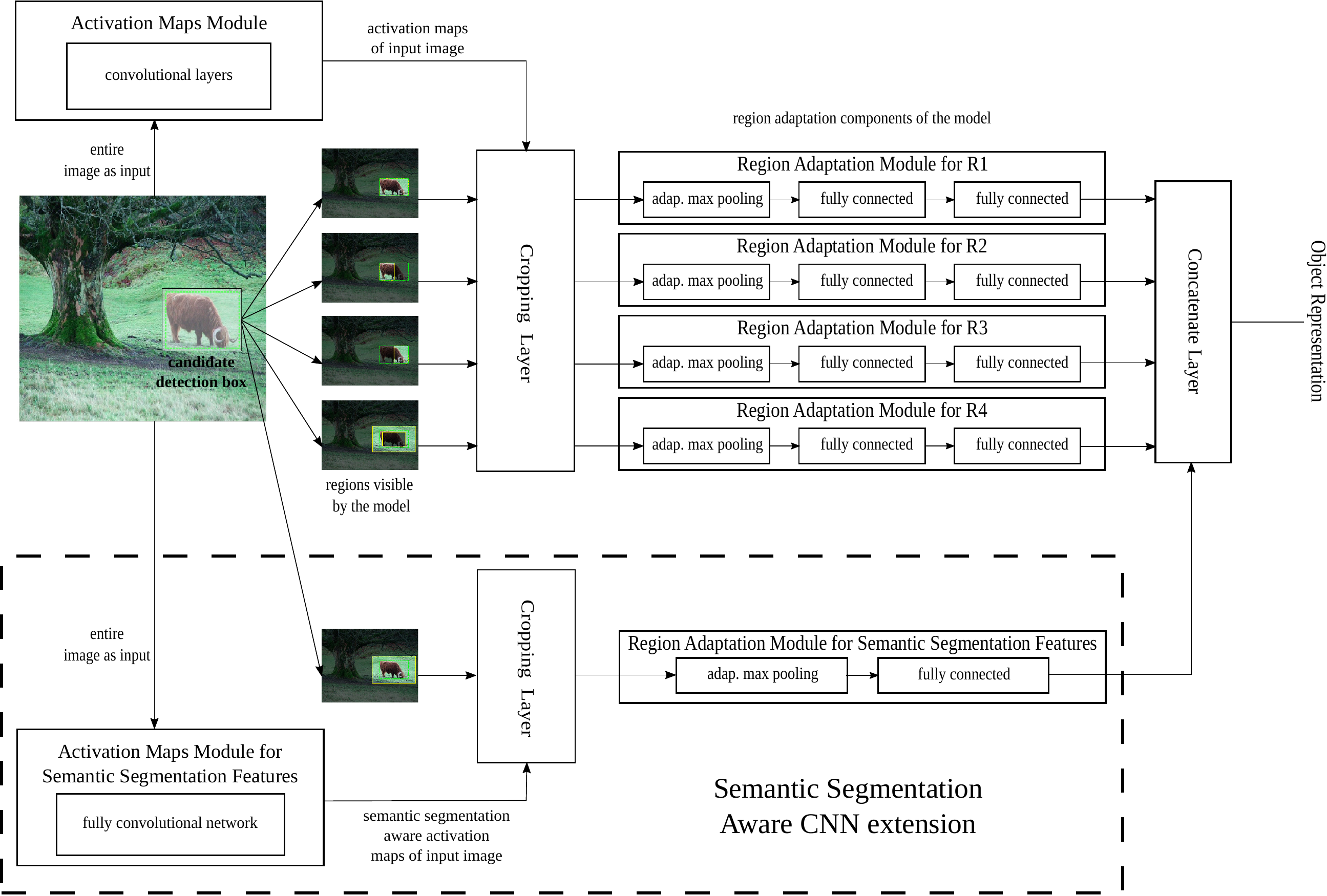}
\end{center}
\vspace{5pt}
   \caption{Multi Region CNN architecture extended with the semantic segmentation-aware CNN features.}
\label{fig:MRCNN_ext_SCNN}
\vspace{-0.6em}
\end{figure*}

We  utilize 2 types of region shapes: rectangles and rectangular rings, where the latter type is defined in terms of an inner and outer rectangle.
We describe below all of the  regions that we employ, while their specifications are given in the caption of figure~\ref{fig:BoxAreas}.
  
\emph{Original candidate box}: this is the candidate detection box itself as being used on R-CNN~\cite{girshick2014rich} (figure~\ref{fig:A}). A network trained on this type of region is guided to capture the appearance information of the entire object. 
When it is used alone consists the baseline of our work.

\emph{Half boxes}: those are the left/right/up/bottom half parts of a candidate box (figures~\ref{fig:B}, \ref{fig:C}, \ref{fig:D}, and \ref{fig:E}). Networks trained on each of them, are guided to learn the appearance characteristics present only on each half part of an object or on each side of the objects borders, aiming also to make  the representation more robust with respect to occlusions. 

\emph{Central Regions}: there are two type of central regions in our model (figures~\ref{fig:F} and~\ref{fig:G}). 
The networks trained on them are guided to capture the pure appearance characteristics of the central part of an object that is probably less interfered from other objects next to it or its background. 

\emph{Border Regions}: 
we include two such regions, with the shape of rectangular rings (figures~\ref{fig:H} and~\ref{fig:I}).
We expect that the dedicated on them networks will be guided to focus on the joint appearance characteristics on both sides of the object borders, also aiming to make the representation more sensitive to inaccurate localization.  

\emph{Contextual Region}: there is one region of this type that has rectangular ring shape (figure~\ref{fig:J}).
Its assigned network is driven to focus on the contextual appearance that surrounds an object such as the appearance of its background or of other objects next to it. 

\textbf{Role on  detection.} Concerning the general role of the  regions on object detection, we briefly focus below on two of the reasons
why using these regions helps:

\emph{Discriminative feature diversification.} Our hypothesis is that having regions that render visible to their network-components only a limited part of the object or only its immediate surrounding forces each network-component to discriminate image boxes solely based on the visual information that is apparent on them thus diversifying the discriminative factors captured by our overall recognition model. 
For example, if the border region depicted on figure \ref{fig:I} is replaced with one that includes its whole inner content, then we would expect that the network-component dedicated on it will not pay the desired attention on the visual content that is concentrated around the borders of an object. We tested such a hypothesis by conducting an experiment where we trained and tested two Multi-Region CNN models that consist of two regions each. 
Model A included the original box region (figure \ref{fig:A}) and the border region of figure  \ref{fig:I} that does not contain the central part of the object. 
On model B, we replaced the latter region (figure  \ref{fig:I}), which is a rectangular ring, with a normal box of the same size.
Both of them were trained on PASCAL VOC2007~\cite{everingham2008pascal} trainval set and tested on the test set of the same challenge. Model A achieved $64.1\%$ mAP while Model B achieved $62.9\%$ mAP which is $1.2$ points lower and validates our assumption.

\emph{Localization-aware representation.} 
We argue that our multi-region architecture as well as the type of regions included, address to a certain extent one of the major problems on the detection task, which is the inaccurate object localization. 
We believe that having multiple regions with network-components dedicated on each of them imposes soft constraints regarding the visual content allowed on each type of region for a given candidate detection box.
We experimentally justify this argument by referring to sections \ref{sec:det_error_an} and \ref{sec:loc_awareness}. 

\section{Semantic Segmentation-Aware CNN Model} \label{sec:segmentation_features}

To further  diversify the features encoded by our representation, we extend the Multi-Region CNN model so that it  also learns semantic segmentation-aware CNN features.
The motivation for this comes from the close connection  between segmentation and detection as well as from the fact that segmentation related cues are empirically known to often help  object detection \cite{dong2014towards,hariharan2014simultaneous,mottaghi2014role}.
In the context of our multi-region CNN network, the incorporation of the semantic segmentation-aware features is done by  adding properly adapted versions of the two main modules of the network, \ie, the `{activation-maps}' and  `{region-adaptation}' modules (see architecture in figure~\ref{fig:MRCNN_ext_SCNN}). We hereafter refer to the resulting modules as:
\begin{itemize}
\item \emph{Activation maps module for semantic segmentation-aware features.}
\item \emph{Region adaptation module for semantic segmentation-aware features.}
\end{itemize}
It is important to note that the modules for the semantic segmentation-aware features are trained \emph{without the use of any additional annotation}. Instead, they are trained in a \emph{weakly supervised manner} using only the provided bounding box annotations for detection.

We combine the Multi-Region CNN features and the semantic segmentation aware CNN features by concatenating them (see figure~\ref{fig:MRCNN_ext_SCNN}).
The resulting network thus jointly learns deep features of both types during training.

\subsection{Activation maps module for semantic segmentation-aware features}
\emph{\textbf{Fully Convolutional Nets.}} 
In order to serve the purpose of exploiting semantic segmentation aware features, for this module we adopt a Fully Convolutional Network~\cite{long2014fully}, abbreviated hereafter as FCN, trained to predict class specific foreground probabilities (we refer the interested reader to~\cite{long2014fully} for more details about FCN where it is being used for the task of semantic segmentation). 

\begin{figure}[t!]
\center
\renewcommand{\figurename}{Figure}
\renewcommand{\captionlabelfont}{\bf}
\renewcommand{\captionfont}{\small} 
        \begin{center}
        \begin{subfigure}[b]{0.15\textwidth}
                \includegraphics[width=\textwidth]{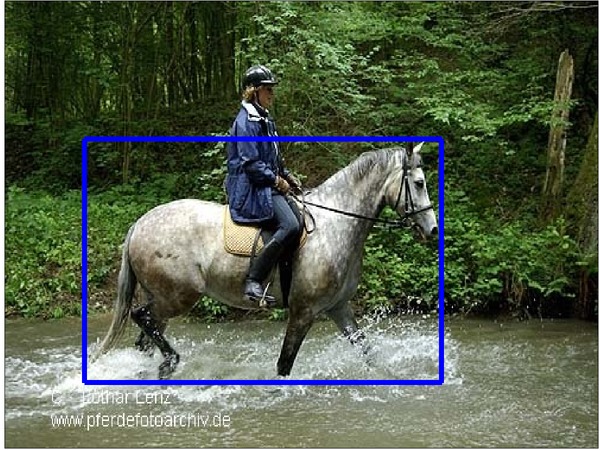}
        \end{subfigure}
        \hspace{0.05cm} 
        \begin{subfigure}[b]{0.15\textwidth}
                \includegraphics[width=\textwidth]{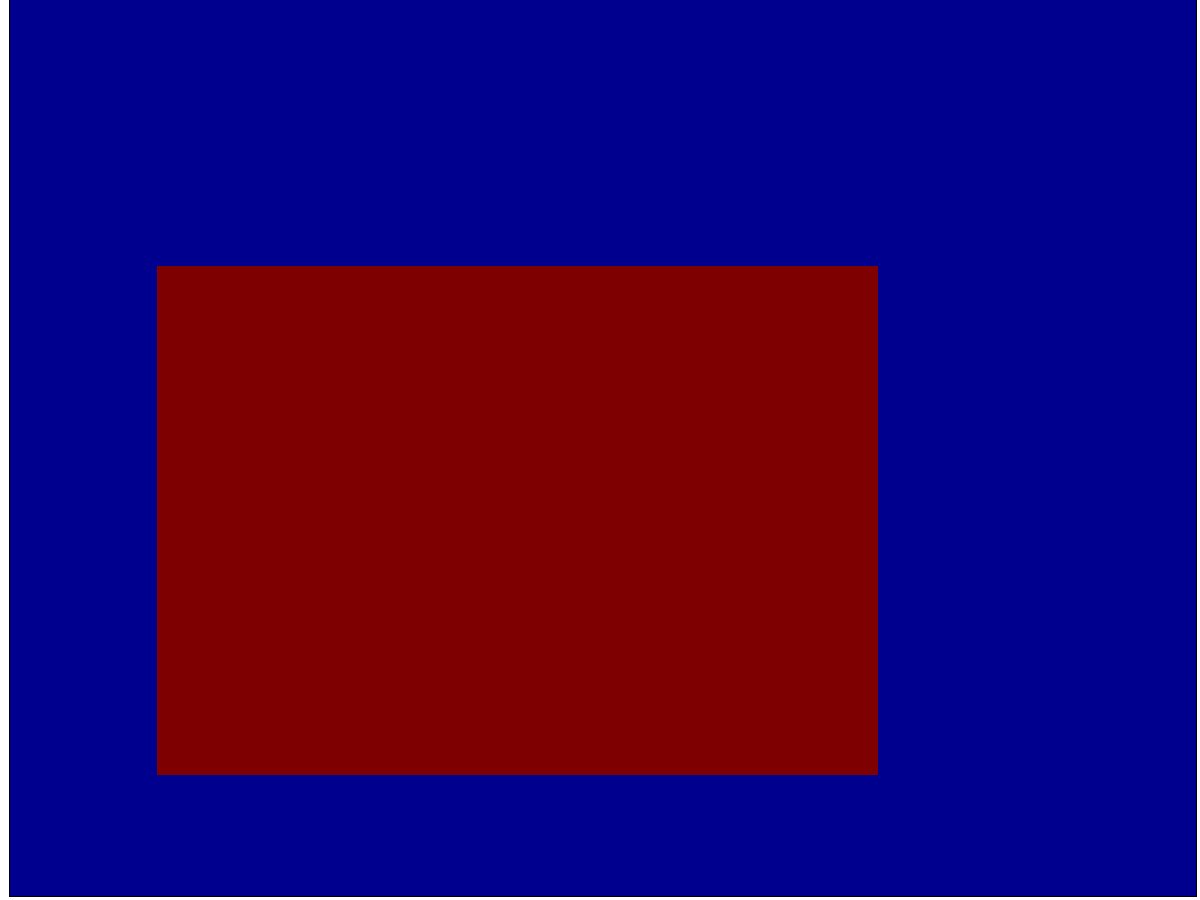}
        \end{subfigure}
        \hspace{0.05cm} 
        \begin{subfigure}[b]{0.15\textwidth}
                \includegraphics[width=\textwidth]{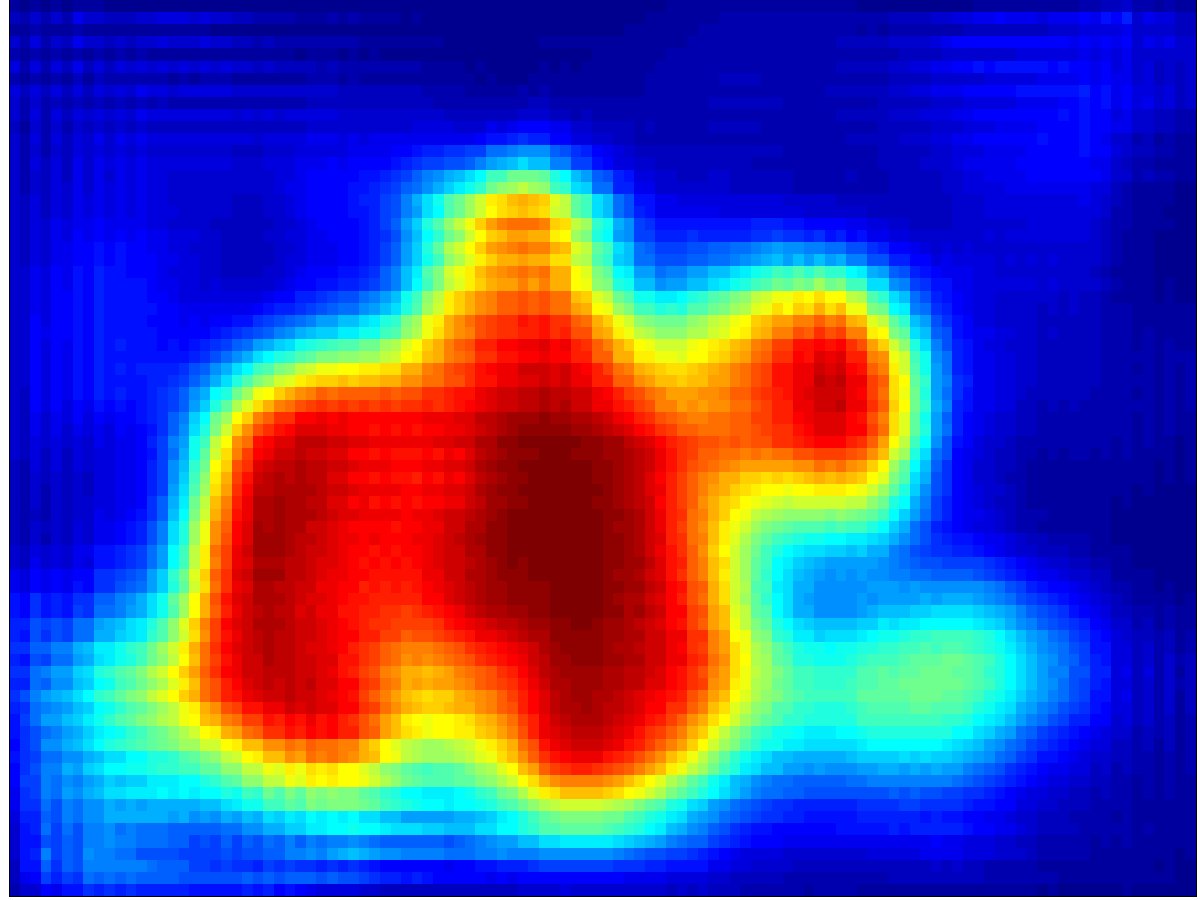}
        \end{subfigure}
        \vspace{-5pt}
        
        \begin{subfigure}[b]{0.15\textwidth}
                \includegraphics[width=\textwidth]{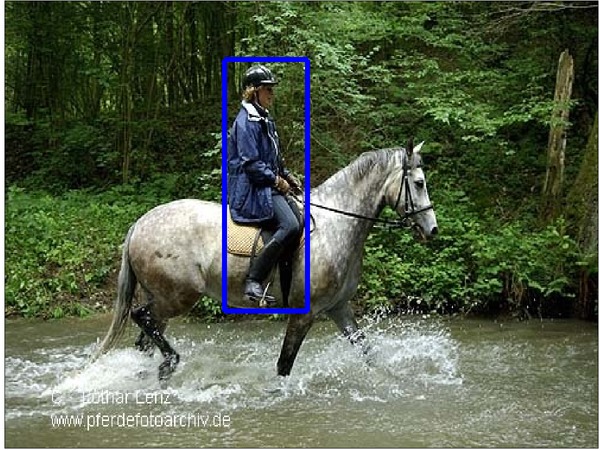}
        \end{subfigure}
        \hspace{0.05cm} 
        \begin{subfigure}[b]{0.15\textwidth}
                \includegraphics[width=\textwidth]{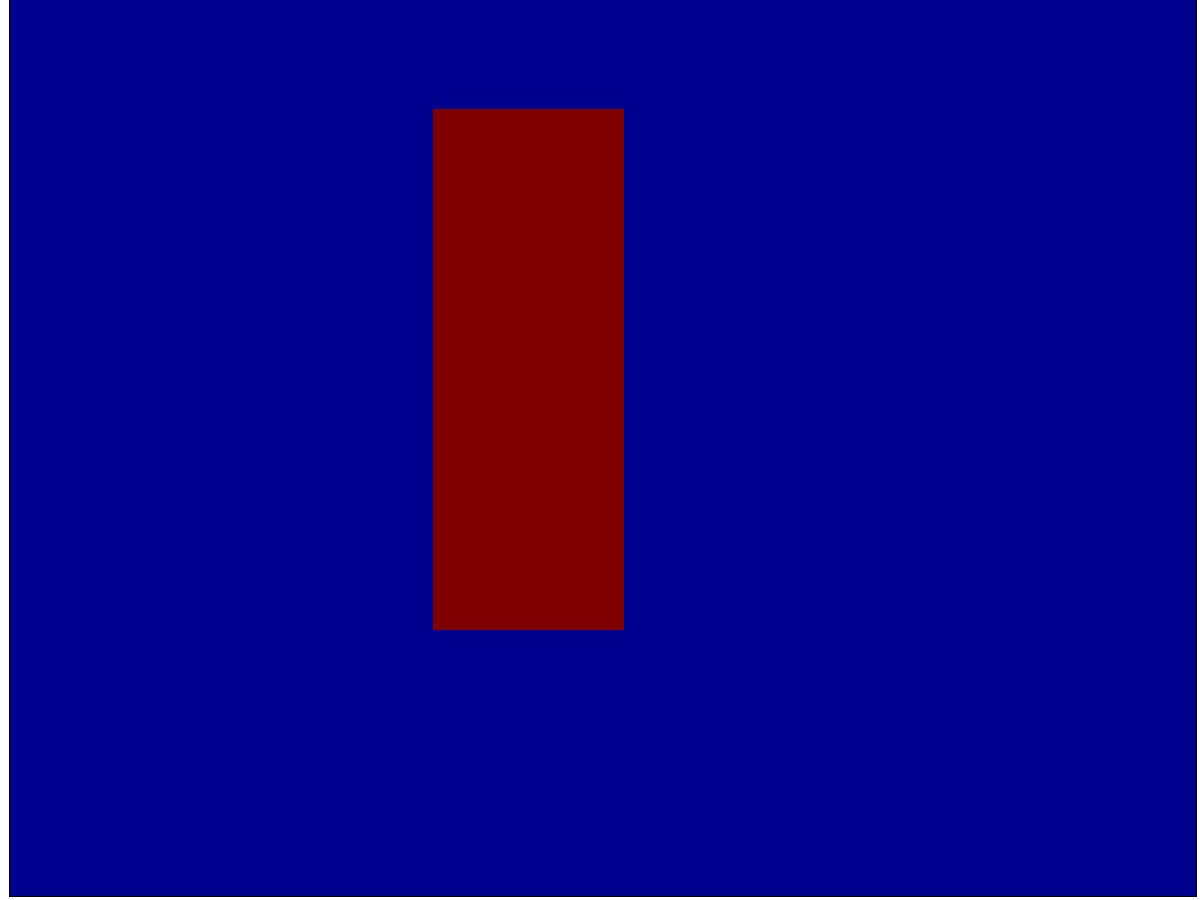}
        \end{subfigure}
        \hspace{0.05cm} 
        \begin{subfigure}[b]{0.15\textwidth}
                \includegraphics[width=\textwidth]{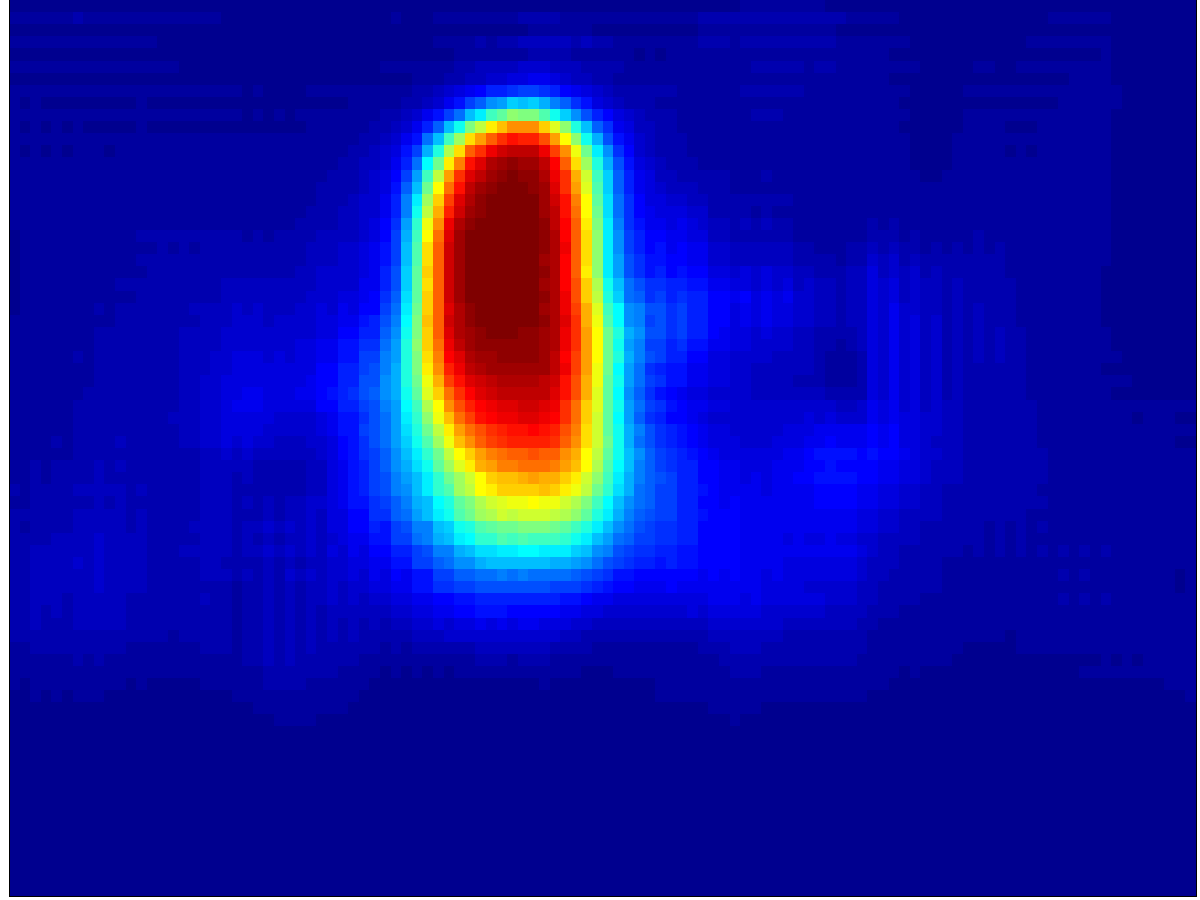}
        \end{subfigure}   
        \vspace{-5pt}
        
        \begin{subfigure}[b]{0.15\textwidth}
                \includegraphics[width=\textwidth]{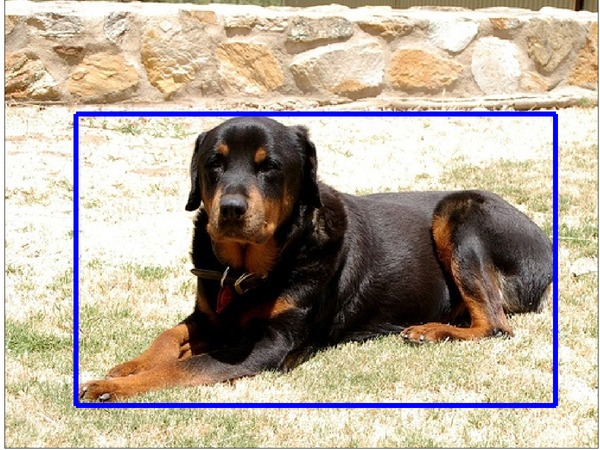}
        \end{subfigure}
        \hspace{0.05cm} 
        \begin{subfigure}[b]{0.15\textwidth}
                \includegraphics[width=\textwidth]{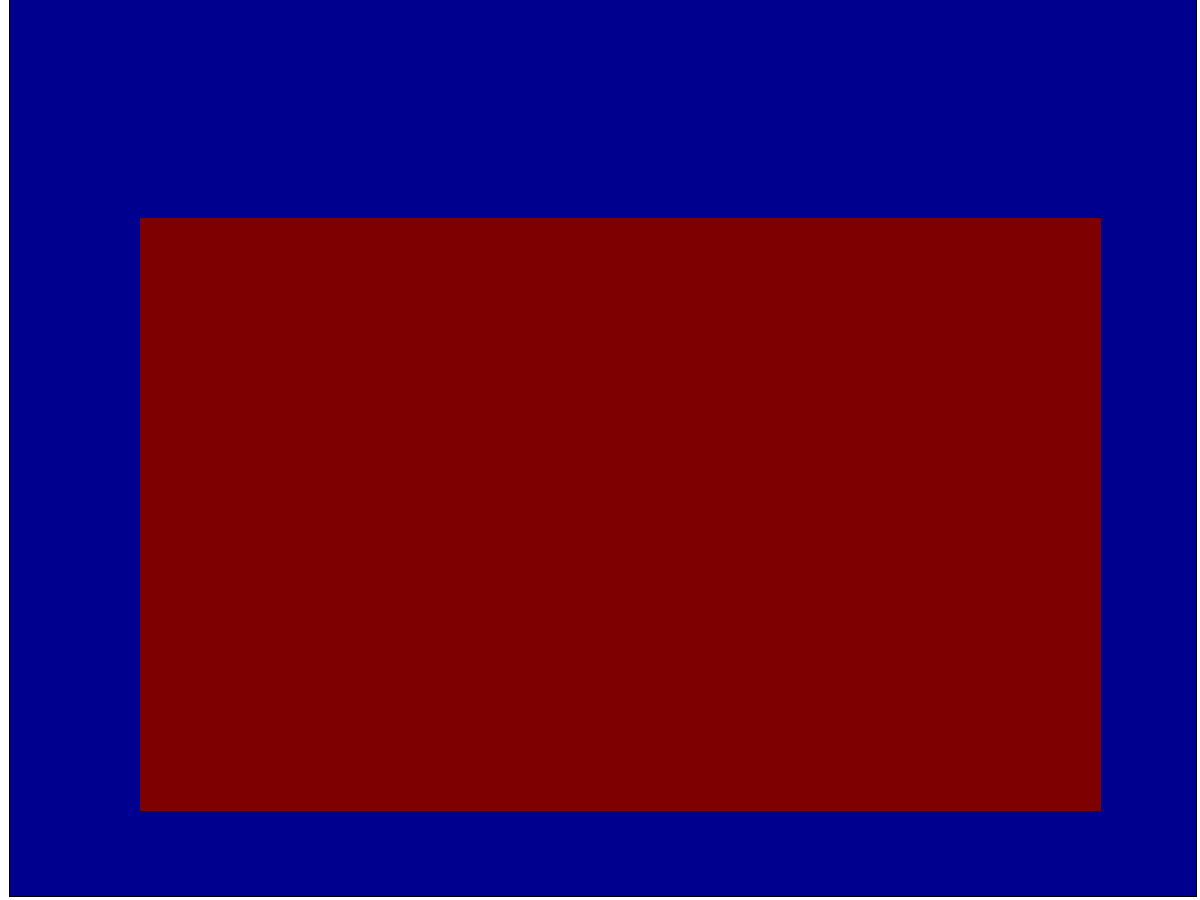}
        \end{subfigure}
        \hspace{0.05cm} 
        \begin{subfigure}[b]{0.15\textwidth}
                \includegraphics[width=\textwidth]{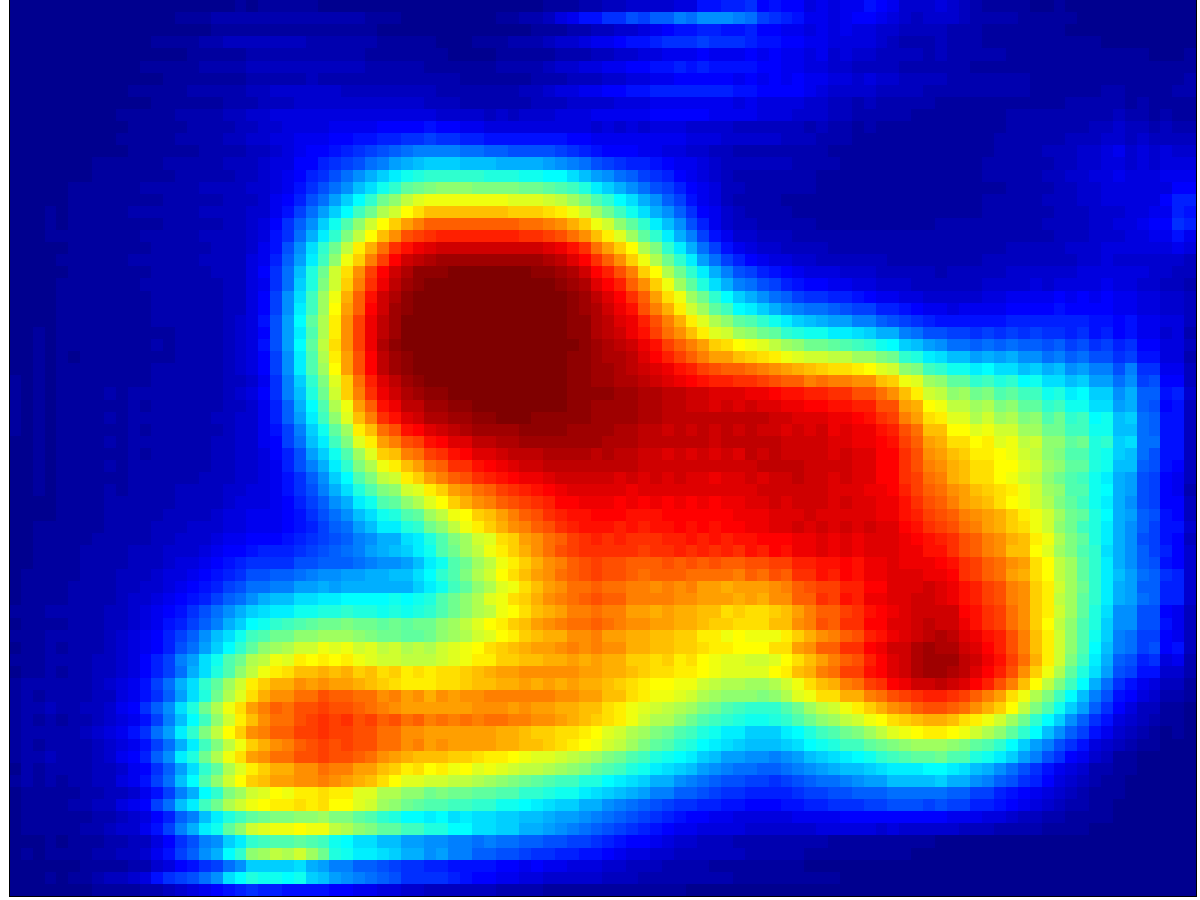}
        \end{subfigure}                   
        \end{center}
        \vspace{5pt}
        \caption{Illustration of the weakly supervised training of the FCN~\cite{long2014fully} used as activation maps module for the semantic segmentation aware CNN features.
        \textbf{Left column:} images with the ground truth bounding boxes drawn on them. 
        The classes depicted from top to down order are horse, human, and dog.
        \textbf{Middle column:} 
        the segmentation target values used during training of the FCN. They are artificially generated from the ground truth bounding box(es) on the left column. We use blue color for the background and red color for the foreground.      
        \textbf{Right column:} the foreground probabilities estimated from our trained FCN model. These clearly verify that, despite the weakly supervised training, our extracted features carry significant semantic segmentation information.}         
        \label{fig:SemanticSegmentationExamples}
        \vspace{-0.6em}
\end{figure}

\emph{\textbf{Weakly Supervised Training.}}
To train the activation maps module for the class-specific foreground segmentation task, 
we only use the annotations provided on object detection challenges (so as to make the training of our overall system independent of the availability of segmentation annotations). 
To that end, we follow a weakly supervised training strategy and we create artificial foreground class-specific segmentation masks using bounding box annotations. 
More specifically, the ground truth bounding boxes of an image are projected on the spatial domain of the last hidden layer of the FCN, and the "pixels" that lay inside the projected boxes are labelled as foreground while the rest are labelled as background (see left and middle column in figure~\ref{fig:SemanticSegmentationExamples}). 
The aforementioned process is performed independently for each class and yields as many segmentation target images as the number of our classes.
As can be seen in figure \ref{fig:SemanticSegmentationExamples} right column, despite  the weakly supervised way of training, the resulting activations still carry significant semantic segmentation information, enough even to delineate the boundaries of the object and separate the object from its background.

\emph{\textbf{Activation Maps.}} 
After the FCN has been trained on the auxiliary task of foreground segmentation, we drop the last classification layer and we use the rest of the FCN network in order to extract from images semantic segmentation aware activation maps.

\subsection{Region adaptation module for semantic segmentation-aware features}
We exploit the above activation maps by treating them as mid-level features and adding on top of them a single region adaptation module trained for our primary task of object detection.
In this case, we choose to use a single region obtained by enlarging the candidate detection box by a factor of $1.5$ (such a region contains semantic information also from the surrounding of a candidate detection box). 
The reason that we do not repeat the same regions as in the initial Multi-Region CNN architecture is for efficiency  as these are already used for capturing the appearance cues of an object.

\section{Object Localization} \label{sec:object_localization}
As already explained, our Multi-Region CNN  recognition model exhibits the localization awareness property that is necessary for accurate object localization. 
However, by itself it is not enough.
In order to make full use of it, our recognition model needs to be presented with well localized candidate boxes that in turn will be scored with high confidence from it. The solution that we adopt consists of 3 main components:

  \textbf{\emph{CNN region adaptation module for bounding box regression.}}
  We introduce an extra region adaptation module that, instead of being used for object recognition, is trained to predict the object bounding box. 
  It is applied on top of the activation maps produced from the Multi-Region CNN model 
  and, instead of a typical one-layer ridge regression model~\cite{girshick2014rich}, consists of two hidden fully connected layers and one prediction layer that outputs $4$ values (\ie, a bounding box) per category.
  In order to allow it to predict the location of object instances that are not in the close proximity of any of the initial candidate boxes, we use as region a box obtained by enlarging the candidate box by a factor of $1.3$. 
  This combination offers a significant boost on the detection performance of our system by allowing it to make more accurate predictions and for more distant objects.

  \textbf{\emph{Iterative Localization.}} 
  Our localization scheme starts from the selective search proposals~\cite{van2011segmentation} and works by iteratively scoring them and refining their coordinates.
  Specifically, let $\textbf{B}^{t}_c = \{B_{i,c}^{t}\}_{i=1}^{N_{c,t}}$ denote the set of $N_{c,t}$ bounding boxes generated on iteration $t$ for class $c$ and image $X$. 
  For each iteration\footnote{In practice $T\!=\!2$ iterations were enough for convergence.} $t=1,...,T$, the boxes from the previous iteration $\textbf{B}^{t-1}_c$ are scored with $s^{t}_{i,c} = \mathcal{F}_{rec}(B_{i,c}^{t-1}|c,X)$  by our recognition model and refined into $B_{i,c}^{t} = \mathcal{F}_{reg}(B_{i,c}^{t-1}|c,X)$ by our CNN regression model, thus forming the set of candidate detections $\textbf{D}^t_c = \{(s_{i,c}^t,B_{i,c}^t)\}_{i=1}^{N_{c,t}}$.
  For the first iteration $t=1$, the box proposals $\textbf{B}^0_c$ are coming from selective search~\cite{van2011segmentation} and are common between all the classes. 
  Also, those with score $s^{0}_{i,c}$  below a threshold $\tau_{s}$ are rejected\footnote{We use $\tau_{s} = -2.1$, which was selected such that the average number of box proposals per image from all the classes together to be around 250.} in order to reduce the computational burden of the subsequent iterations. 
  This way, we obtain a sequence of candidate detection sets $\{\textbf{D}^t_c\}_{t=1}^{T}$ that all-together both exhibit high recall of the objects on an image and are well localized on them.

 \textbf{\emph{Bounding box voting.}} After the last iteration $T$, 
 the candidate detections $\{\textbf{D}^t_c\}_{t=1}^{T}$ produced on each iteration $t$ are merged together $\textbf{D}_c=\cup_{t=1}^{T}{\textbf{D}_c^t}$.
 Because of the multiple regression steps, the generated boxes will be highly concentrated around the actual objects of interest. 
We exploit this "by-product" of the iterative localization scheme by adding a step of bounding box voting.  
First, standard non-max suppression~\cite{girshick2014rich} is applied on $\textbf{D}_c$ and produces the detections $\textbf{Y}_c = \{ (s_{i,c}, B_{i,c})\}$ using an IoU overlap threshold of 0.3.
Then, the final bounding box coordinates $B_{i,c}$ are further refined by having each box $B_{j,c}\in\mathcal{N}(B_{i,c})$ (where  $\mathcal{N}(B_{i,c})$ denotes the set of boxes in  $\textbf{D}_c$ that overlap with $B_{i,c}$ by more than $0.5$ on IoU metric) to vote for the bounding box location using as weight its score $w_{j,c} = max(0, s_{j,c})$, or
\begin{equation} \label{eq:boxVoting} 
B^{'}_{i,c} = \frac{\sum_{\scriptscriptstyle{j:B_{j,c} \in \mathcal{N}(B_{i,c})}} {w_{j,c} \cdot B_{j,c}}}{\sum_{\scriptscriptstyle{j:B_{j,c} \in \mathcal{N}(B_{i,c})}} w_{j,c}} .
\end{equation}
The final set of object detections for class $c$ will be $\textbf{Y}^{'}_c = \{ (s_{i,c}, B^{'}_{i,c})\}$.

\begin{figure}[t!]
\center
\renewcommand{\figurename}{Figure}
\renewcommand{\captionlabelfont}{\bf}
\renewcommand{\captionfont}{\small} 
        \begin{center}
        \begin{subfigure}[b]{0.22\textwidth}
        \center
                \renewcommand{\figurename}{Figure}
                \renewcommand{\captionlabelfont}{\bf}
                \renewcommand{\captionfont}{\footnotesize} 
                \includegraphics[width=\textwidth]{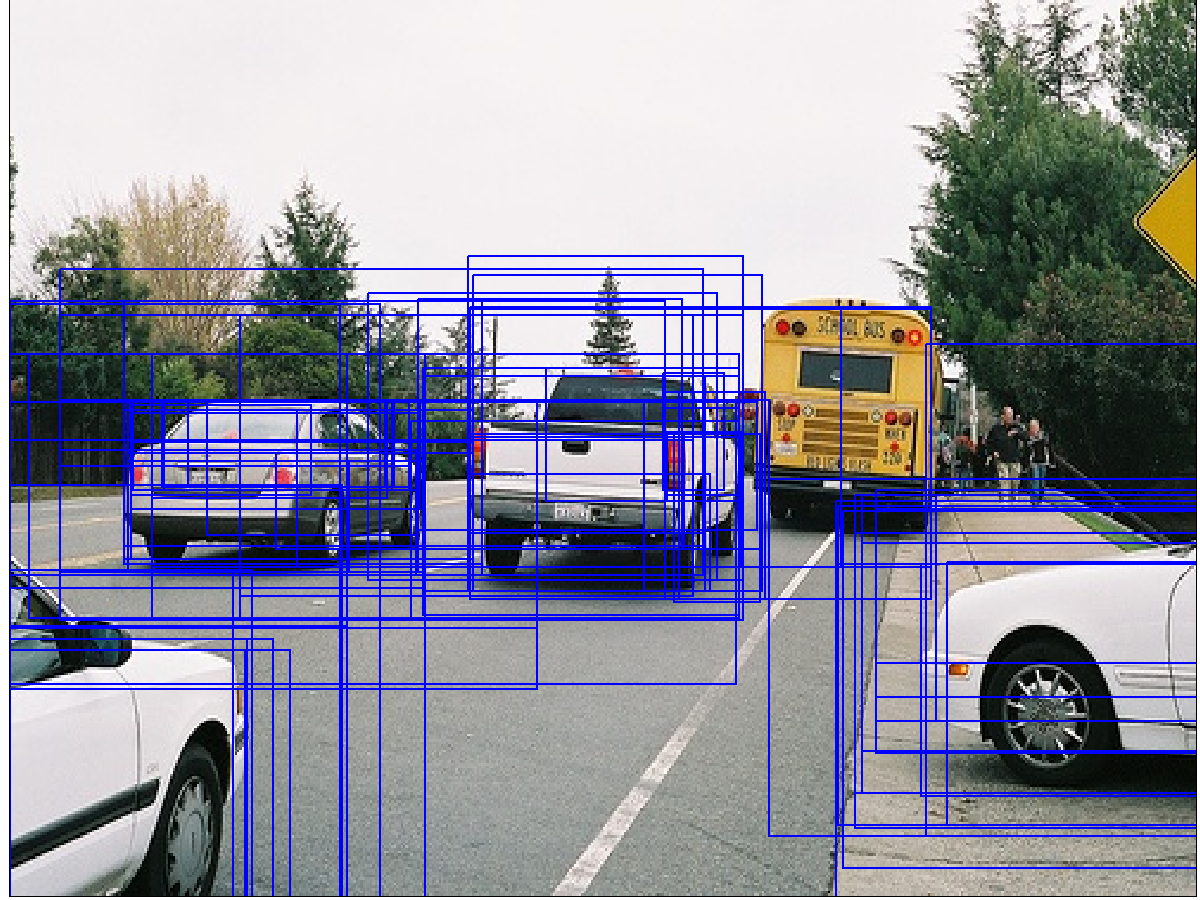}
                \caption{\footnotesize{Step 1}}
                \label{fig:Step1}
        \end{subfigure}
        \hspace{0.05cm} 
        \begin{subfigure}[b]{0.22\textwidth}
        \center
                \renewcommand{\figurename}{Figure}
                \renewcommand{\captionlabelfont}{\bf}
                \renewcommand{\captionfont}{\footnotesize} 
                \includegraphics[width=\textwidth]{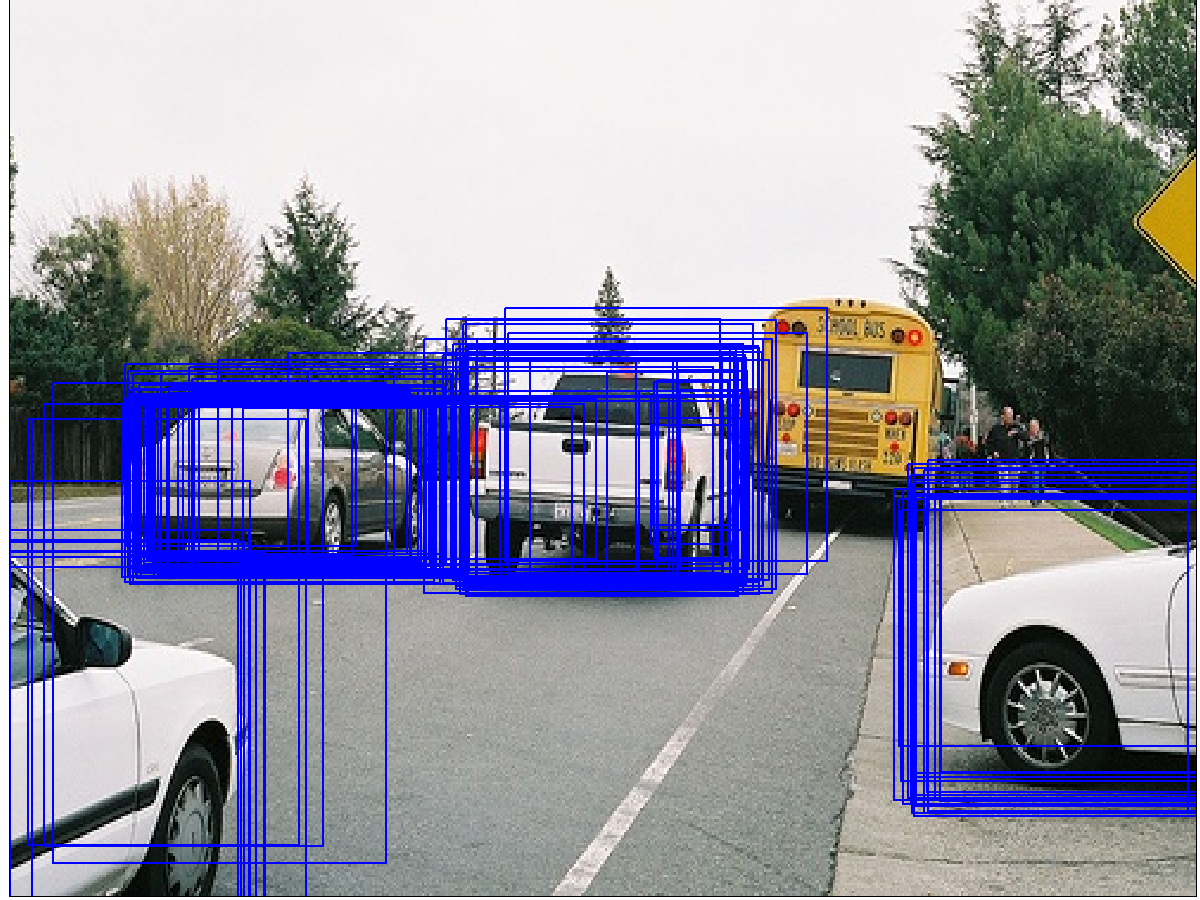}
                \caption{\footnotesize{Step 2}}
                \label{fig:Step2}
        \end{subfigure}
        \vspace{0.1cm}
        
        \begin{subfigure}[b]{0.22\textwidth}
        \center
                \renewcommand{\figurename}{Figure}
                \renewcommand{\captionlabelfont}{\bf}
                \renewcommand{\captionfont}{\footnotesize} 
                \includegraphics[width=\textwidth]{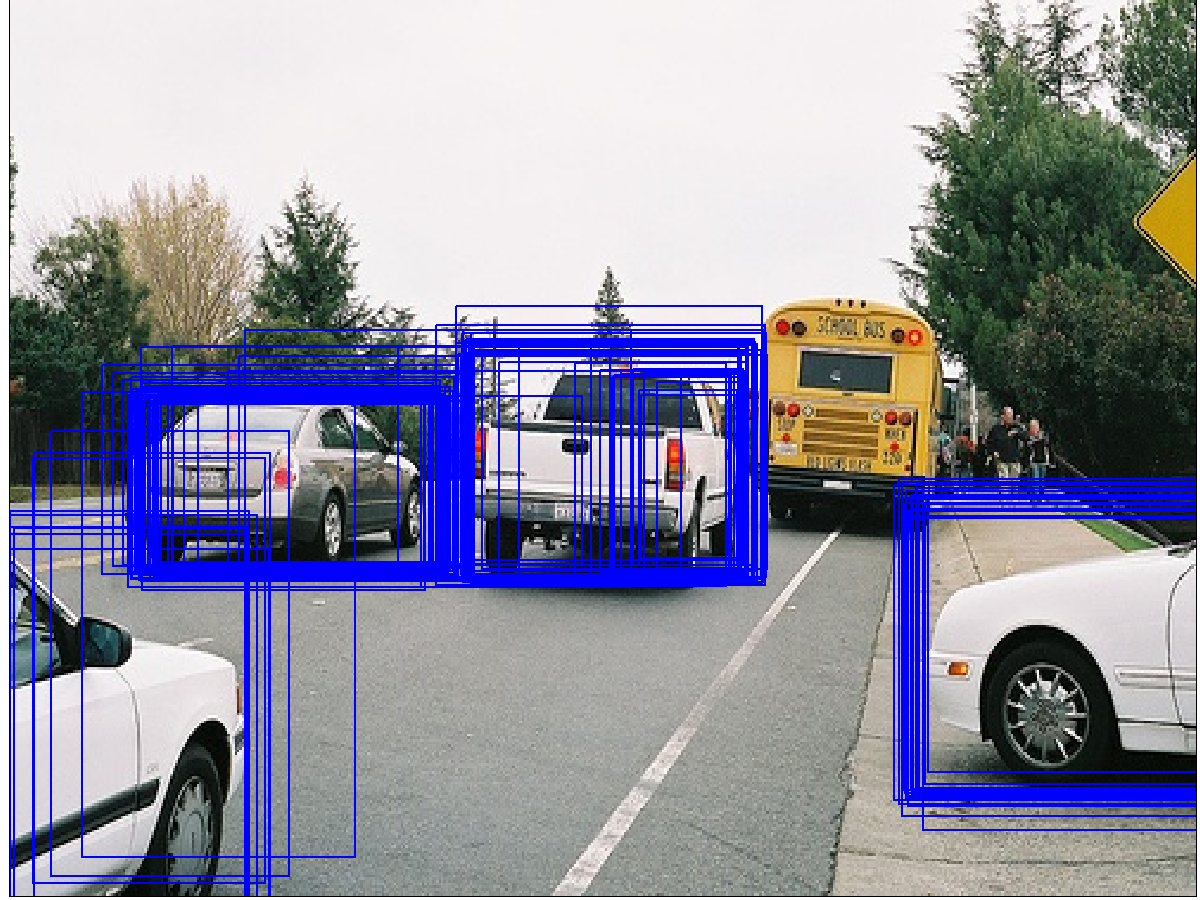}
                \caption{\footnotesize{Step 3}}
                \label{fig:Step3}
        \end{subfigure}
        \hspace{0.05cm} 
        \begin{subfigure}[b]{0.22\textwidth}
        \center
                \renewcommand{\figurename}{Figure}
                \renewcommand{\captionlabelfont}{\bf}
                \renewcommand{\captionfont}{\footnotesize} 
                \includegraphics[width=\textwidth]{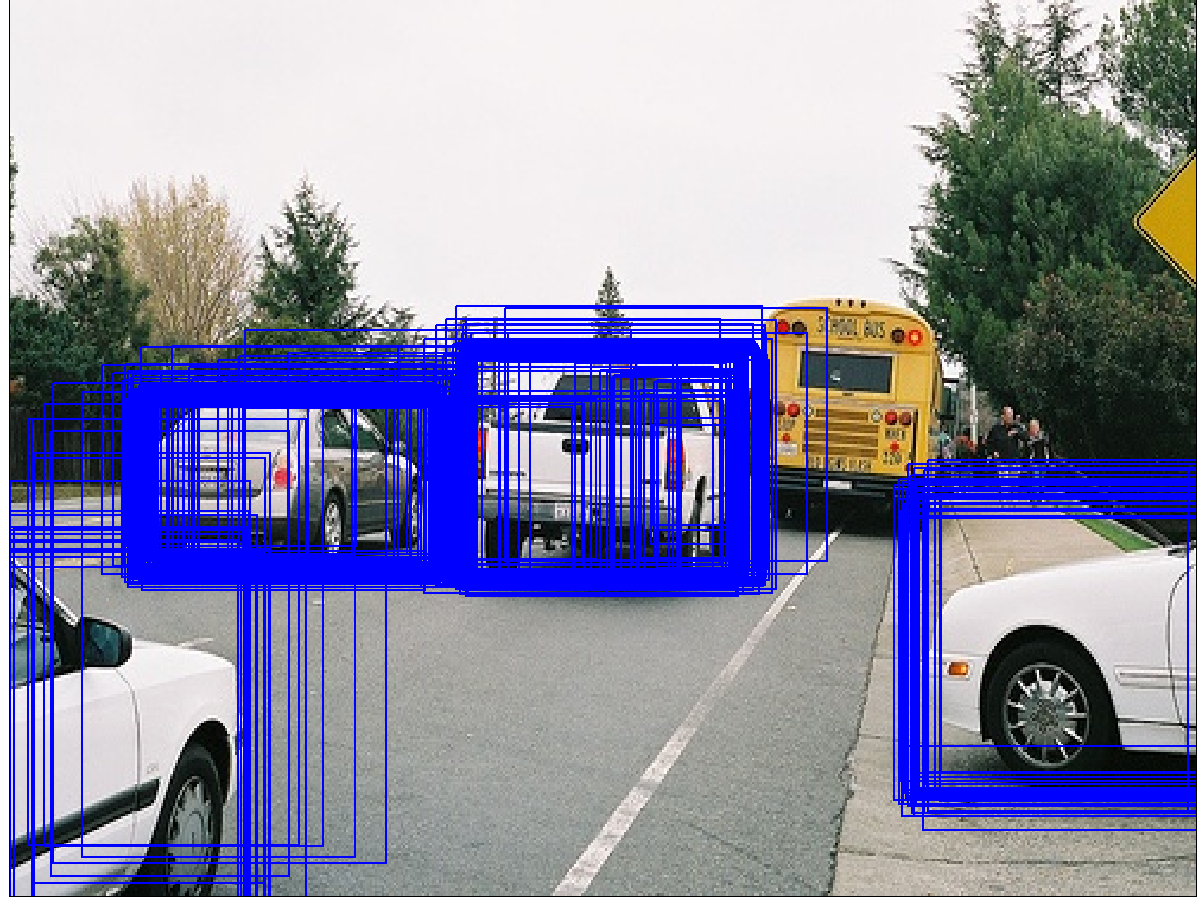}
                \caption{\footnotesize{Step 4}}
                \label{fig:Step4}
        \end{subfigure}
        \vspace{0.1cm}
        \begin{subfigure}[b]{0.44\textwidth}
        \center
                \renewcommand{\figurename}{Figure}
                \renewcommand{\captionlabelfont}{\bf}
                \renewcommand{\captionfont}{\footnotesize} 
                \includegraphics[width=\textwidth]{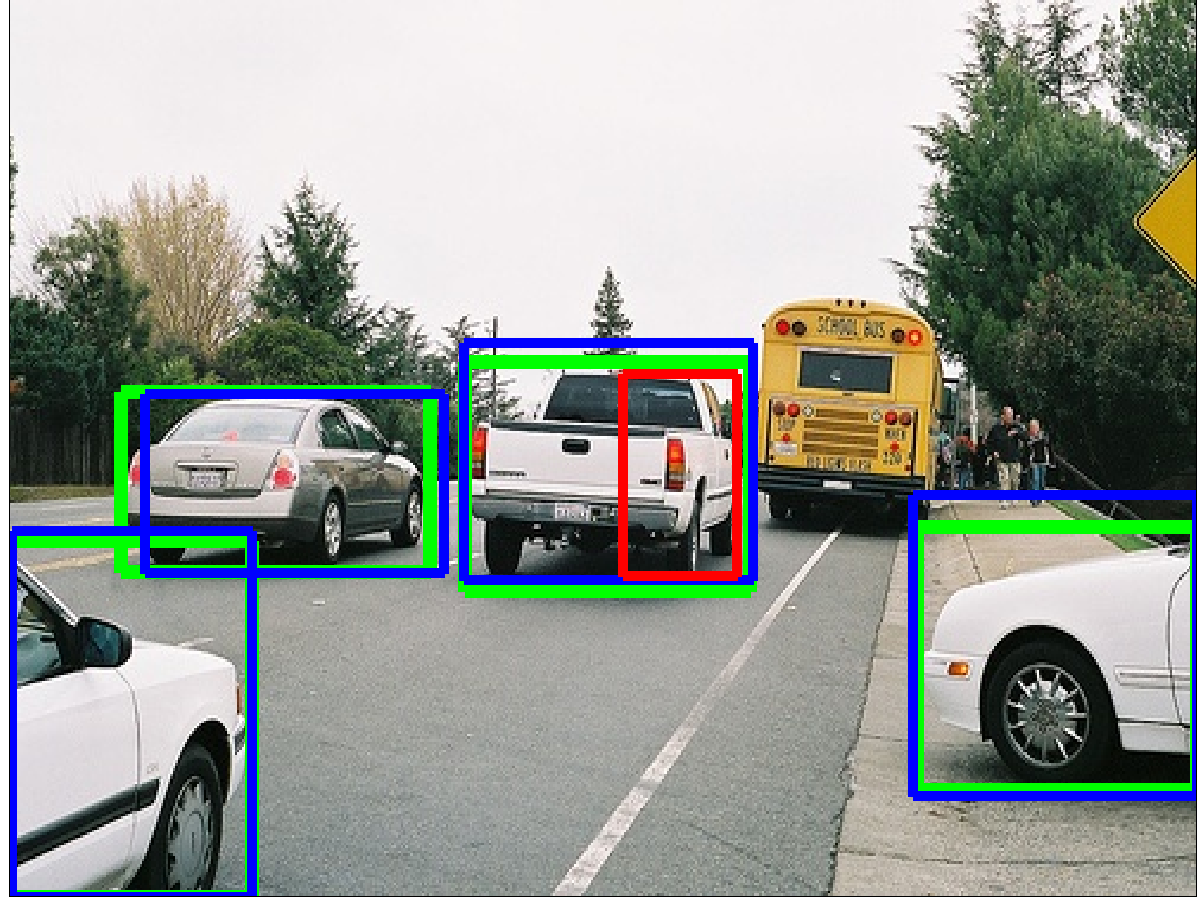}
                \caption{\footnotesize{Step 5}}
                \label{fig:Step5}
        \end{subfigure}            
        \end{center}
        \vspace{1pt}
        \caption{Illustration of the object localization scheme for instances of the class car. We describe the images from left to right and top to down order.
        \textbf{Step 1:} the initial box proposal of the image. For clarity we visualize only the box proposals that are not rejected after the first scoring step. 
        \textbf{Step 2:} the new box locations obtained after performing CNN based bounding box regression on the boxes of Step 1. 
        \textbf{Step 3:} the boxes obtained after a second step of box scoring and regressing on the boxes of Step 2. 
        \textbf{Step 4:} the boxes of Step 2 and Step 3 merged together.
        \textbf{Step 5:} the detected boxes after applying non-maximum-suppression and box voting on the boxes of Step 4.  On the final detections we use blue color for the true positives and red color for the false positives. Also, the ground truth bounding boxes are drawn with green color. 
        The false positive that we see after the last step is a duplicate detection that survived from non-maximum-suppression.}
        \label{fig:ObjectLocalization}
        \vspace{-0.6em}
\end{figure}

In figure~\ref{fig:ObjectLocalization} we provide a visual illustration of the object localization.

\section{Implementation Details} \label{sec:implementation_details}

For all the CNN models involved in our proposed system, we used the publicly available 16-layers VGG model~\cite{simonyan2014very} pre-trained on ImageNet~\cite{deng2009imagenet} for the task of image classification\footnote{https://gist.github.com/ksimonyan/}. 
For simplicity, we fine-tuned only the fully connected layers (fc6 and fc7) of each model while we preserved the pre-trained weights for the convolutional layers (conv1\_1 to conv5\_3), which are shared among all the models of our system. 

\textbf{Multi-Region CNN model.}
Its activation maps module consists of the convolutional part (layers conv1\_1 to conv5\_3) of the 16-layers VGG-Net that outputs $512$ feature channels. The max-pooling layer right after the last convolutional layer is omitted on this module.
Each region adaptation module inherits the fully connected layers of the 16-layers VGG-Net and is fine-tuned separately from the others. 
Regarding the regions that are rectangular rings, both the inner and outer box are projected on the activation maps and then the activations that lay inside the inner box are masked out by setting them to zero (similar to the Convolutional Feature Masking layer proposed on~\cite{dai2014convolutional}).
In order to train the region adaptation modules, we follow the guidelines of R-CNN~\cite{girshick2014rich}. 
As an optimization objective we use the softmax-loss and the minimization is performed with stochastic gradient descent (SGD). 
The momentum is set to $0.9$, the learning rate is initially set to $0.001$ and then reduced by a factor of $10$ every $30k$ iterations, and the minibatch has $128$ samples.
The positive samples are defined as the selective search proposals~\cite{van2011segmentation} that overlap a ground-truth bounding box by at least $0.5$. As negative samples we use the proposals that overlap with a ground-truth bounding box on the range $[0.1, 0.5)$. 
The labelling of the training samples is relative to the original candidate boxes and is the same across all the different regions.

\textbf{Activation maps module for semantic segmentation aware features.} 
Its architecture consists of the 16-layers VGG-Net without the last classification layer and transformed to a FCN~\cite{long2014fully} (by reshaping the fc6 and fc7 fully connected layers to convolutional ones with kernel size of $7 \times 7$ and $1 \times 1$ correspondingly). 
For efficiency purposes, we reduce the output channels of the fc7 layer from $4096$ to $512$.
In order to learn the semantic segmentation aware features, we use an auxiliary fc8 convolutional classification layer (of kernel size $1 \times 1$) that outputs as many channels as our classes and a binary (foreground vs background) logistic loss applied on each spatial cell and for each class independently. 
Initially, we train the FCN with the $4096$ channels on the fc7 layer until convergence.  
Then, we replace the fc7 layer with another one that has $512$ output channels, which is initialized from a Gaussian distribution, and the training of the FCN starts from the beginning and is continued until convergence again.
For loss minimization we use SGD with minibatch of size $10$. The momentum is set to $0.9$ and the learning rate is initialized to $0.01$ and decreased by a factor of $10$ every $20$ epochs. For faster convergence, the learning rate of the randomly initialized fc7 layer with the $512$ channels is multiplied by a factor of $10$.

\textbf{Region adaptation module for semantic segmentation aware features.}
Its architecture consists of a spatially adaptive max-pooling layer~\cite{long2014fully} that outputs feature maps of $512$ channels on a $9 \times 9$ grid, and a fully connected layer with $2096$ channels.
In order to train it, we use the same procedure as for the region components of the Multi-Region CNN model. 
During training, we only learn the weights of the region adaptation module layers that are randomly initialized from a Gaussian distribution.

\textbf{Classification SVMs.} In order to train the SVMs we follow the same principles as in \cite{girshick2014rich}. 
As positive samples are considered the ground truth bounding boxes and as negative samples are considered the selective search proposals~\cite{van2011segmentation} that overlap with the ground truth boxes by less than 0.3. 
We use hard negative mining the same way as in~\cite{girshick2014rich, felzenszwalb2010object}.

\textbf{CNN region adaptation module for bounding box regression.}
The activation maps module used as input in this case is common with the Multi-Region CNN model.
The region adaptation module for bounding box regression inherits the fully connected hidden layers of the 16-layers VGG-Net.
As a loss function we use the euclidean distance between the target values and the network predictions. 
For training samples we use the box proposals~\cite{van2011segmentation} that overlap by at least $0.4$ with the ground truth bounding boxes. 
The target values are defined the same way as in R-CNN~\cite{girshick2014rich}.
The learning rate is initially set to $0.01$ and reduced by a factor of $10$ every $40k$ iterations. The momentum is set to $0.9$ and the minibatch size is $128$.

\emph{\textbf{Multi-Scale Implementation.}}
In our system we adopt a similar multi-scale implementation as in SPP-Net~\cite{he2014spatial}.
More specifically, we apply the activation maps modules of our models on multiple scales of an image and then a single scale is selected for each region adaptation module independently.
\begin{itemize}
\item
\emph{Multi-Region CNN model:} The activation maps module is applied on $7$ scales of an image with their shorter dimension being in $\{480, 576, 688, 874, 1200, 1600, 2100\}$.
For training, the region adaptation modules are applied on a random scale and for testing, a single scale is used such that the area of the scaled region is closest to $224 \times 224$ pixels. 
In the case of rectangular ring regions, the scale is selected based on the area of the scaled outer box of the rectangular ring.
\item
\emph{Semantic Segmentation-Aware CNN model:} The activation maps module is applied on $3$ scales of an image with their shorter dimension being in $\{576, 874, 1200\}$.
For training, the region adaptation module is applied on a random scale and for testing, a single scale is selected such that the area of the scaled region is closest to $288 \times 288$ pixels.
\item
\emph{Bounding Box Regression CNN model:} The activation maps module is applied on $7$ scales of an image with their shorter dimension being in $\{480, 576, 688, 874, 1200, 1600, 2100\}$. 
Both during training and testing, a single scale is used such that the area of the scaled region is closest to $224 \times 224$ pixels.
\end{itemize}

\emph{\textbf{Training/Test Time.}}
On a Titan GPU and on PASCAL VOC2007 train+val dataset, the training time of each region adaptation module is approximately 12 hours, of the activation maps module for the semantic segmentation features is approximately 4 days, and of the linear SVM is approximately 16 hours.
In order to speed up the above steps, the activation maps (conv5\_3 features and the fc7 semantic segmentation aware features) were pre-cashed on a SSD. Finally, the per image runtime is around 30 seconds.

\section{Experimental Evaluation} \label{sec:experimental_results}
\begin{table*}[t!]
\centering
\renewcommand{\figurename}{Table}
\renewcommand{\captionlabelfont}{\bf}
\renewcommand{\captionfont}{\small} 
\resizebox{\textwidth}{!}{
{\setlength{\extrarowheight}{2pt}\scriptsize
{\begin{tabular}{l <{\hspace{-0.3em}}|>{\hspace{-0.5em}} c >{\hspace{-1em}}c >{\hspace{-1em}}c >{\hspace{-1em}}c >{\hspace{-1em}}c >{\hspace{-1em}}c >{\hspace{-1em}}c >{\hspace{-1em}}c >{\hspace{-1em}}c >{\hspace{-1em}}c >{\hspace{-1em}}c >{\hspace{-1em}}c >{\hspace{-1em}}c >{\hspace{-1em}}c >{\hspace{-1em}}c >{\hspace{-1em}}c >{\hspace{-1em}}c >{\hspace{-1em}}c >{\hspace{-1em}}c >{\hspace{-1em}}c <{\hspace{-0.3em}}| >{\hspace{-0.3em}}c}
\hline
Adaptation Modules & areo & bike & bird & boat & bottle & bus & car & cat & chair & cow & table & dog & horse & mbike & person & plant & sheep & sofa & train & tv & mAP \\
\hline
\emph{Original Box fig. \ref{fig:A}} & \textbf{0.729} & \textbf{0.715} & \textbf{0.593} & \textbf{0.478} & \textbf{0.405} & \textbf{0.713} & \textbf{0.725} & \textbf{0.741} & \textbf{0.418} & \textbf{0.694} & 0.591 & \textbf{0.713} & 0.662 & \textbf{0.725} & \textbf{0.560} & \textbf{0.312} & \textbf{0.601} & 0.565 & 0.669 & \textbf{0.731} & \textbf{0.617} \\
\emph{Left Half Box fig. \ref{fig:B}} &  0.635 & 0.659 & 0.455 & 0.364 & 0.322 & 0.621 & 0.640 & 0.589 & 0.314 & 0.620 & 0.463 & 0.573 & 0.545 & 0.641 & 0.477 & 0.300 & 0.532 & 0.442 & 0.546 & 0.621 & 0.518 \\
\emph{Right Half Box fig. \ref{fig:C}} &  0.626 & 0.605 & 0.470 & 0.331 & 0.314 & 0.607 & 0.616 & 0.641 & 0.278 & 0.487 & 0.513 & 0.548 & 0.564 & 0.585 & 0.459 & 0.262 & 0.469 & 0.465 & 0.573 & 0.620 & 0.502 \\
\emph{Up Half Box fig. \ref{fig:D}} & 0.591 & 0.651 & 0.470 & 0.266 & 0.361 & 0.629 & 0.656 & 0.641 & 0.305 & 0.604 & 0.511 & 0.604 & 0.643 & 0.588 & 0.466 & 0.220 & 0.545 & 0.528 & 0.590 & 0.570 & 0.522 \\
\emph{Bottom Half Box fig. \ref{fig:E}} & 0.607 & 0.631 & 0.406 & 0.397 & 0.233 & 0.594 & 0.626 & 0.559 & 0.285 & 0.417 & 0.404 & 0.520 & 0.490 & 0.649 & 0.387 & 0.233 & 0.457 & 0.344 & 0.566 & 0.617 & 0.471 \\
\emph{Central Region fig. \ref{fig:F}} & 0.552 & 0.622 & 0.413 & 0.244 & 0.283 & 0.502 & 0.594 & 0.603 & 0.282 & 0.523 & 0.424 & 0.516 & 0.495 & 0.584 & 0.386 & 0.232 & 0.527 & 0.358 & 0.533 & 0.587 & 0.463 \\
\emph{Central Region fig. \ref{fig:G}} & 0.674 & 0.705 & 0.547 & 0.367 & 0.337 & 0.678 & 0.698 & 0.687 & 0.381 & 0.630 & 0.538 & 0.659 & 0.667 & 0.679 & 0.507 & 0.309 & 0.557 & 0.530 & 0.611 & 0.694 & 0.573 \\
\emph{Border Region fig. \ref{fig:H}} & 0.694 & 0.696 & 0.552 & 0.470 & 0.389 & 0.687 & 0.706 & 0.703 & 0.398 & 0.631 & 0.515 & 0.660 & 0.643 & 0.686 & 0.539 & 0.307 & 0.582 & 0.537 & 0.618 & 0.717 & 0.586 \\
\emph{Border Region fig. \ref{fig:I}} & 0.651 & 0.649 & 0.504 & 0.407 & 0.333 & 0.670 & 0.704 & 0.624 & 0.323 & 0.625 & 0.533 & 0.594 & 0.656 & 0.627 & 0.517 & 0.223 & 0.533 & 0.515 & 0.604 & 0.663 & 0.548  \\
\emph{Contextual Region fig. \ref{fig:J}} & 0.624 & 0.568 & 0.425 & 0.380 & 0.255 & 0.609 & 0.650 & 0.545 & 0.222 & 0.509 & 0.522 & 0.427 & 0.563 & 0.541 & 0.431 & 0.163 & 0.482 & 0.392 & 0.597 & 0.532 & 0.472 \\
\emph{Semantic-aware region.} & 0.652 & 0.684 & 0.549 & 0.407 & 0.225 & 0.658 & 0.676 & 0.738 & 0.316 & 0.596 & \textbf{0.635} & 0.705 & \textbf{0.670} & 0.689 & 0.545 & 0.230 & 0.522 & \textbf{0.598} & \textbf{0.680} & 0.548 & 0.566 \\
\hline
\end{tabular}}}}
\vspace{1pt}
\caption{Detection performance of individual regions on VOC2007 test set. They were trained on VOC2007 train+val set.}
\label{tab:single_regions}
\vspace{-0.6em} 
\end{table*}
\begin{table*}[t!]
\centering
\renewcommand{\figurename}{Table}
\renewcommand{\captionlabelfont}{\bf}
\renewcommand{\captionfont}{\small} 
\resizebox{\textwidth}{!}{
{\setlength{\extrarowheight}{2pt}\scriptsize
{\begin{tabular}{l <{\hspace{-0.3em}}|>{\hspace{-0.5em}} c >{\hspace{-1em}}c >{\hspace{-1em}}c >{\hspace{-1em}}c >{\hspace{-1em}}c >{\hspace{-1em}}c >{\hspace{-1em}}c >{\hspace{-1em}}c >{\hspace{-1em}}c >{\hspace{-1em}}c >{\hspace{-1em}}c >{\hspace{-1em}}c >{\hspace{-1em}}c >{\hspace{-1em}}c >{\hspace{-1em}}c >{\hspace{-1em}}c >{\hspace{-1em}}c >{\hspace{-1em}}c >{\hspace{-1em}}c >{\hspace{-1em}}c <{\hspace{-0.3em}}| >{\hspace{-0.3em}}c}
\hline
Approach & areo & bike & bird & boat & bottle & bus & car & cat & chair & cow & table & dog & horse & mbike & person & plant & sheep & sofa & train & tv & mAP \\
\hline
\emph{R-CNN with VGG-Net} & 0.716 & 0.735 & 0.581 & 0.422 & 0.394 & 0.707 & 0.760  & 0.745 & 0.387 & 0.710 & 0.569 & 0.745 & 0.679 & 0.696 & 0.593 & 0.357 & 0.621 & 0.640 & 0.665  & 0.712 & 0.622 \\
\emph{R-CNN with VGG-Net \& bbox reg.} & 0.734 & 0.770 & 0.634 & 0.454 & 0.446 & 0.751 & 0.781 & 0.798 & 0.405 & 0.737 & 0.622 & 0.794 & 0.781 & 0.731 & 0.642 & 0.356 & 0.668 & 0.672  & 0.704 & 0.711 & 0.660 \\
\emph{Best approach of~\cite{yuting2015improving}}  & 0.725 & 0.788 &  0.67 & 0.452 & 0.510 & 0.738 & 0.787 & 0.783 & 0.467 & 0.738 & 0.615 & 0.771 & 0.764 & 0.739 & 0.665 & 0.392 & 0.697 & 0.594 & 0.668 & 0.729 & 0.665 \\
\emph{Best approach of~\cite{yuting2015improving} \& bbox reg.}  & 0.741 & \textbf{0.832} & 0.670 & 0.508 & 0.516 & 0.762 & 0.814 & 0.772 & 0.481 & 0.789 & 0.656 & 0.773 & 0.784 & 0.751 & 0.701 & 0.414 & 0.696 & 0.608 & 0.702 & 0.737 & 0.685 \\
\hline
\emph{Original Box fig.~\ref{fig:A}} & 0.729 & 0.715 & 0.593 & 0.478 & 0.405 & 0.713 & 0.725 & 0.741 & 0.418 & 0.694 & 0.591 & 0.713 & 0.662 & 0.725 & 0.560 & 0.312 & 0.601 & 0.565 & 0.669 & 0.731 & 0.617 \\
\emph{MR-CNN} & 0.749 & 0.757 & 0.645 & 0.549 & 0.447 & 0.741 & 0.755 & 0.760 & 0.481 & 0.724 & 0.674 & 0.765 & 0.724 & 0.749 & 0.617 & 0.348 & 0.617 & 0.640 & 0.735 & 0.760 & 0.662 \\
\emph{MR-CNN \& S-CNN} &  0.768 & 0.757 & 0.676 & 0.551 & 0.456 & 0.776 & 0.765 & 0.784 & 0.467 & 0.747 & 0.688 & 0.793 & 0.742 & 0.770 & 0.625 & 0.374 & 0.643 & 0.638 & 0.740 & 0.747 & 0.675 \\
\emph{MR-CNN \& S-CNN \& Loc.}  & \textbf{0.787} & 0.818 & \textbf{0.767} & \textbf{0.666} & \textbf{0.618} & \textbf{0.817} & \textbf{0.853} & \textbf{0.827} & \textbf{0.570} & \textbf{0.819} & \textbf{0.732} & \textbf{0.846} & \textbf{0.860} & \textbf{0.805} & \textbf{0.749} & \textbf{0.449} & \textbf{0.717} & \textbf{0.697} & \textbf{0.787} & \textbf{0.799} & \textbf{0.749} \\
\hline
\end{tabular}}}}
\vspace{1pt}
\caption{\small{Detection performance of our modules on VOC2007 test set. Each model was trained on VOC2007 train+val set.}}
\label{tab:multi_regions}
\vspace{-0.6em} 
\end{table*}
\begin{table*}[t!]
\centering
\renewcommand{\figurename}{Table}
\renewcommand{\captionlabelfont}{\bf}
\renewcommand{\captionfont}{\small} 
\resizebox{\textwidth}{!}{
{\setlength{\extrarowheight}{2pt}\scriptsize
{\begin{tabular}{l <{\hspace{-0.3em}}|>{\hspace{-0.5em}} c >{\hspace{-1em}}c >{\hspace{-1em}}c >{\hspace{-1em}}c >{\hspace{-1em}}c >{\hspace{-1em}}c >{\hspace{-1em}}c >{\hspace{-1em}}c >{\hspace{-1em}}c >{\hspace{-1em}}c >{\hspace{-1em}}c >{\hspace{-1em}}c >{\hspace{-1em}}c >{\hspace{-1em}}c >{\hspace{-1em}}c >{\hspace{-1em}}c >{\hspace{-1em}}c >{\hspace{-1em}}c >{\hspace{-1em}}c >{\hspace{-1em}}c <{\hspace{-0.3em}}| >{\hspace{-0.3em}}c}
\hline
Approach & areo & bike & bird & boat & bottle & bus & car & cat & chair & cow & table & dog & horse & mbike & person & plant & sheep & sofa & train & tv & mAP \\
\hline
\emph{R-CNN with VGG-Net from~\cite{yuting2015improving}} & 0.402 & 0.433 & 0.234 & 0.144 & 0.133 & 0.482 & 0.445 & 0.364 & 0.171 & 0.340 & 0.279 & 0.363 & 0.268 & 0.282 & 0.212 & 0.103 & 0.337 & 0.366 & 0.316 & 0.489 & 0.308 \\
\emph{Best approach of~\cite{yuting2015improving}} & 0.463 & 0.581 & 0.311 & 0.216 & 0.258 & 0.571 & 0.582 & 0.435 & 0.230 & 0.464 & 0.290 & 0.407 & 0.406 & 0.463 & 0.334 & 0.106 & 0.413 & 0.409 & 0.458 & 0.563 & 0.398 \\
\emph{Best approach of~\cite{yuting2015improving} \& bbox reg.}  & 0.471 & \textbf{0.618} & 0.352 & 0.181 & 0.297 & \textbf{0.660} & 0.647 & 0.480 & \textbf{0.253} & 0.504 & 0.349 & 0.437 & 0.508 & 0.494 & 0.368 & 0.137 & 0.447 & 0.436 & 0.498 & \textbf{0.605} & 0.437 \\
\hline
\emph{Original Candidate Box} & 0.449 & 0.426 & 0.237 & 0.175 & 0.157 & 0.441 & 0.444 & 0.377 & 0.182 & 0.295 & 0.303 & 0.312 & 0.249 & 0.332 & 0.187 & 0.099 & 0.302 & 0.286 & 0.337 & 0.499 & 0.305 \\
\emph{MR-CNN} & 0.495 & 0.505 & 0.292 & 0.235 & 0.179 & 0.513 & 0.504 & 0.481 & 0.206 & 0.381 & 0.375 & 0.387 & 0.296 & 0.403 & 0.239 & 0.151 & 0.341 & 0.389 & 0.422 & 0.521 & 0.366 \\
\emph{MR-CNN \& S-CNN} & 0.507 & 0.523 & 0.316 & 0.266 & 0.177 & 0.547 & 0.513 & 0.492 & 0.210 & 0.450 & 0.361 & 0.433 & 0.309 & 0.408 & 0.246 & 0.151 & 0.359 & 0.427 & 0.438 & 0.534 & 0.383 \\
\emph{MR-CNN \& S-CNN \& Loc.} & \textbf{0.549} & 0.613 & \textbf{0.430} & \textbf{0.315} & \textbf{0.383} & 0.646 & \textbf{0.650} & \textbf{0.512} & \textbf{0.253} & \textbf{0.544} & \textbf{0.505} & \textbf{0.521} & \textbf{0.591} & \textbf{0.540} & \textbf{0.393} & \textbf{0.159} & \textbf{0.485} & \textbf{0.468} & \textbf{0.553} & 0.573 & \textbf{0.484} \\
\hline
\end{tabular}}}}
\vspace{1pt}
\caption{
Detection performance of our modules on VOC2007 test set.
In this case, the IoU overlap threshold for positive detections is $0.7$.
Each model was trained on VOC2007 train+val set.}
\label{tab:multi_regions_IoU07}
\vspace{-0.6em} 
\end{table*}

We evaluate our detection system on PASCAL VOC2007~\cite{everingham2008pascal} and on PASCAL VOC2012~\cite{everingham2012pascal}.
During the presentation of the results, we will use as baseline either the \emph{Original candidate box} region alone (figure~\ref{fig:A}) and/or the R-CNN framework with VGG-Net~\cite{simonyan2014very}.
We note that, when the \emph{Original candidate box} region alone is used then the resulted model is a realization of the SPP-Net~\cite{he2014spatial} object detection framework with the 16-layers VGG-Net~\cite{simonyan2014very}.
Except if otherwise stated, for all the PASCAL VOC2007 results, we trained our models on the trainval set and tested them on the test set of the same year.

\subsection{Results on PASCAL VOC2007}

First, we asses the significance of each of the region adaptation modules alone on the object detection task. 
Results are reported in table \ref{tab:single_regions}.
As we expected, the best performing component is the \emph{Original candidate box}. 
What is surprising is the high detection performance of individual regions like the \emph{Border Region} on figure \ref{fig:I} $54.8\%$ or the \emph{Contextual Region} on figure \ref{fig:J} $47.2\%$. 
Despite the fact that the area visible by them includes limited or not at all portion of the object, they outperform previous detection systems that were based on hand crafted features.
Also interesting, is the high detection performance of the semantic segmentation aware region, $56.6\%$. 

In table \ref{tab:multi_regions}, we report the detection performance of our proposed modules.
The Multi-Region CNN model without the semantic segmentation aware CNN features (\emph{MR-CNN}), achieves $66.2\%$ mAP, which is $4.2$ points higher than \emph{R-CNN with VGG-Net} ($62.0\%$) and $4.5$ points higher than the \emph{Original candidate box} region alone ($61.7\%$). 
Moreover, its detection performance slightly exceeds that of \emph{R-CNN with VGG-Net} and bounding box regression ($66.0\%$).
Extending the Multi-Region CNN model with the semantic segmentation aware CNN features (\emph{MR-CNN \& S-CNN}), boosts the performance of our recognition model another $1.3$ points and reaches the total of $67.5\%$ mAP.
Comparing to the recently published method of Yuting et al.~\cite{yuting2015improving}, 
our \emph{MR-CNN \& S-CNN} model scores 
$1$ point higher than their best performing method that includes 
generation of extra box proposals via Bayesian optimization and structured loss during the fine-tuning of the VGG-Net.
Significant is also the improvement that we get when we couple our recognition model with the CNN model for bounding box regression under the iterative localization scheme proposed (\emph{MR-CNN \& S-CNN \& Loc.}). 
Specifically, the detection performance is raised from $67.5\%$ to $74.9\%$ setting the new state-of-the-art on this test set and for this set of training data (VOC2007 train+val set).

In table \ref{tab:multi_regions_IoU07}, we report the detection performance of our system when the overlap threshold for considering a detection positive is set to $0.7$. This metric was proposed from~\cite{yuting2015improving} in order to reveal the localization capability of their method. 
From the table we observe that each of our modules exhibit very good localization capability, which was our goal when designing them, and our overall system exceeds in that metric the approach of~\cite{yuting2015improving}.

\subsection{Detection error analysis} \label{sec:det_error_an}
\begin{figure*}[t!]
\center
\renewcommand{\figurename}{Figure}
\renewcommand{\captionlabelfont}{\bf}
\renewcommand{\captionfont}{\small} 
        \begin{center}
        \begin{subfigure}[b]{0.22\textwidth}
                \includegraphics[width=\textwidth]{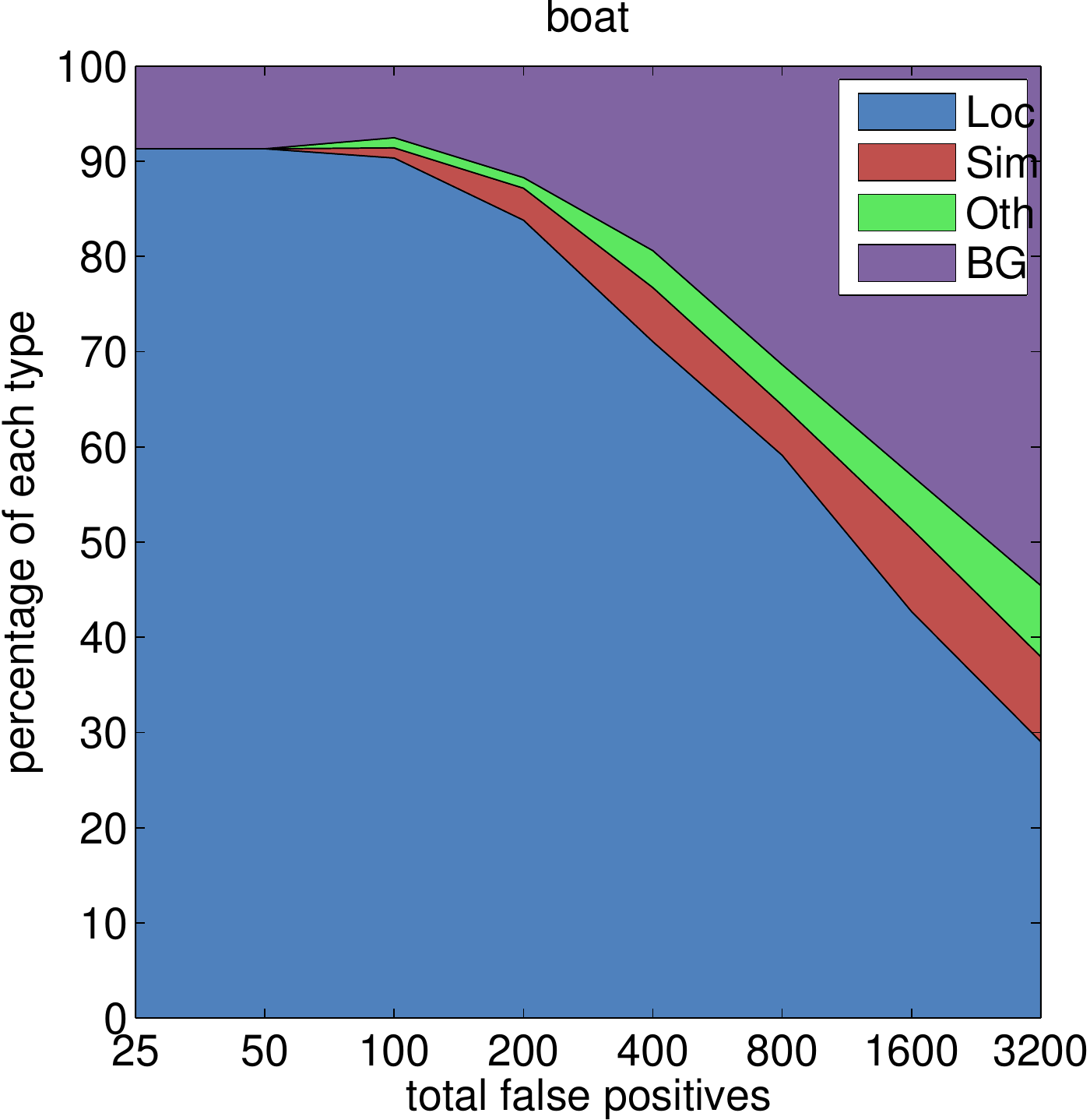}
        \end{subfigure}%
        \hspace{0.05cm} 
       \begin{subfigure}[b]{0.22\textwidth}
               \includegraphics[width=\textwidth]{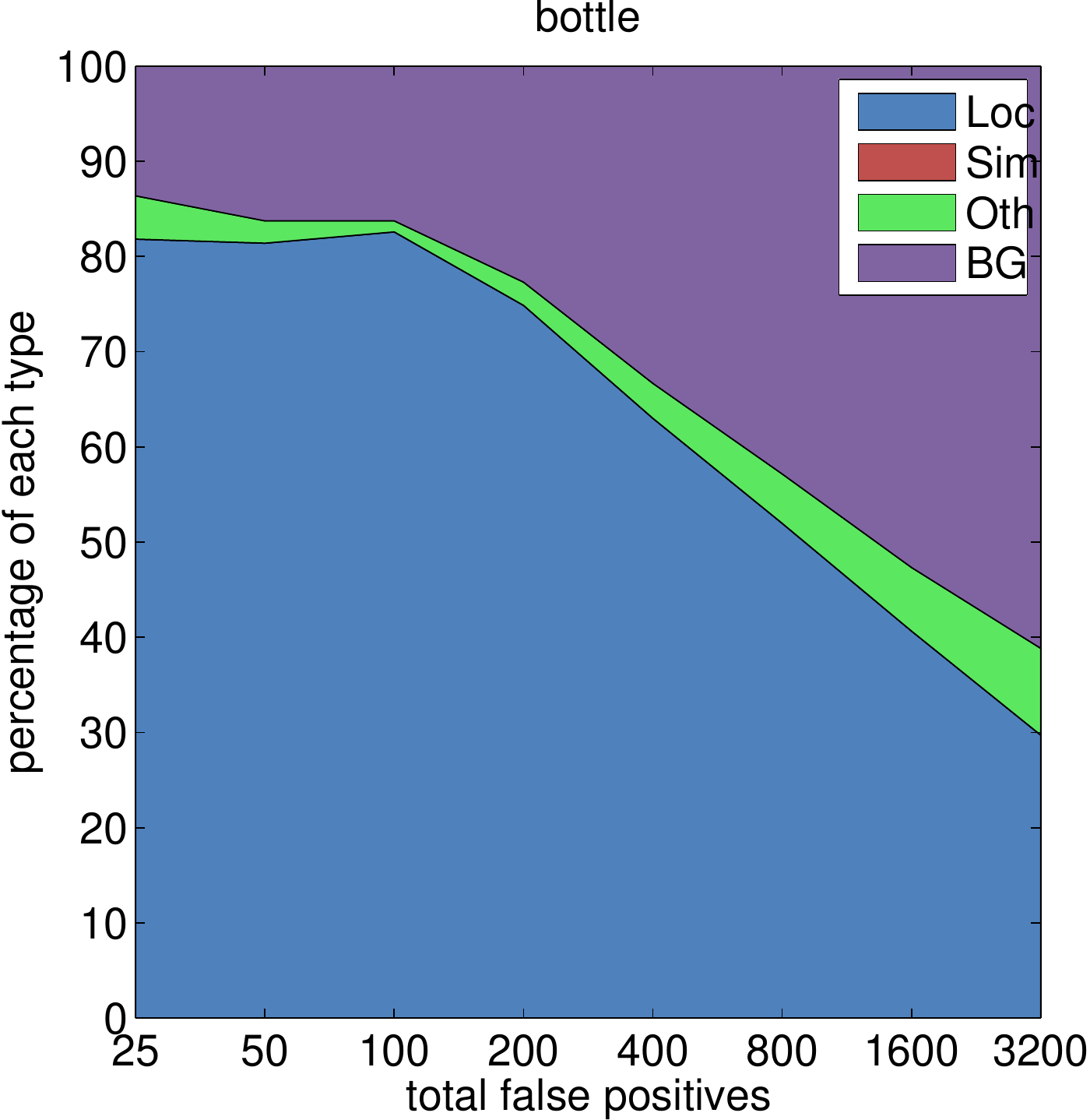}
       \end{subfigure}
        \hspace{0.05cm}
       \begin{subfigure}[b]{0.22\textwidth}
                \includegraphics[width=\textwidth]{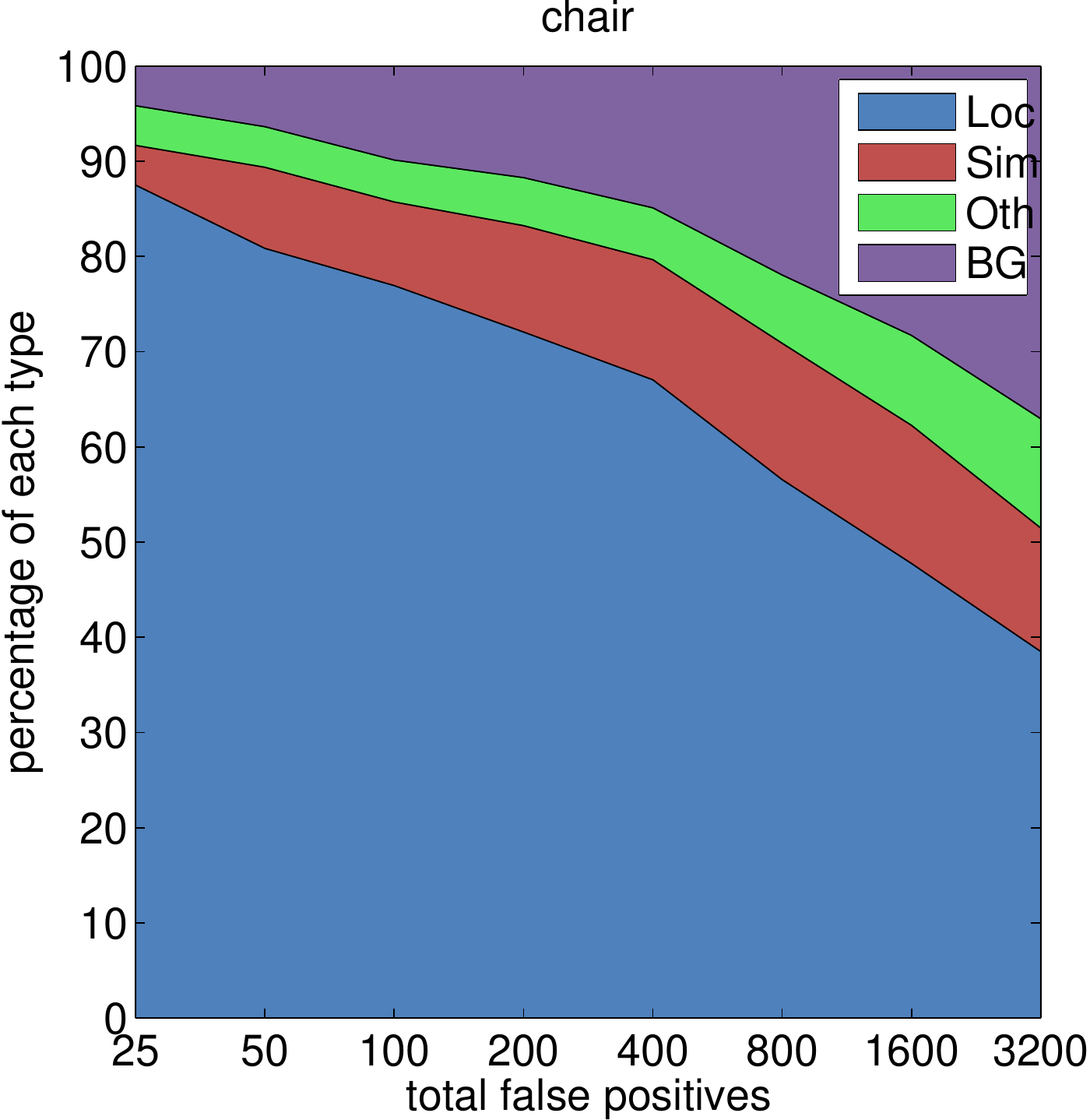}
        \end{subfigure}
        \hspace{0.05cm}
        \begin{subfigure}[b]{0.22\textwidth}
                \includegraphics[width=\textwidth]{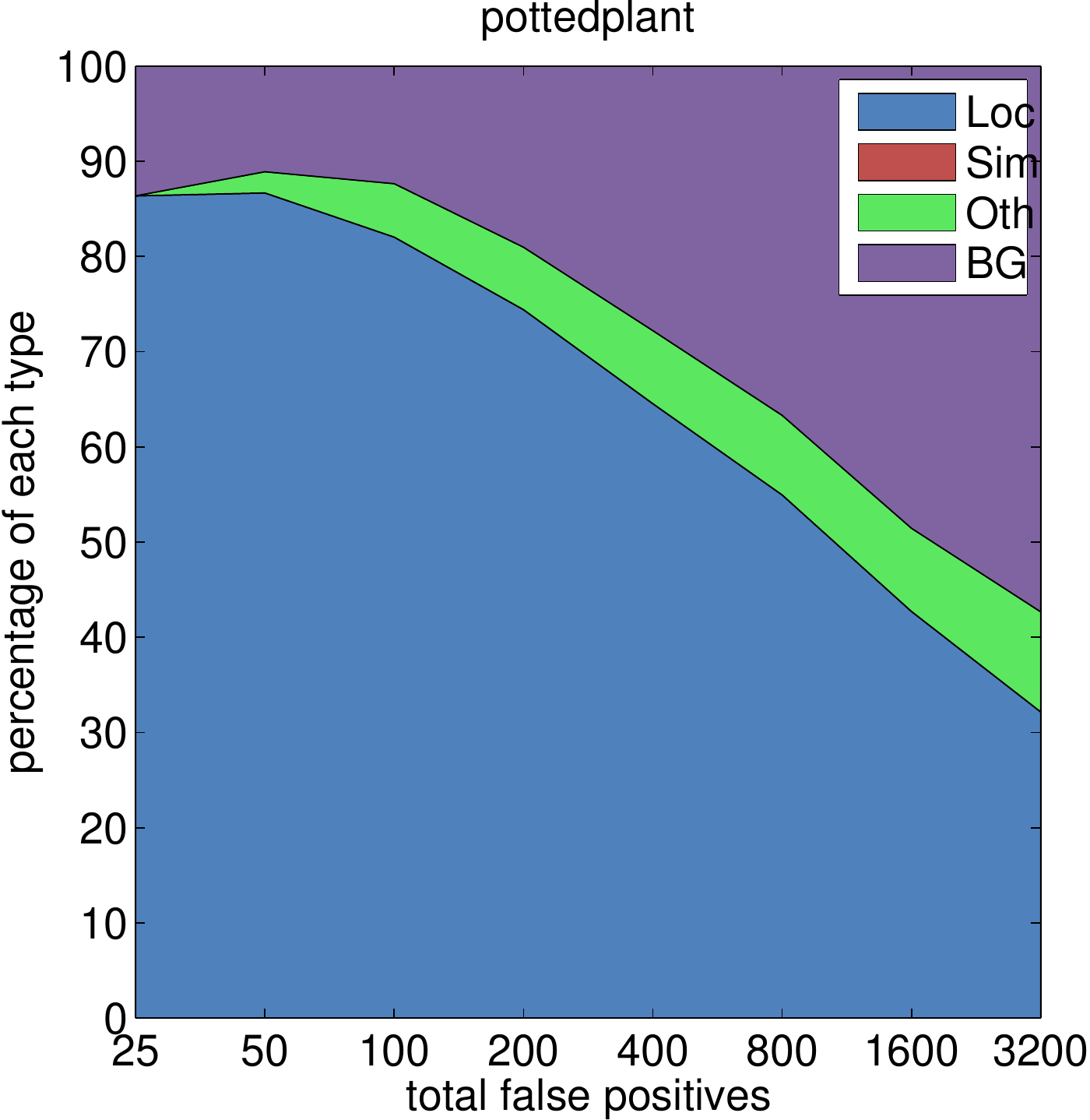}
        \end{subfigure}
        
        \begin{subfigure}[b]{0.22\textwidth}
               \includegraphics[width=\textwidth]{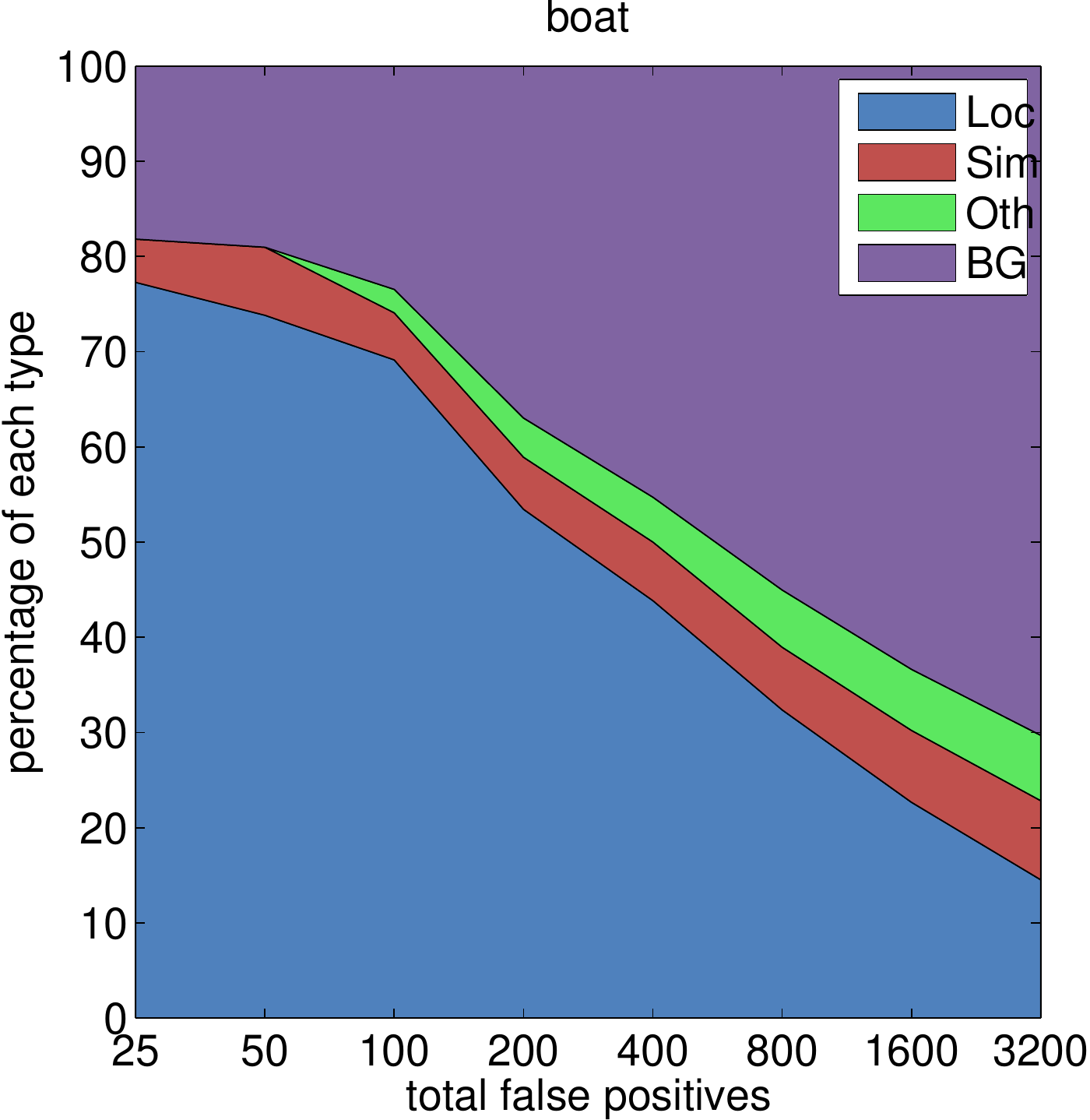}
        \end{subfigure}%
        \hspace{0.05cm} 
        \begin{subfigure}[b]{0.22\textwidth}
                \includegraphics[width=\textwidth]{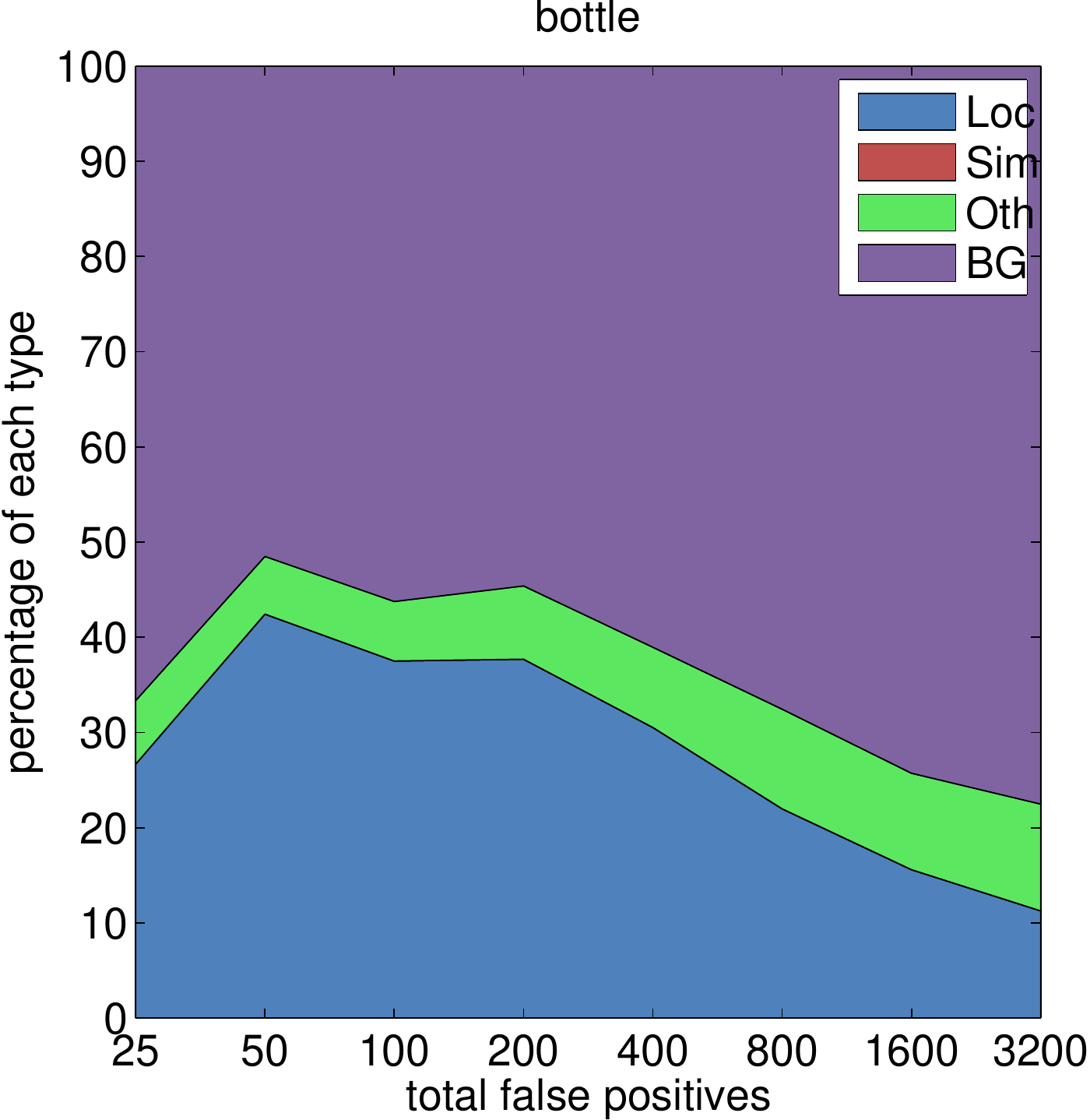}
        \end{subfigure}
        \hspace{0.05cm} 
        \begin{subfigure}[b]{0.22\textwidth}
                \includegraphics[width=\textwidth]{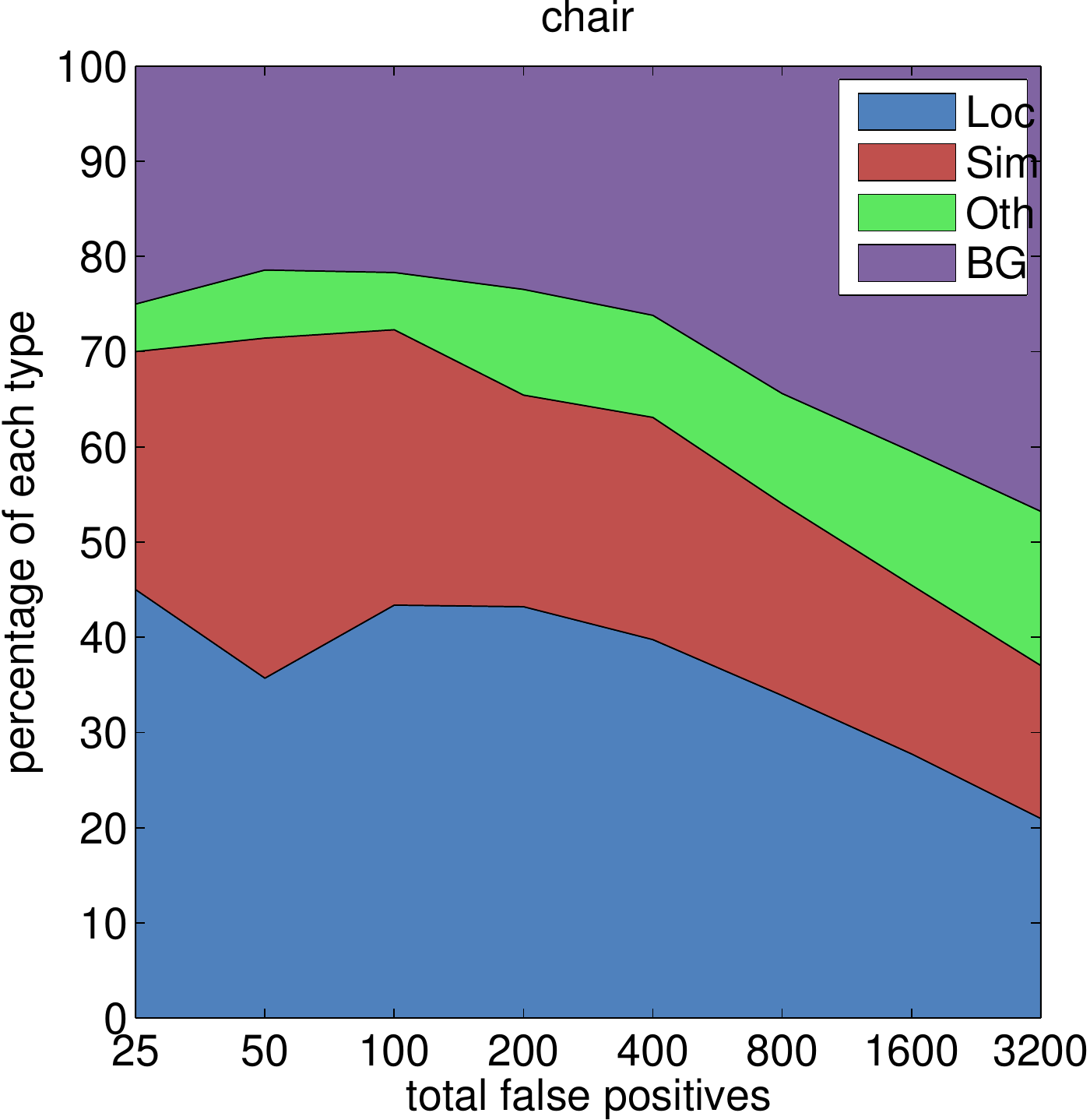}
        \end{subfigure}
        \hspace{0.05cm}
        \begin{subfigure}[b]{0.22\textwidth}
                \includegraphics[width=\textwidth]{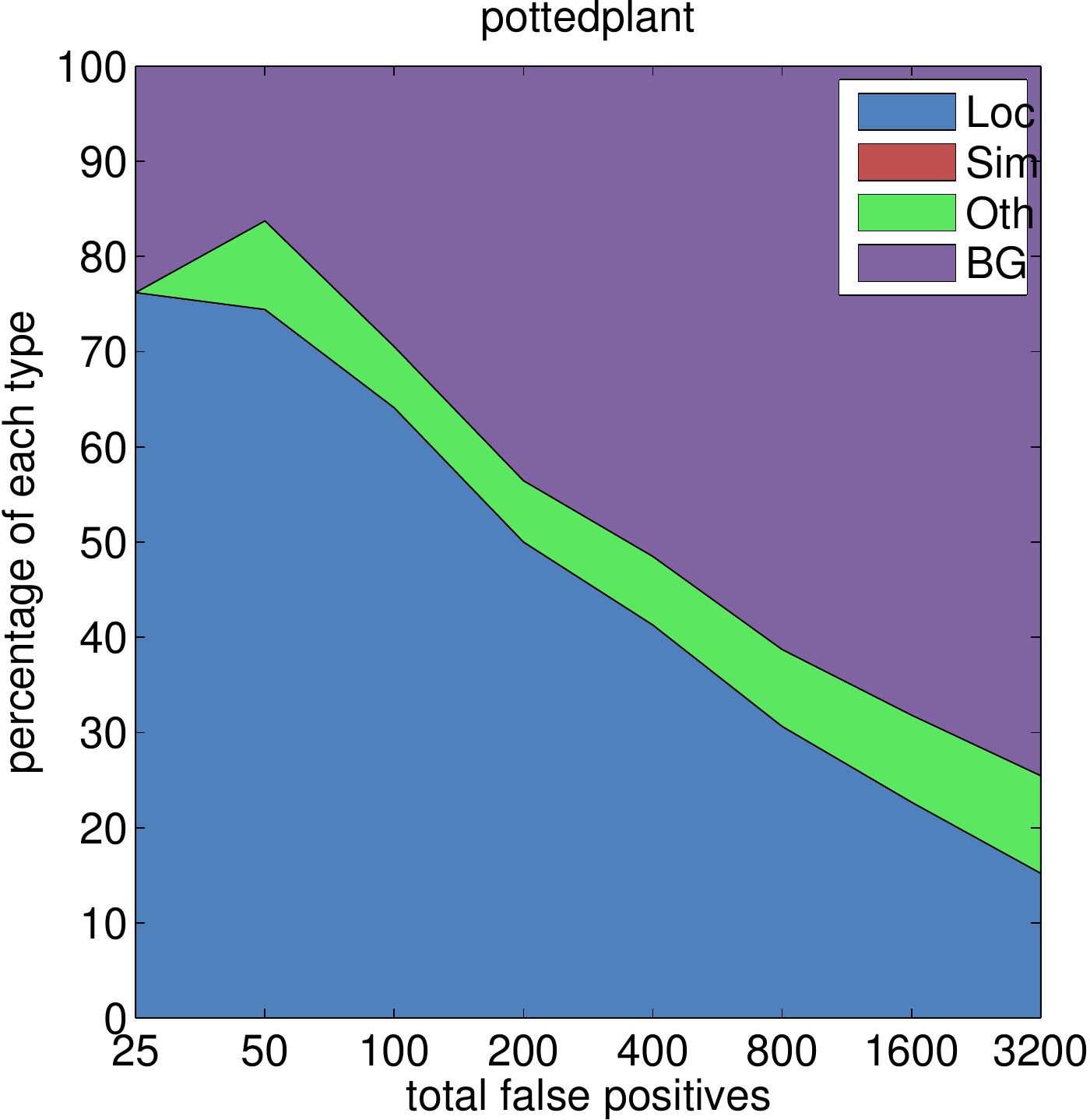}
        \end{subfigure}                
        \end{center}
        \vspace{1pt}
        \caption{Top ranked false positive types.
\textbf{Top row:} our baseline which is the \emph{original candidate box} only model. 
\textbf{Bottom row:} our overall system. 
We present only the graphs for the classes boat, bottle, chair, and pottedplant because of space limitations and the fact that they are the most difficult ones of PASCAL VOC challenge.}
       \label{fig:TopRankedFalsePositives}
       \vspace{-0.6em} 
\end{figure*}
\begin{figure}[t!]
\center
\renewcommand{\figurename}{Figure}
\renewcommand{\captionlabelfont}{\bf}
\renewcommand{\captionfont}{\small} 
        \begin{center}
                \begin{center}
        \begin{subfigure}[b]{0.152\textwidth}
                \includegraphics[width=\textwidth]{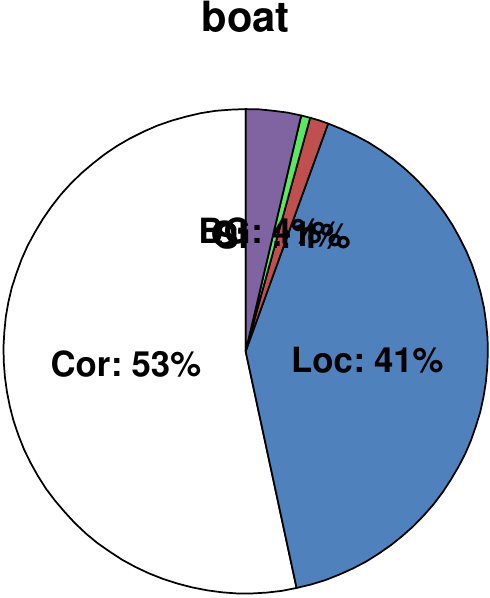}
        \end{subfigure}%
        \hspace{0.01cm} 
        \begin{subfigure}[b]{0.152\textwidth}
                \includegraphics[width=\textwidth]{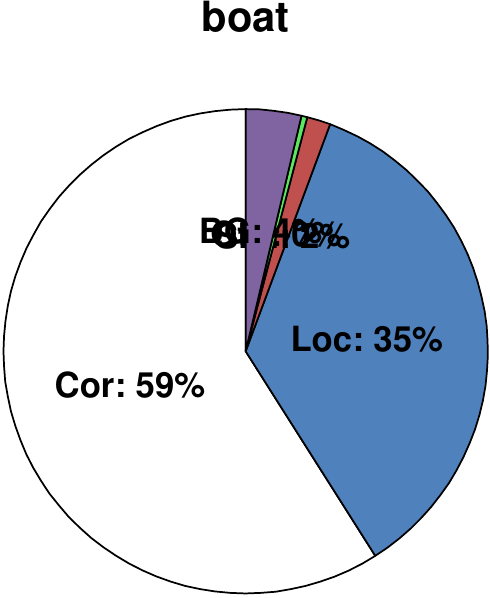}
        \end{subfigure}
        \hspace{0.01cm} 
        \begin{subfigure}[b]{0.152\textwidth}
                \includegraphics[width=\textwidth]{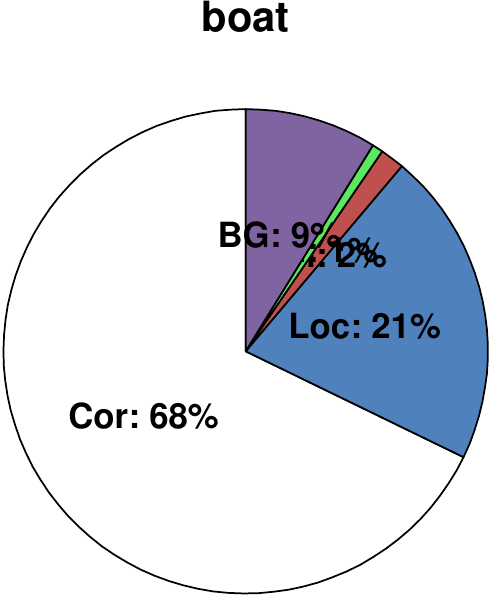}
        \end{subfigure}
        \vspace{1pt}
        \begin{subfigure}[b]{0.152\textwidth}
                \includegraphics[width=\textwidth]{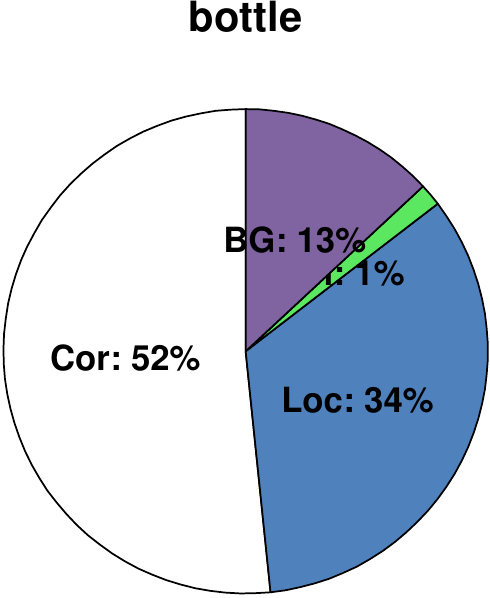}
        \end{subfigure}
        \hspace{0.01cm} 
        \begin{subfigure}[b]{0.152\textwidth}
                \includegraphics[width=\textwidth]{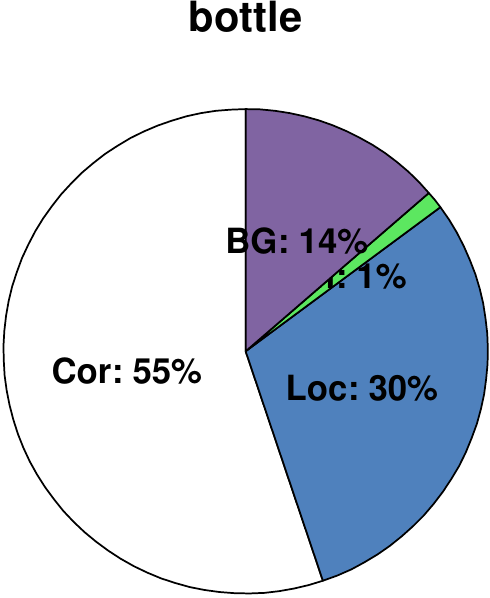}
        \end{subfigure}
        \hspace{0.01cm}
        \begin{subfigure}[b]{0.152\textwidth}
                \includegraphics[width=\textwidth]{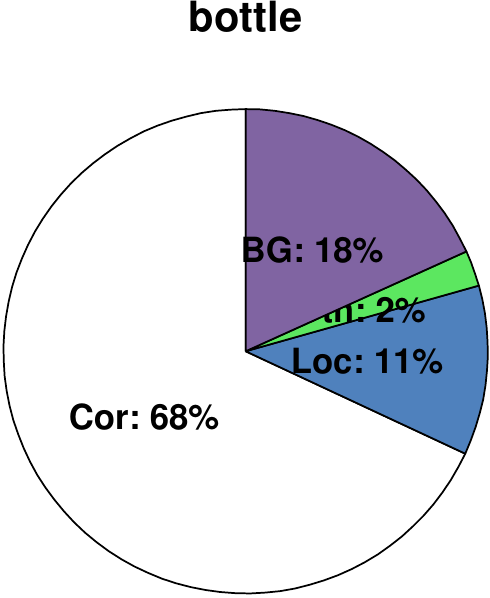}
        \end{subfigure} 
        \vspace{1pt}
        \begin{subfigure}[b]{0.152\textwidth}
                \includegraphics[width=\textwidth]{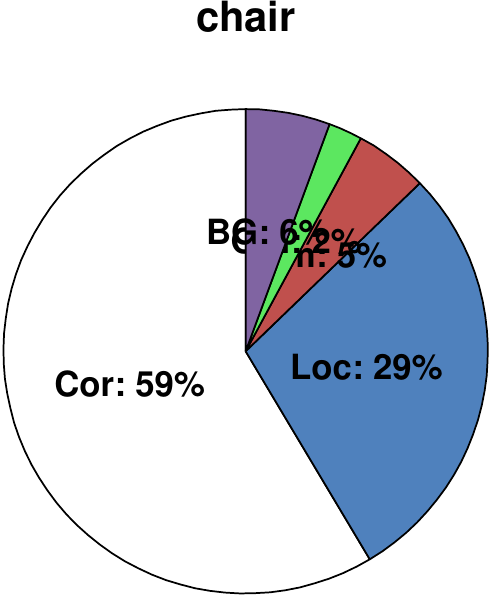}
        \end{subfigure}
        \hspace{0.01cm} 
        \begin{subfigure}[b]{0.152\textwidth}
                \includegraphics[width=\textwidth]{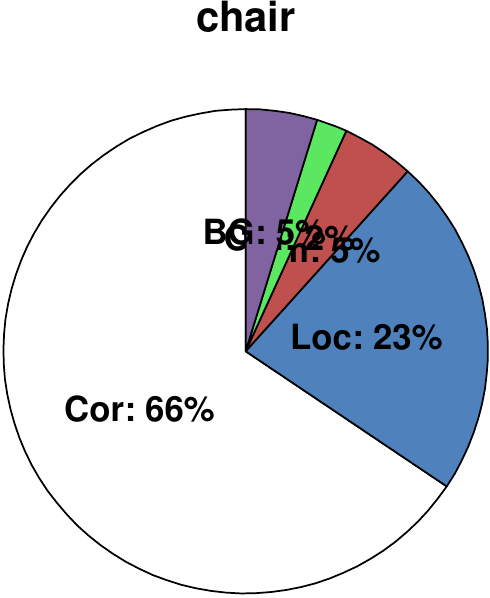}
        \end{subfigure}
        \hspace{0.01cm}
        \begin{subfigure}[b]{0.152\textwidth}
                \includegraphics[width=\textwidth]{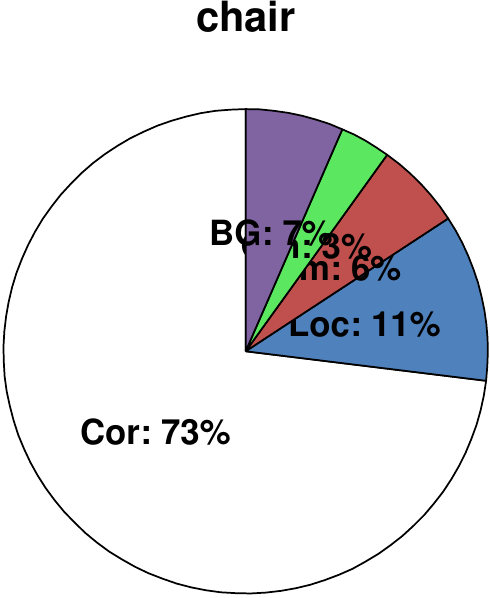}
        \end{subfigure} 
        \vspace{1pt}
        \begin{subfigure}[b]{0.152\textwidth}
                \includegraphics[width=\textwidth]{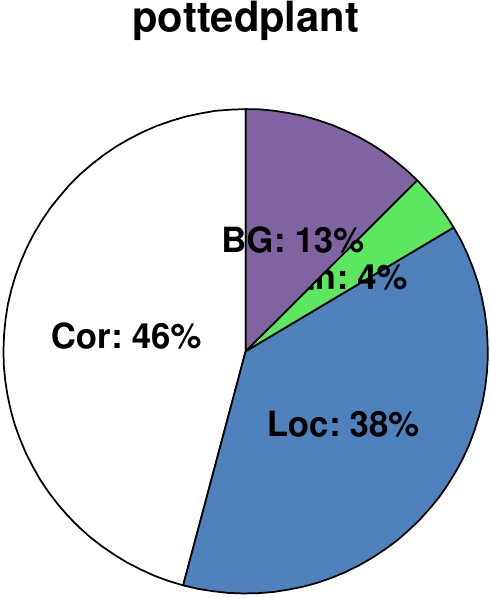}
        \end{subfigure}
        \hspace{0.01cm} 
        \begin{subfigure}[b]{0.152\textwidth}
                \includegraphics[width=\textwidth]{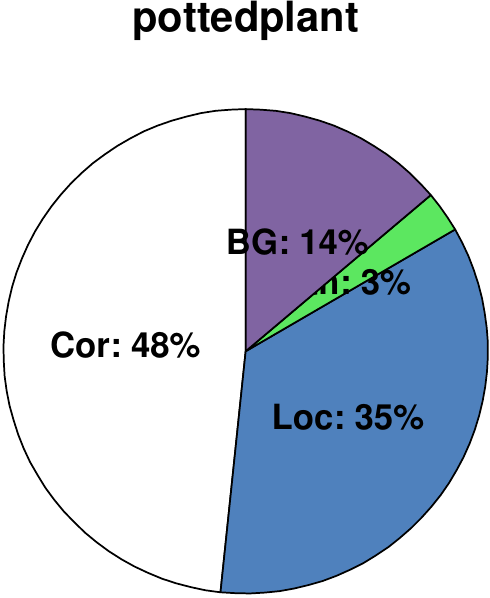}
        \end{subfigure}
        \hspace{0.01cm}
        \begin{subfigure}[b]{0.152\textwidth}
                \includegraphics[width=\textwidth]{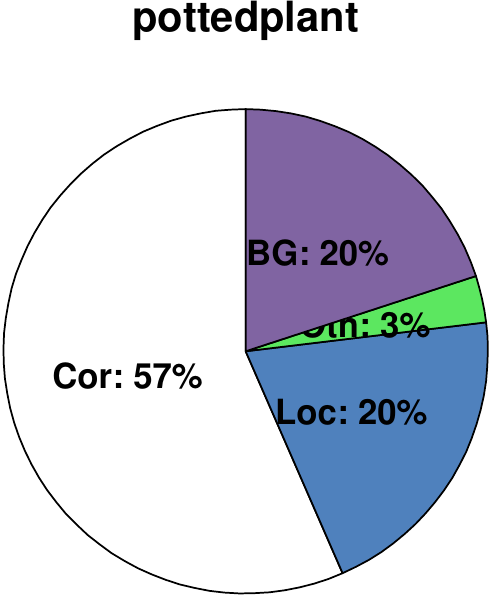}
        \end{subfigure} 
        \end{center}                
        \end{center}
        \vspace{1pt}
        \caption{Fraction of top N detections (N=num of objs in category) that are correct (Cor), or false positives due to poor localization (Loc), confusion with similar objects (Sim), confusion with other VOC objects (Oth), or confusion with background or unlabelled objects (BG). 
\textbf{Left column:} our baseline which is the \emph{original candidate box} only model.
\textbf{Middle column:} Multi-Region CNN model without the semantic segmentation aware CNN features.
\textbf{Right column:} our overall system. We present only the pie charts for the classes boat, bottle, chair, and pottedplant because of space limitations and the fact that they are the most difficult categories of PASCAL VOC challenge.}
        \label{fig:DetectionAnalysis}
        \vspace{-0.6em} 
\end{figure}
\begin{table*}[t!]
\centering
\renewcommand{\figurename}{Table}
\renewcommand{\captionlabelfont}{\bf}
\renewcommand{\captionfont}{\small} 
\resizebox{\textwidth}{!}{
{\setlength{\extrarowheight}{2pt}\scriptsize
{\begin{tabular}{l <{\hspace{-0.3em}}|>{\hspace{-0.5em}} c >{\hspace{-1em}}c >{\hspace{-1em}}c >{\hspace{-1em}}c >{\hspace{-1em}}c >{\hspace{-1em}}c >{\hspace{-1em}}c >{\hspace{-1em}}c >{\hspace{-1em}}c >{\hspace{-1em}}c >{\hspace{-1em}}c >{\hspace{-1em}}c >{\hspace{-1em}}c >{\hspace{-1em}}c >{\hspace{-1em}}c >{\hspace{-1em}}c >{\hspace{-1em}}c >{\hspace{-1em}}c >{\hspace{-1em}}c >{\hspace{-1em}}c <{\hspace{-0.3em}}}
\hline
Approach & areo & bike & bird & boat & bottle & bus & car & cat & chair & cow & table & dog & horse & mbike & person & plant & sheep & sofa & train & tv \\
\hline
\emph{Original candidate box-Baseline} & 0.7543 & 0.7325 & 0.6634 & 0.5816 & 0.5775 & 0.7109 & 0.7390 & 0.7277 & 0.5718 & 0.7112 & 0.6007 & 0.7000 & 0.7039 & 0.7194 & 0.6607 & 0.5339 & 0.6855 & 0.6461 & 0.6903 & 0.7359 \\
\emph{MR-CNN} & 0.7938 & 0.7864 & 0.7180 & 0.6424 & 0.6222 & 0.7609 & 0.7918 & 0.7758 & 0.6186 & 0.7483 & 0.6802 & 0.7448 &  0.7562 & 0.7569 & 0.7166 & 0.5753 & 0.7268 & 0.7148 & 0.7391 & 0.7556 \\
\hline
\end{tabular}}}}
\vspace{1pt}
\caption{Correlation between the IoU overlap of selective search box proposals~\cite{van2011segmentation} (with the closest ground truth bounding box) and the scores assigned to them.}
\label{tab:Correlation_IoU_Scores}
\vspace{-0.6em} 
\end{table*}
\begin{table*}[t!]
\centering
\renewcommand{\figurename}{Table}
\renewcommand{\captionlabelfont}{\bf}
\renewcommand{\captionfont}{\small} 
\resizebox{\textwidth}{!}{
{\setlength{\extrarowheight}{2pt}\scriptsize
{\begin{tabular}{l <{\hspace{-0.3em}}|>{\hspace{-0.5em}} c >{\hspace{-1em}}c >{\hspace{-1em}}c >{\hspace{-1em}}c >{\hspace{-1em}}c >{\hspace{-1em}}c >{\hspace{-1em}}c >{\hspace{-1em}}c >{\hspace{-1em}}c >{\hspace{-1em}}c >{\hspace{-1em}}c >{\hspace{-1em}}c >{\hspace{-1em}}c >{\hspace{-1em}}c >{\hspace{-1em}}c >{\hspace{-1em}}c >{\hspace{-1em}}c >{\hspace{-1em}}c >{\hspace{-1em}}c >{\hspace{-1em}}c <{\hspace{-0.3em}}}
\hline
Approach & areo & bike & bird & boat & bottle & bus & car & cat & chair & cow & table & dog & horse & mbike & person & plant & sheep & sofa & train & tv \\
\hline
\emph{Original candidate box-Baseline} & 0.9327 & 0.9324 & 0.9089 & 0.8594 & 0.8570 & 0.9389 & 0.9455 & 0.9250 & 0.8603 & 0.9237 & 0.8806 & 0.9209 & 0.9263 & 0.9317 & 0.9151 & 0.8415 & 0.8932 & 0.9060 & 0.9241 & 0.9125\\
\emph{MR-CNN} & 0.9462 & 0.9479 & 0.9282 & 0.8843 & 0.8740 & 0.9498 & 0.9593 & 0.9355 & 0.8790 & 0.9338 & 0.9127 & 0.9358 & 0.9393 & 0.9440 & 0.9341 & 0.8607 & 0.9120 & 0.9314 & 0.9413 & 0.9210\\
\hline
\end{tabular}}}}
\vspace{1pt}
\caption{The Area-Under-Curve (AUC) measure for the well-localized box proposals against the mis-localized box proposals.}
\label{tab:AUC_from_ROC}
\vspace{-0.6em} 
\end{table*}

We use the tool of Hoiem et al.~\cite{hoiem2012diagnosing} to analyse the detection errors of our system. 
In figure~\ref{fig:DetectionAnalysis}, we plot pie charts with the percentage of detections that are false positive due to bad localization, confusion with similar category, confusion with other category, and triggered on the background or an unlabelled object.
We use the tool of Hoiem et al.~\cite{hoiem2012diagnosing} to analyse the detection errors of our system. 
In figure \ref{fig:DetectionAnalysis}, we plot pie charts with the percentage of detections that are false positive due to bad localization, confusion with similar category, confusion with other category, and triggered on the background or an unlabelled object.
We observe that, by using the Multi-Region CNN model instead of the \emph{Original Candidate Box} region alone, a considerable reduction in the percentage of false positives due to bad localization is achieved. 
This validates our argument that focusing on multiple regions of an object increases the localization sensitivity of our model.
Furthermore, when our recognition model is integrated on the localization module developed for it, the reduction of false positives due to bad localization is huge. A similar observation can be deducted from figure~\ref{fig:TopRankedFalsePositives} where we plot the top-ranked false positive types of the baseline and of our overall proposed system. 

\subsection{Localization awareness of Multi-Region CNN model} \label{sec:loc_awareness}
Two extra experiments are presented here that indicate the localization awareness of our Multi-Region CNN model without the semantic segmentation aware CNN features (\emph{MR-CNN}) against the model that uses only the original candidate box (\emph{Baseline}). 

\emph{\textbf{Correlation between the scores and the IoU overlap of box proposals.}} 
In this experiment, we estimate the correlation between the IoU overlap of box proposals~\cite{van2011segmentation} (with the closest ground truth bounding box) and the score assigned to them from the two examined models. 
High correlation coefficient means that better localized box proposals will tend to be scored higher than mis-localized ones.
We report the correlation coefficients of the aforementioned quantities both for the \emph{Baseline} and  \emph{MR-CNN} models in table~\ref{tab:Correlation_IoU_Scores}.
Because with this experiment we want to emphasize on the localization aspect of the Multi-Region CNN model, we use proposals that overlap with the ground truth bounding boxes by at least $0.1$ IoU.

\emph{\textbf{Area-Under-the-Curve of well-localized proposals against mis-localized proposals.}} 
The ROC curves are typically used to illustrate the capability of a classifier to distinguish between two classes.
This discrimination capability can be measured by computing the Area-Under-the-Curve (AUC) metric.
The higher the AUC measure is, the more discriminative is the classifier between the two classes.
In our case, the set of well-localized box proposals is the positive class and the set of miss-localized box proposals is the negative class.
As well-localized are considered the box proposals that overlap with a ground-truth bounding box in the range $[0.5, 1.0]$ and as mis-localized are considered the box proposals that overlap with a ground truth bounding box in the range $[0.1, 0.5)$.
In table~\ref{tab:AUC_from_ROC}, we report the AUC measure for each class separately and both for the \emph{MR-CNN} and the \emph{Baseline} models.

\subsection{Results on PASCAL VOC2012}
\begin{table*}[t!]
\centering
\renewcommand{\figurename}{Table}
\renewcommand{\captionlabelfont}{\bf}
\renewcommand{\captionfont}{\small} 
\resizebox{\textwidth}{!}{
{\setlength{\extrarowheight}{2pt}\scriptsize
{\begin{tabular}{l <{\hspace{-0.3em}}|>{\hspace{-0.5em}} l | >{\hspace{-0.5em}}c >{\hspace{-1em}}c >{\hspace{-1em}}c >{\hspace{-1em}}c >{\hspace{-1em}}c >{\hspace{-1em}}c >{\hspace{-1em}}c >{\hspace{-1em}}c >{\hspace{-1em}}c >{\hspace{-1em}}c >{\hspace{-1em}}c >{\hspace{-1em}}c >{\hspace{-1em}}c >{\hspace{-1em}}c >{\hspace{-1em}}c >{\hspace{-1em}}c >{\hspace{-1em}}c >{\hspace{-1em}}c >{\hspace{-1em}}c >{\hspace{-1em}}c <{\hspace{-0.3em}}| >{\hspace{-0.3em}}c}
\hline
Approach & trained on & areo & bike & bird & boat & bottle & bus & car & cat & chair & cow & table & dog & horse & mbike & person & plant & sheep & sofa & train & tv & mAP \\
\hline
\emph{R-CNN~\cite{girshick2014rich} with VGG-Net \& bbox reg.} & VOC12 & 0.792 & 0.723 & 0.629 & 0.437 & 0.451 & 0.677 & 0.667 & 0.830 & 0.393 & 0.662 & 0.517 & 0.822 & 0.732 & 0.765 & 0.642 & 0.337 & 0.667 & 0.561 & 0.683 & 0.610 & 0.630 \\
\emph{Network In Network~\cite{LinCY2013NIN}} & VOC12 & 0.802 & 0.738 & 0.619 & 0.437 & 0.430 & 0.703 & 0.676 & 0.807 & 0.419 & 0.697 & 0.517 & 0.782 & 0.752 & 0.769 & 0.651 & 0.386 & 0.683 & 0.580 & 0.687 & 0.633 & 0.638 \\
\emph{Best approach of~\cite{yuting2015improving} \& bbox reg.} & VOC12 & 0.829 & 0.761 & 0.641 & 0.446 & 0.494 & 0.703 & 0.712 & \textbf{0.846} & 0.427 & 0.686 & 0.558 & 0.827 & 0.771 & 0.799 & 0.687 & 0.414 & \textbf{0.690} & 0.600 & 0.720 & 0.662 & 0.664 \\
\hline
\emph{MR-CNN \& S-CNN \& Loc. (Ours)} & VOC07 & 0.829 & 0.789 & 0.708 & 0.528 & 0.555 & 0.737 & 0.738 & 0.843 & 0.480 & 0.702 & 0.571 & 0.845 & 0.769 & \textbf{0.819} & 0.755 & \textbf{0.426} & 0.685 & 0.599 & 0.728 & 0.717 & 0.691 \\
\emph{MR-CNN \& S-CNN \& Loc. (Ours)} & VOC12 & \textbf{0.850} & \textbf{0.796} & \textbf{0.715} & \textbf{0.553} & \textbf{0.577} & \textbf{0.760} & \textbf{0.739} & \textbf{0.846} & \textbf{0.505} & \textbf{0.743} & \textbf{0.617} & \textbf{0.855} & \textbf{0.799} & 0.817 & \textbf{0.764} & 0.410 & \textbf{0.690} & \textbf{0.612} & \textbf{0.777} & \textbf{0.721} & \textbf{0.707} \\
\hline
\end{tabular}}}}
\vspace{1pt}
\caption{\small{Comparative results on VOC 2012 test set.}}
\label{tab:final_system_voc2012_test}
\vspace{-0.6em} 
\end{table*}
\begin{table*}[t!]
\centering
\renewcommand{\figurename}{Table}
\renewcommand{\captionlabelfont}{\bf}
\renewcommand{\captionfont}{\small} 
\resizebox{\textwidth}{!}{
{\setlength{\extrarowheight}{2pt}\scriptsize
{\begin{tabular}{l <{\hspace{-0.3em}}|>{\hspace{-0.5em}} l | >{\hspace{-0.5em}}c >{\hspace{-1em}}c >{\hspace{-1em}}c >{\hspace{-1em}}c >{\hspace{-1em}}c >{\hspace{-1em}}c >{\hspace{-1em}}c >{\hspace{-1em}}c >{\hspace{-1em}}c >{\hspace{-1em}}c >{\hspace{-1em}}c >{\hspace{-1em}}c >{\hspace{-1em}}c >{\hspace{-1em}}c >{\hspace{-1em}}c >{\hspace{-1em}}c >{\hspace{-1em}}c >{\hspace{-1em}}c >{\hspace{-1em}}c >{\hspace{-1em}}c <{\hspace{-0.3em}}| >{\hspace{-0.3em}}c}
\hline
Approach & trained on &  areo & bike & bird & boat & bottle & bus & car & cat & chair & cow & table & dog & horse & mbike & person & plant & sheep & sofa & train & tv & mAP \\
\hline
\emph{MR-CNN \& S-CNN \& Loc. (Ours)} & VOC07+12 & \textbf{0.803} & \textbf{0.841} & \textbf{0.785} & \textbf{0.708} & \textbf{0.685} & \textbf{0.880} & \textbf{0.859} & \textbf{0.878} & \textbf{0.603} & \textbf{0.852} & \textbf{0.737} & \textbf{0.872} & \textbf{0.865} & \textbf{0.850} & 0.764 & \textbf{0.485} & \textbf{0.763} & 0.755 & \textbf{0.850} & \textbf{0.810} & \textbf{0.782}\\
\emph{MR-CNN \& S-CNN \& Loc. (Ours)} & VOC07 & 0.787 & 0.818 & 0.767 & 0.666 & 0.618 & 0.817 & 0.853 & 0.827 & 0.570 & 0.819 & 0.732 & 0.846 & 0.860 & 0.805 & 0.749 & 0.449 & 0.717 & 0.697 & 0.787 & 0.799 & 0.749\\
\hline
\emph{Faster R-CNN~\cite{shaoqing2015faster}} & VOC07+12 & 0.765 & 0.790 & 0.709 & 0.655 & 0.521 &  0.831 & 0.847 & 0.864 & 0.520 & 0.819 & 0.657 & 0.848 & 0.846 & 0.775 & \textbf{0.767} & 0.388 & 0.736 & 0.739 & 0.830 & 0.726 & 0.732\\
\emph{NoC~\cite{ren2015object}} & VOC07+12  & 0.763 & 0.814 & 0.744 & 0.617 & 0.608 & 0.847 & 0.782 & 0.829 & 0.530 & 0.792 & 0.692 & 0.832 & 0.832 & 0.785 & 0.680 & 0.450 & 0.716 & \textbf{0.767} & 0.822 & 0.757 & 0.733\\
\emph{Fast R-CNN~\cite{girshick2015fast}} & VOC07+12 & 0.770 & 0.781 & 0.693 & 0.594 & 0.383 & 0.816 & 0.786 & 0.867 & 0.428 & 0.788 & 0.689 & 0.847 & 0.820 & 0.766 & 0.699 & 0.318 & 0.701 & 0.748 & 0.804 & 0.704 & 0.700\\
\hline
\end{tabular}}}}
\vspace{1pt}
\caption{\small{Comparative results on VOC 2007 test set for models trained with extra data.}}
\label{tab:final_system_voc2007_test_extra_data}
\vspace{-0.6em} 
\end{table*}
\begin{table*}[t!]
\centering
\renewcommand{\figurename}{Table}
\renewcommand{\captionlabelfont}{\bf}
\renewcommand{\captionfont}{\small} 
\resizebox{\textwidth}{!}{
{\setlength{\extrarowheight}{2pt}\scriptsize
{\begin{tabular}{l <{\hspace{-0.3em}}|>{\hspace{-0.5em}} l | >{\hspace{-0.5em}}c >{\hspace{-1em}}c >{\hspace{-1em}}c >{\hspace{-1em}}c >{\hspace{-1em}}c >{\hspace{-1em}}c >{\hspace{-1em}}c >{\hspace{-1em}}c >{\hspace{-1em}}c >{\hspace{-1em}}c >{\hspace{-1em}}c >{\hspace{-1em}}c >{\hspace{-1em}}c >{\hspace{-1em}}c >{\hspace{-1em}}c >{\hspace{-1em}}c >{\hspace{-1em}}c >{\hspace{-1em}}c >{\hspace{-1em}}c >{\hspace{-1em}}c <{\hspace{-0.3em}}| >{\hspace{-0.3em}}c}
\hline
Approach & trained on &  areo & bike & bird & boat & bottle & bus & car & cat & chair & cow & table & dog & horse & mbike & person & plant & sheep & sofa & train & tv & mAP \\
\hline
\emph{MR-CNN \& S-CNN \& Loc. (Ours)} & VOC07+12 &  \textbf{0.855} & \textbf{0.829} & \textbf{0.766} & \textbf{0.578} & \textbf{0.627} & \textbf{0.794} & \textbf{0.772} & 0.866 & \textbf{0.550} & \textbf{0.791} & \textbf{0.622} & 0.870 & \textbf{0.834} & \textbf{0.847} & 0.789 & 0.453 & \textbf{0.734} & 0.658 & 0.803 & \textbf{0.740} & \textbf{0.739} \\
\emph{MR-CNN \& S-CNN \& Loc. (Ours)} & VOC12 & 0.850 & 0.796 & 0.715 & 0.553 & 0.577 & 0.760 & 0.739 & 0.846 & 0.505 & 0.743 & 0.617 & 0.855 & 0.799 & 0.817 & 0.764 & 0.410 & 0.690 & 0.612 & 0.777 & 0.721 & 0.707 \\
\hline
\emph{Faster R-CNN~\cite{shaoqing2015faster}} & VOC07+12 & 0.849 & 0.798 & 0.743 & 0.539 & 0.498 & 0.775 & 0.759 & 0.885 & 0.456 & 0.771 & 0.553 & 0.869 & 0.817 & 0.809 & \textbf{0.796} & 0.401 & 0.726 & 0.609 & 0.812 & 0.615 & 0.704\\
\emph{Fast R-CNN \& YOLO~\cite{redmon2015yolo}} & VOC07+12 & 0.830 & 0.785 & 0.737 & 0.558 & 0.431 & 0.783 & 0.730 & 0.892 & 0.491 & 0.743 & 0.566 & 0.872 & 0.805 & 0.805 & 0.747 & 0.421 & 0.708 & 0.683 & \textbf{0.815} & 0.670 & 0.704\\
\emph{Deep Ensemble COCO~\cite{GuoG15a}} & VOC07+12, COCO~\cite{lin2014microsoft} & 0.840 & 0.794 & 0.716 & 0.519 & 0.511 & 0.741 & 0.721 & 0.886 & 0.483 & 0.734 & 0.578 & 0.861 & 0.800 & 0.807 & 0.704 & \textbf{0.466} & 0.696 & \textbf{0.688} & 0.759 & 0.714 & 0.701\\
\emph{NoC~\cite{ren2015object}} & VOC07+12 & 0.828 & 0.790 & 0.716 & 0.523 & 0.537 & 0.741 & 0.690 & 0.849 &  0.469 & 0.743 & 0.531 & 0.850 & 0.813 & 0.795 & 0.722 & 0.389 & 0.724 & 0.595 & 0.767 & 0.681 & 0.688\\
\emph{Fast R-CNN~\cite{girshick2015fast}} & VOC07+12 & 0.823 & 0.784 & 0.708 & 0.523 & 0.387 & 0.778 & 0.716 & \textbf{0.893} & 0.442 & 0.730 & 0.550 & \textbf{0.875} & 0.805 & 0.808 & 0.720 & 0.351        & 0.683 & 0.657 &       0.804 & 0.642 & 0.684 \\
\hline
\end{tabular}}}}
\vspace{1pt}
\caption{\small{Comparative results on VOC 2012 test set for models trained with extra data.}}
\label{tab:final_system_voc2012_test_extra_data}
\vspace{-0.6em} 
\end{table*}

In table~\ref{tab:final_system_voc2012_test}, we compare our detection system against other published work on the test set of PASCAL VOC2012~\cite{everingham2012pascal}. 
Our overall system involves the Multi-Region CNN model enriched with the semantic segmentation aware CNN features and coupled with the CNN based bounding box regression under the iterative localization scheme. 
We tested two instances of our system. Both of them have exactly the same components but they have being trained on different datasets. 
For the first one, the fine-tuning of the networks as well as the training of the detection SVMs was performed on VOC2007 train+val dataset that includes $5011$ annotated images.
For the second one, the fine-tuning of the networks was performed on VOC2012 train dataset that includes $5717$ annotated images and the training of the detection SVMs was performed on VOC2012 train+val dataset that includes $11540$ annotated images.
As we observe from table~\ref{tab:final_system_voc2012_test},
we achieve excellent mAP ($69.1\%$ and $70.7\%$ correspondingly) in both cases setting the new state-of-the-art on this test set and for those training sets.

\subsection{Training with extra data and comparison with contemporary work}
Approaches contemporary to ours~\cite{ren2015object,girshick2015fast,shaoqing2015faster,redmon2015yolo}, train their models with extra data in order to improve the accuracy of their systems. 
We follow the same practice and we report results on tables~\ref{tab:final_system_voc2007_test_extra_data} and~\ref{tab:final_system_voc2012_test_extra_data}. Specifically, we trained our models on VOC 2007 and 2012 train+val datasets using both selective search~\cite{van2011segmentation} and EdgeBox~\cite{zitnick2014edge} proposals. During test time we only use EdgeBox proposals that are faster to be computed. 
From the tables, it is apparent that our methods outperforms the other approaches even when trained with less data. 
Currently (08/06/15), our entries are ranked 1st and 2nd on the leader board of PASCAL VOC2012 object detection comp4 benchmark (see table~\ref{tab:final_system_voc2012_test_extra_data}) and the difference of our top performing entry from the 3rd is $3.5$ points.

\section{Qualitative Results} \label{sec:qualitative_results}

\begin{figure*}[t!]
\center
\renewcommand{\figurename}{Figure}
\renewcommand{\captionlabelfont}{\bf}
\renewcommand{\captionfont}{\small} 
\vspace{5pt}
\begin{subfigure}[b]{\textwidth}
\center
\renewcommand{\figurename}{Figure}
\renewcommand{\captionlabelfont}{\bf}
\renewcommand{\captionfont}{\footnotesize} 
        \begin{center}
        \begin{subfigure}[b]{0.18\textwidth}
        \includegraphics[width=\textwidth]{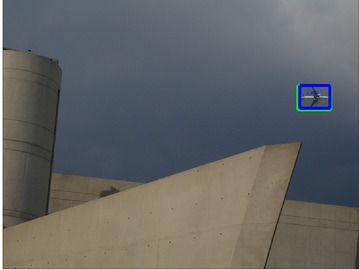}
        \end{subfigure} 
        \hspace{0.05cm} 
        \begin{subfigure}[b]{0.18\textwidth}
        \includegraphics[width=\textwidth]{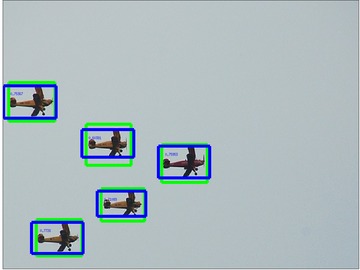}
        \end{subfigure} 
        \hspace{0.05cm} 
        \begin{subfigure}[b]{0.18\textwidth}
        \includegraphics[width=\textwidth]{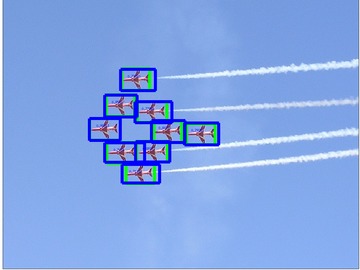}
        \end{subfigure} 
        \hspace{0.05cm} 
        \begin{subfigure}[b]{0.18\textwidth}
        \includegraphics[width=\textwidth]{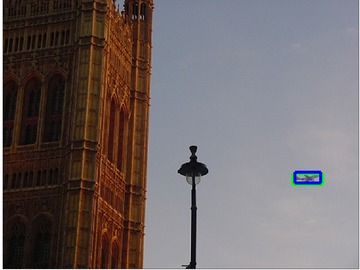}
        \end{subfigure} 
        \hspace{0.05cm} 
        \begin{subfigure}[b]{0.18\textwidth}
        \includegraphics[width=\textwidth]{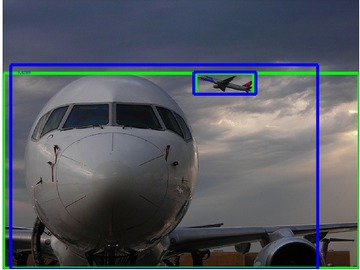}
        \end{subfigure} 
        \end{center}
        \caption{Aeroplane detections.}
        \label{fig:aeroplane}
        \vspace{5pt}
\end{subfigure}  
\begin{subfigure}[b]{\textwidth}
\center
\renewcommand{\figurename}{Figure}
\renewcommand{\captionlabelfont}{\bf}
\renewcommand{\captionfont}{\footnotesize} 
        \begin{center}
        \begin{subfigure}[b]{0.18\textwidth}
        \includegraphics[width=\textwidth]{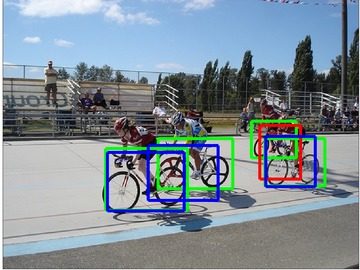}
        \end{subfigure} 
        \hspace{0.05cm} 
        \begin{subfigure}[b]{0.18\textwidth}
        \includegraphics[width=\textwidth]{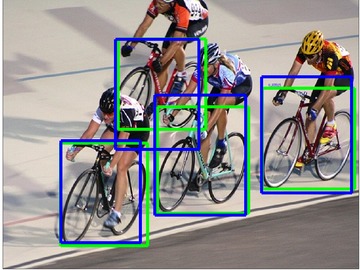}
        \end{subfigure} 
        \hspace{0.05cm} 
        \begin{subfigure}[b]{0.18\textwidth}
        \includegraphics[width=\textwidth]{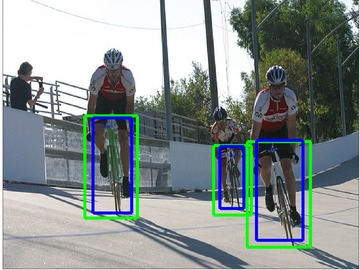}
        \end{subfigure} 
        \hspace{0.05cm} 
        \begin{subfigure}[b]{0.18\textwidth}
        \includegraphics[width=\textwidth]{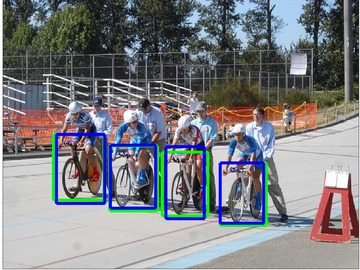}
        \end{subfigure} 
        \hspace{0.05cm} 
        \begin{subfigure}[b]{0.18\textwidth}
        \includegraphics[width=\textwidth]{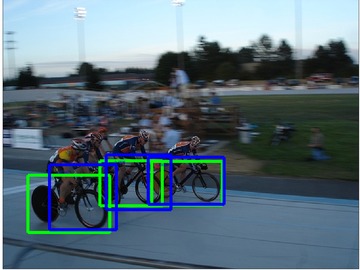}
        \end{subfigure} 
        \end{center}
        \caption{Bicycle detections.}
        \label{fig:bicycle}   
        \vspace{5pt}    
\end{subfigure}  
\begin{subfigure}[b]{\textwidth}
\center
\renewcommand{\figurename}{Figure}
\renewcommand{\captionlabelfont}{\bf}
\renewcommand{\captionfont}{\footnotesize} 
        \begin{center}
        \begin{subfigure}[b]{0.18\textwidth}
        \includegraphics[width=\textwidth]{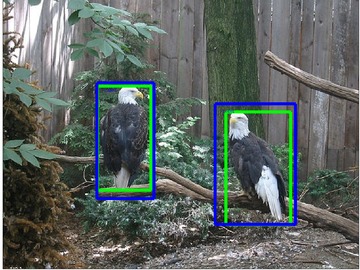}
        \end{subfigure} 
        \hspace{0.05cm} 
        \begin{subfigure}[b]{0.18\textwidth}
        \includegraphics[width=\textwidth]{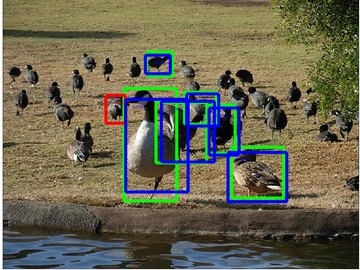}
        \end{subfigure} 
        \hspace{0.05cm} 
        \begin{subfigure}[b]{0.18\textwidth}
        \includegraphics[width=\textwidth]{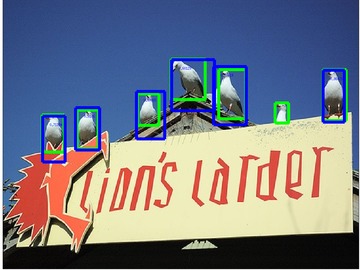}
        \end{subfigure} 
        \hspace{0.05cm} 
        \begin{subfigure}[b]{0.18\textwidth}
        \includegraphics[width=\textwidth]{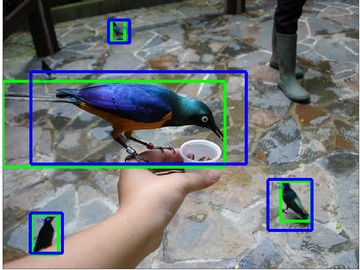}
        \end{subfigure} 
        \hspace{0.05cm} 
        \begin{subfigure}[b]{0.18\textwidth}
        \includegraphics[width=\textwidth]{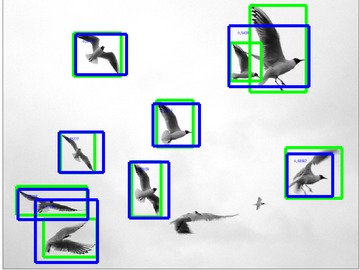}
        \end{subfigure} 
        \end{center}
        \caption{Bird detections.}
        \label{fig:bird}
                \vspace{5pt}    
\end{subfigure}
\begin{subfigure}[b]{\textwidth}
\center
\renewcommand{\figurename}{Figure}
\renewcommand{\captionlabelfont}{\bf}
\renewcommand{\captionfont}{\footnotesize} 
        \begin{center}
        \begin{subfigure}[b]{0.18\textwidth}
        \includegraphics[width=\textwidth]{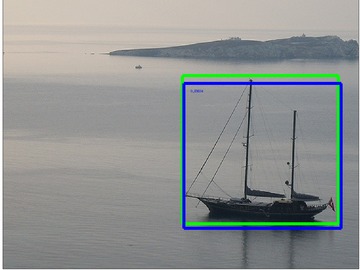}
        \end{subfigure} 
        \hspace{0.05cm} 
        \begin{subfigure}[b]{0.18\textwidth}
        \includegraphics[width=\textwidth]{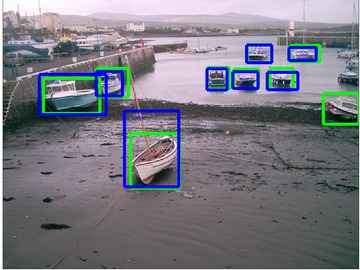}
        \end{subfigure} 
        \hspace{0.05cm} 
        \begin{subfigure}[b]{0.18\textwidth}
        \includegraphics[width=\textwidth]{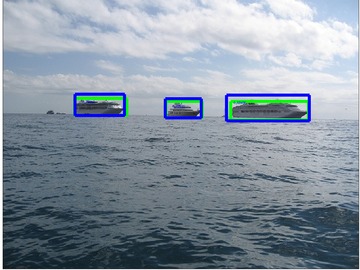}
        \end{subfigure} 
        \hspace{0.05cm} 
        \begin{subfigure}[b]{0.18\textwidth}
        \includegraphics[width=\textwidth]{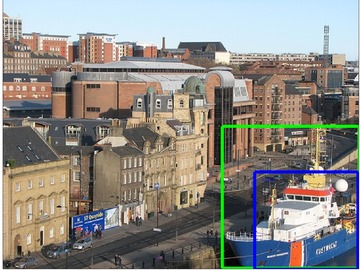}
        \end{subfigure} 
        \hspace{0.05cm} 
        \begin{subfigure}[b]{0.18\textwidth}
        \includegraphics[width=\textwidth]{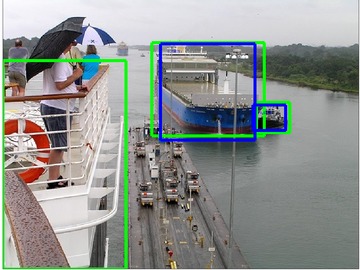}
        \end{subfigure} 
        \end{center}
        \caption{Boat detections.}
        \label{fig:boat}
                \vspace{5pt}
\end{subfigure}  
\begin{subfigure}[b]{\textwidth}
\center
\renewcommand{\figurename}{Figure}
\renewcommand{\captionlabelfont}{\bf}
\renewcommand{\captionfont}{\footnotesize} 
        \begin{center}
        \begin{subfigure}[b]{0.18\textwidth}
        \includegraphics[width=\textwidth]{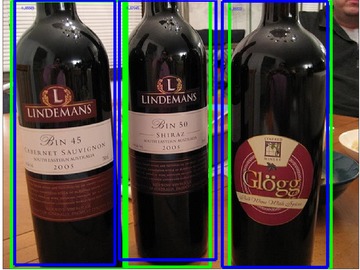}
        \end{subfigure} 
        \hspace{0.05cm} 
        \begin{subfigure}[b]{0.18\textwidth}
        \includegraphics[width=\textwidth]{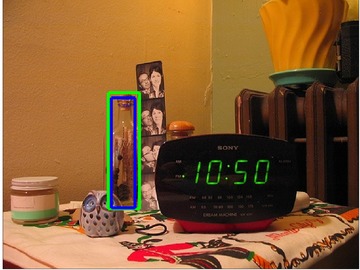}
        \end{subfigure} 
        \hspace{0.05cm} 
        \begin{subfigure}[b]{0.18\textwidth}
        \includegraphics[width=\textwidth]{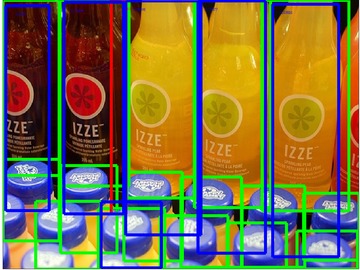}
        \end{subfigure} 
        \hspace{0.05cm} 
        \begin{subfigure}[b]{0.18\textwidth}
        \includegraphics[width=\textwidth]{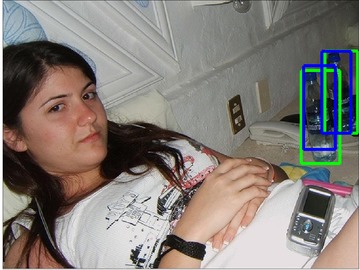}
        \end{subfigure} 
        \hspace{0.05cm} 
        \begin{subfigure}[b]{0.18\textwidth}
        \includegraphics[width=\textwidth]{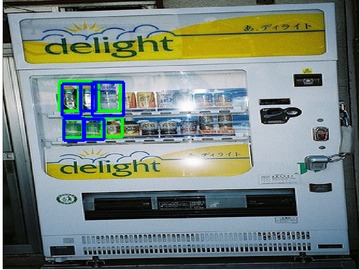}
        \end{subfigure} 
        \end{center}
        \caption{Bottle detections.}
        \label{fig:bottle}
                \vspace{5pt}        
\end{subfigure}  
\begin{subfigure}[b]{\textwidth}
\center
\renewcommand{\figurename}{Figure}
\renewcommand{\captionlabelfont}{\bf}
\renewcommand{\captionfont}{\footnotesize} 
        \begin{center}
        \begin{subfigure}[b]{0.18\textwidth}
        \includegraphics[width=\textwidth]{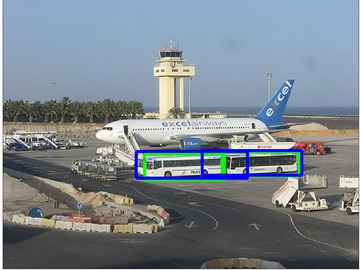}
        \end{subfigure} 
        \hspace{0.05cm} 
        \begin{subfigure}[b]{0.18\textwidth}
        \includegraphics[width=\textwidth]{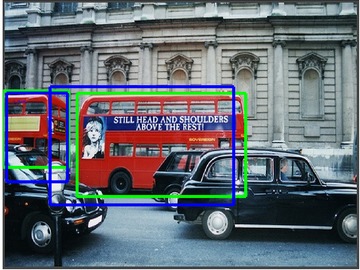}
        \end{subfigure} 
        \hspace{0.05cm} 
        \begin{subfigure}[b]{0.18\textwidth}
        \includegraphics[width=\textwidth]{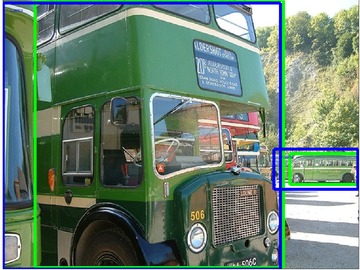}
        \end{subfigure} 
        \hspace{0.05cm} 
        \begin{subfigure}[b]{0.18\textwidth}
        \includegraphics[width=\textwidth]{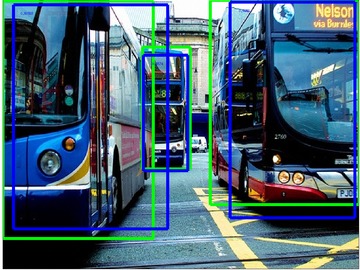}
        \end{subfigure} 
        \hspace{0.05cm} 
        \begin{subfigure}[b]{0.18\textwidth}
        \includegraphics[width=\textwidth]{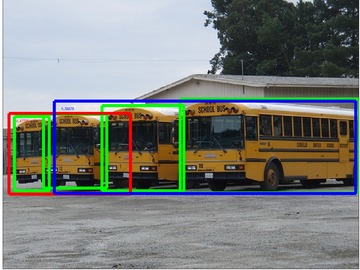}
        \end{subfigure} 
        \end{center}
        \caption{Bus detections.}
        \label{fig:bus}
                \vspace{5pt}        
\end{subfigure} 
\begin{subfigure}[b]{\textwidth}
\center
\renewcommand{\figurename}{Figure}
\renewcommand{\captionlabelfont}{\bf}
\renewcommand{\captionfont}{\footnotesize} 
        \begin{center}
        \begin{subfigure}[b]{0.18\textwidth}
        \includegraphics[width=\textwidth]{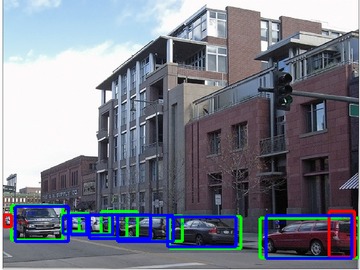}
        \end{subfigure} 
        \hspace{0.05cm} 
        \begin{subfigure}[b]{0.18\textwidth}
        \includegraphics[width=\textwidth]{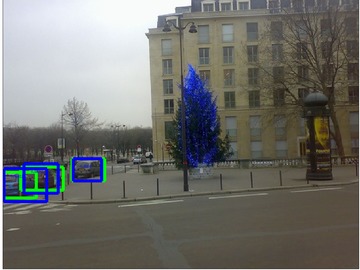}
        \end{subfigure} 
        \hspace{0.05cm} 
        \begin{subfigure}[b]{0.18\textwidth}
        \includegraphics[width=\textwidth]{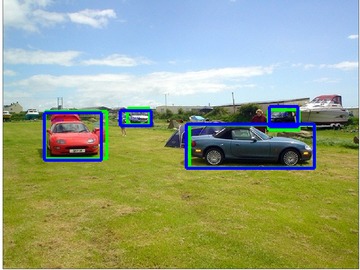}
        \end{subfigure} 
        \hspace{0.05cm} 
        \begin{subfigure}[b]{0.18\textwidth}
        \includegraphics[width=\textwidth]{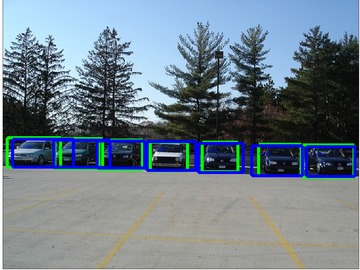}
        \end{subfigure} 
        \hspace{0.05cm} 
        \begin{subfigure}[b]{0.18\textwidth}
        \includegraphics[width=\textwidth]{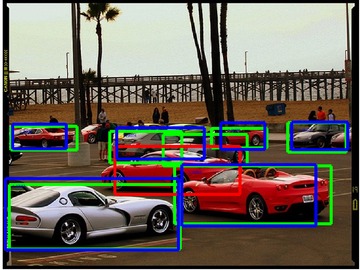}
        \end{subfigure} 
        \end{center}
        \caption{Car detections.}
        \label{fig:car}
\end{subfigure} 
\vspace{5pt}
\caption{We use blue bounding boxes for the true positive detections and red bounding boxes (if any) for the false positive detections. The ground truth bounding boxes are drawn with green color.}
        \label{fig:Det1}
\end{figure*}

\begin{figure*}[t!]
\center
\renewcommand{\figurename}{Figure}
\renewcommand{\captionlabelfont}{\bf}
\renewcommand{\captionfont}{\small} 
\begin{subfigure}[b]{\textwidth}
\center
\renewcommand{\figurename}{Figure}
\renewcommand{\captionlabelfont}{\bf}
\renewcommand{\captionfont}{\footnotesize} 
        \begin{center}
        \begin{subfigure}[b]{0.18\textwidth}
        \includegraphics[width=\textwidth]{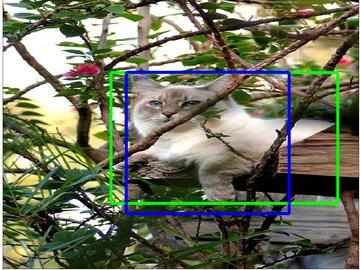}
        \end{subfigure} 
        \hspace{0.05cm} 
        \begin{subfigure}[b]{0.18\textwidth}
        \includegraphics[width=\textwidth]{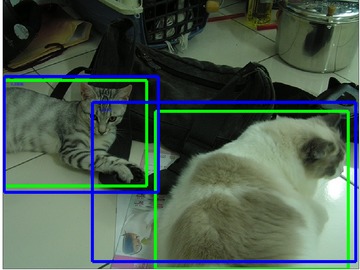}
        \end{subfigure} 
        \hspace{0.05cm} 
        \begin{subfigure}[b]{0.18\textwidth}
        \includegraphics[width=\textwidth]{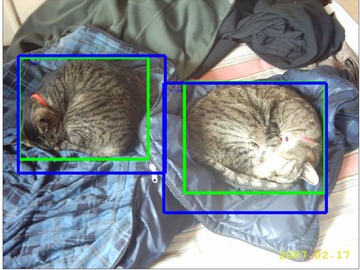}
        \end{subfigure} 
        \hspace{0.05cm} 
        \begin{subfigure}[b]{0.18\textwidth}
        \includegraphics[width=\textwidth]{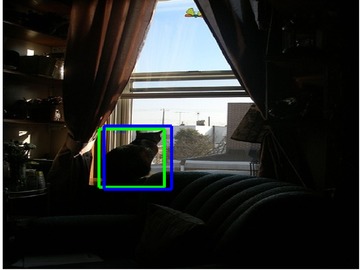}
        \end{subfigure} 
        \hspace{0.05cm} 
        \begin{subfigure}[b]{0.18\textwidth}
        \includegraphics[width=\textwidth]{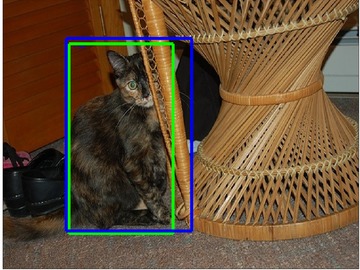}
        \end{subfigure} 
        \end{center}
        \caption{Cat detections.}
        \label{fig:cat}
        \vspace{5pt}
\end{subfigure}   
\begin{subfigure}[b]{\textwidth}
\center
\renewcommand{\figurename}{Figure}
\renewcommand{\captionlabelfont}{\bf}
\renewcommand{\captionfont}{\footnotesize} 
        \begin{center}
        \begin{subfigure}[b]{0.18\textwidth}
        \includegraphics[width=\textwidth]{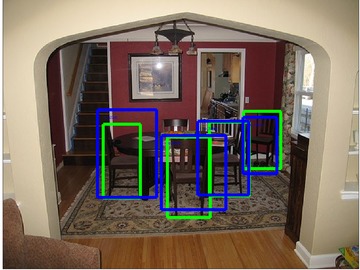}
        \end{subfigure} 
        \hspace{0.05cm} 
        \begin{subfigure}[b]{0.18\textwidth}
        \includegraphics[width=\textwidth]{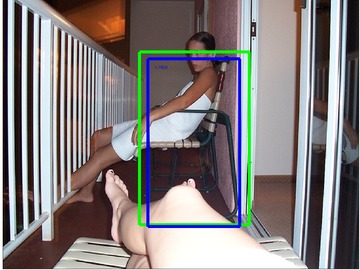}
        \end{subfigure} 
        \hspace{0.05cm} 
        \begin{subfigure}[b]{0.18\textwidth}
        \includegraphics[width=\textwidth]{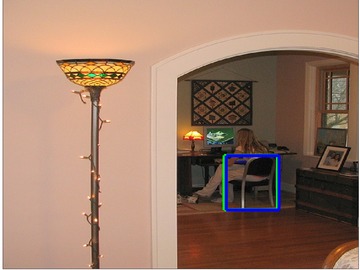}
        \end{subfigure} 
        \hspace{0.05cm} 
        \begin{subfigure}[b]{0.18\textwidth}
        \includegraphics[width=\textwidth]{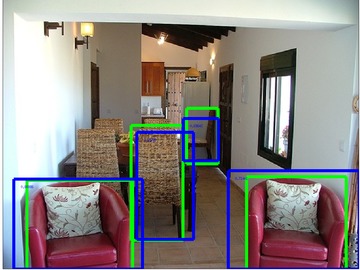}
        \end{subfigure} 
        \hspace{0.05cm} 
        \begin{subfigure}[b]{0.18\textwidth}
        \includegraphics[width=\textwidth]{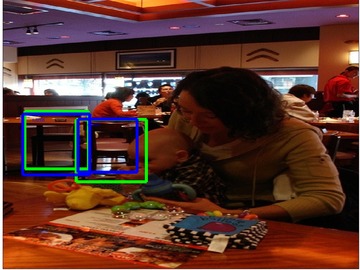}
        \end{subfigure} 
        \end{center}
        \caption{Chair detections.}
        \label{fig:chair}
        \vspace{5pt}
\end{subfigure}   
\begin{subfigure}[b]{\textwidth}
\center
\renewcommand{\figurename}{Figure}
\renewcommand{\captionlabelfont}{\bf}
\renewcommand{\captionfont}{\footnotesize} 
        \begin{center}
        \begin{subfigure}[b]{0.18\textwidth}
        \includegraphics[width=\textwidth]{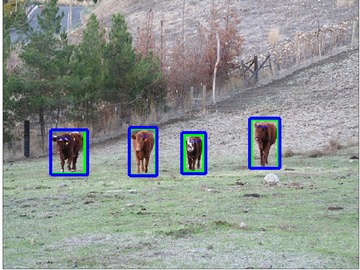}
        \end{subfigure} 
        \hspace{0.05cm} 
        \begin{subfigure}[b]{0.18\textwidth}
        \includegraphics[width=\textwidth]{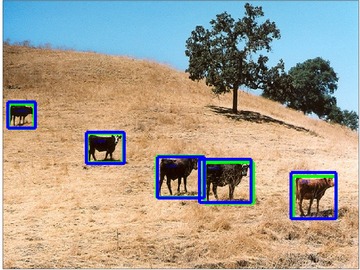}
        \end{subfigure} 
        \hspace{0.05cm} 
        \begin{subfigure}[b]{0.18\textwidth}
        \includegraphics[width=\textwidth]{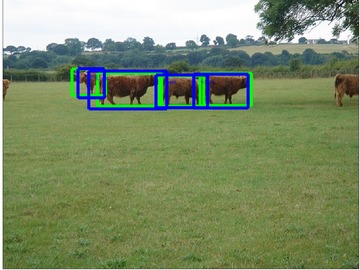}
        \end{subfigure} 
        \hspace{0.05cm} 
        \begin{subfigure}[b]{0.18\textwidth}
        \includegraphics[width=\textwidth]{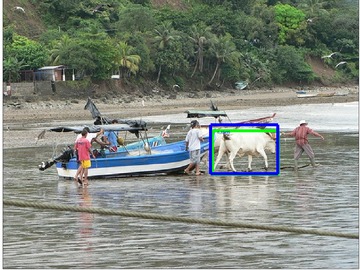}
        \end{subfigure} 
        \hspace{0.05cm} 
        \begin{subfigure}[b]{0.18\textwidth}
        \includegraphics[width=\textwidth]{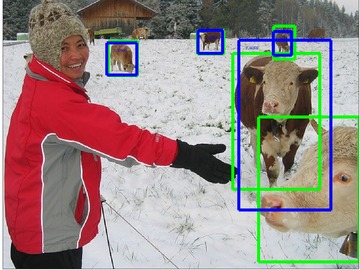}
        \end{subfigure} 
        \end{center}
        \caption{Cow detections.}
        \label{fig:cow}
        \vspace{5pt}
\end{subfigure}   
\begin{subfigure}[b]{\textwidth}
\center
\renewcommand{\figurename}{Figure}
\renewcommand{\captionlabelfont}{\bf}
\renewcommand{\captionfont}{\footnotesize} 
        \begin{center}
        \begin{subfigure}[b]{0.18\textwidth}
        \includegraphics[width=\textwidth]{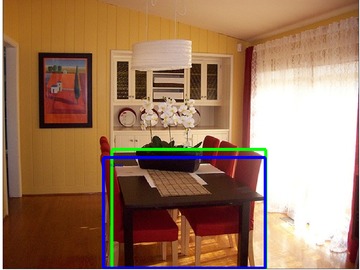}
        \end{subfigure} 
        \hspace{0.05cm} 
        \begin{subfigure}[b]{0.18\textwidth}
        \includegraphics[width=\textwidth]{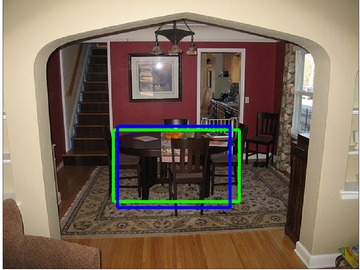}
        \end{subfigure} 
        \hspace{0.05cm} 
        \begin{subfigure}[b]{0.18\textwidth}
        \includegraphics[width=\textwidth]{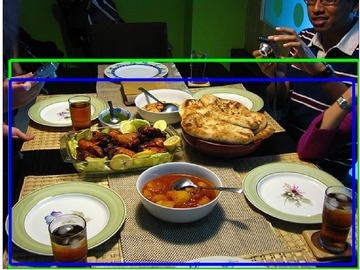}
        \end{subfigure} 
        \hspace{0.05cm} 
        \begin{subfigure}[b]{0.18\textwidth}
        \includegraphics[width=\textwidth]{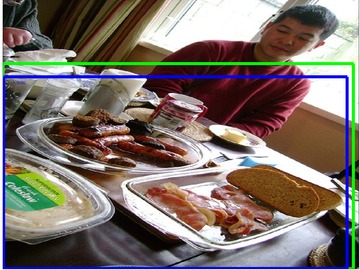}
        \end{subfigure} 
        \hspace{0.05cm} 
        \begin{subfigure}[b]{0.18\textwidth}
        \includegraphics[width=\textwidth]{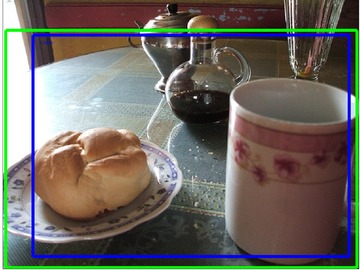}
        \end{subfigure} 
        \end{center}
        \caption{Dinningtable detections.}
        \label{fig:dinningtable}
        \vspace{5pt}
\end{subfigure} 
\begin{subfigure}[b]{\textwidth}
\center
\renewcommand{\figurename}{Figure}
\renewcommand{\captionlabelfont}{\bf}
\renewcommand{\captionfont}{\footnotesize} 
        \begin{center}
        \begin{subfigure}[b]{0.18\textwidth}
        \includegraphics[width=\textwidth]{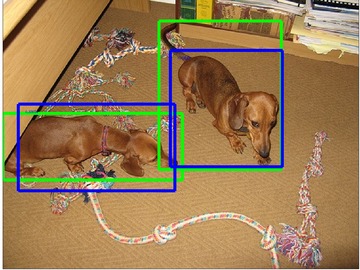}
        \end{subfigure} 
        \hspace{0.05cm} 
        \begin{subfigure}[b]{0.18\textwidth}
        \includegraphics[width=\textwidth]{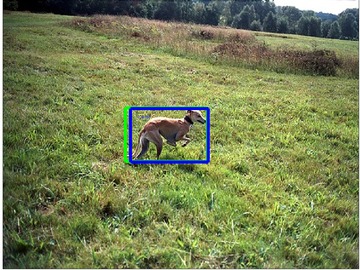}
        \end{subfigure} 
        \hspace{0.05cm} 
        \begin{subfigure}[b]{0.18\textwidth}
        \includegraphics[width=\textwidth]{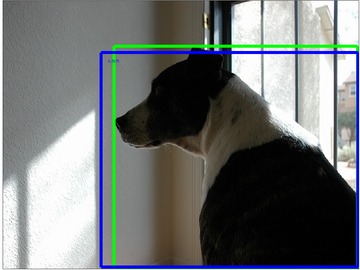}
        \end{subfigure} 
        \hspace{0.05cm} 
        \begin{subfigure}[b]{0.18\textwidth}
        \includegraphics[width=\textwidth]{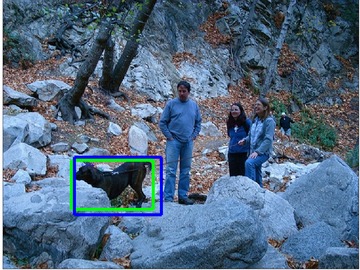}
        \end{subfigure} 
        \hspace{0.05cm} 
        \begin{subfigure}[b]{0.18\textwidth}
        \includegraphics[width=\textwidth]{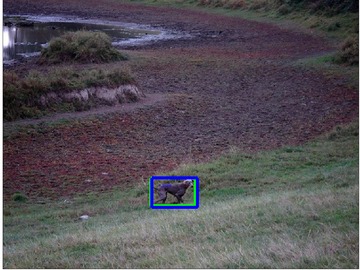}
        \end{subfigure} 
        \end{center}
        \caption{Dog detections.}
        \label{fig:dog}
        \vspace{5pt}
\end{subfigure} 
\begin{subfigure}[b]{\textwidth}
\center
\renewcommand{\figurename}{Figure}
\renewcommand{\captionlabelfont}{\bf}
\renewcommand{\captionfont}{\footnotesize} 
        \begin{center}
        \begin{subfigure}[b]{0.18\textwidth}
        \includegraphics[width=\textwidth]{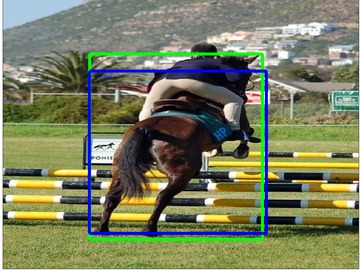}
        \end{subfigure} 
        \hspace{0.05cm} 
        \begin{subfigure}[b]{0.18\textwidth}
        \includegraphics[width=\textwidth]{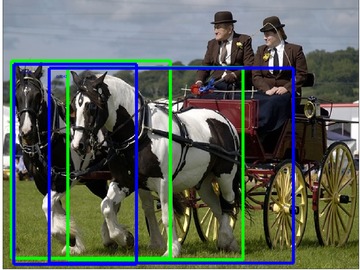}
        \end{subfigure} 
        \hspace{0.05cm} 
        \begin{subfigure}[b]{0.18\textwidth}
        \includegraphics[width=\textwidth]{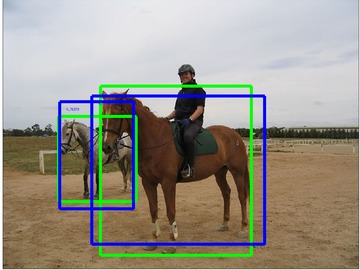}
        \end{subfigure} 
        \hspace{0.05cm} 
        \begin{subfigure}[b]{0.18\textwidth}
        \includegraphics[width=\textwidth]{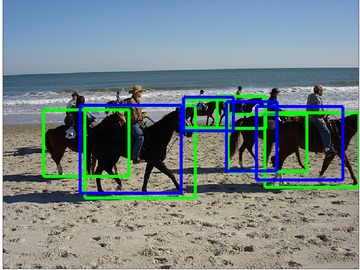}
        \end{subfigure} 
        \hspace{0.05cm} 
        \begin{subfigure}[b]{0.18\textwidth}
        \includegraphics[width=\textwidth]{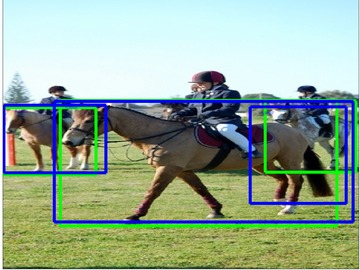}
        \end{subfigure} 
        \end{center}
        \caption{Horse detections.}
        \label{fig:horse}
        \vspace{5pt}
\end{subfigure} 
\begin{subfigure}[b]{\textwidth}
\center
\renewcommand{\figurename}{Figure}
\renewcommand{\captionlabelfont}{\bf}
\renewcommand{\captionfont}{\footnotesize} 
        \begin{center}
        \begin{subfigure}[b]{0.18\textwidth}
        \includegraphics[width=\textwidth]{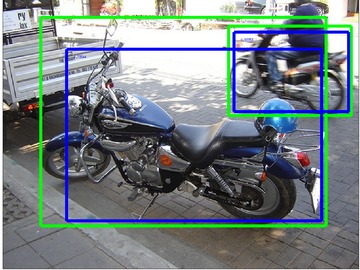}
        \end{subfigure} 
        \hspace{0.05cm} 
        \begin{subfigure}[b]{0.18\textwidth}
        \includegraphics[width=\textwidth]{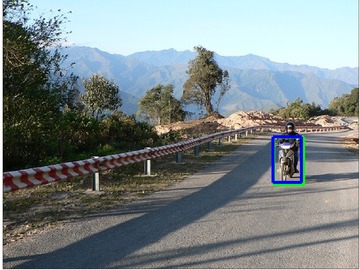}
        \end{subfigure} 
        \hspace{0.05cm} 
        \begin{subfigure}[b]{0.18\textwidth}
        \includegraphics[width=\textwidth]{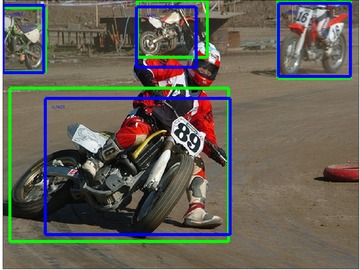}
        \end{subfigure} 
        \hspace{0.05cm} 
        \begin{subfigure}[b]{0.18\textwidth}
        \includegraphics[width=\textwidth]{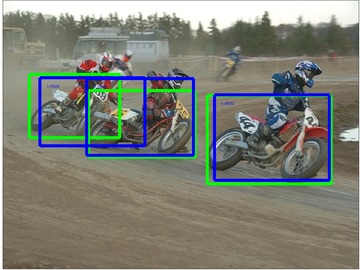}
        \end{subfigure} 
        \hspace{0.05cm} 
        \begin{subfigure}[b]{0.18\textwidth}
        \includegraphics[width=\textwidth]{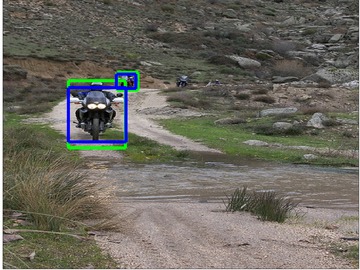}
        \end{subfigure} 
        \end{center}
        \caption{Motorbike detections.}
\end{subfigure} 
\vspace{5pt}
\caption{We use blue bounding boxes for the true positive detections and red bounding boxes (if any) for the false positive detections. The ground truth bounding boxes are drawn with green color.}
        \label{fig:Det2}
\end{figure*}

\begin{figure*}[t!]
\center
\renewcommand{\figurename}{Figure}
\renewcommand{\captionlabelfont}{\bf}
\renewcommand{\captionfont}{\small} 
\begin{subfigure}[b]{\textwidth}
\center
\renewcommand{\figurename}{Figure}
\renewcommand{\captionlabelfont}{\bf}
\renewcommand{\captionfont}{\footnotesize} 
        \begin{center}
        \begin{subfigure}[b]{0.18\textwidth}
        \includegraphics[width=\textwidth]{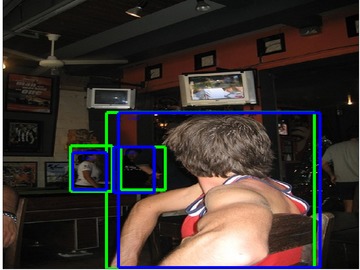}
        \end{subfigure} 
        \hspace{0.05cm} 
        \begin{subfigure}[b]{0.18\textwidth}
        \includegraphics[width=\textwidth]{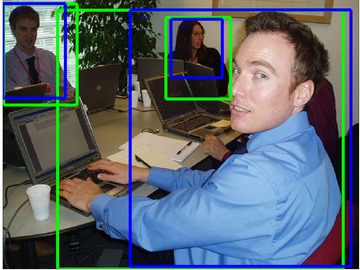}
        \end{subfigure} 
        \hspace{0.05cm} 
        \begin{subfigure}[b]{0.18\textwidth}
        \includegraphics[width=\textwidth]{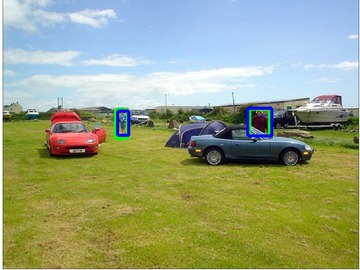}
        \end{subfigure} 
        \hspace{0.05cm} 
        \begin{subfigure}[b]{0.18\textwidth}
        \includegraphics[width=\textwidth]{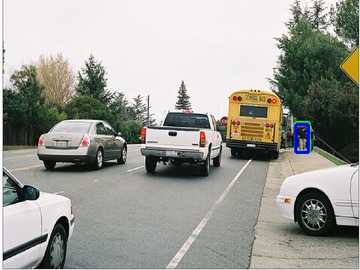}
        \end{subfigure} 
        \hspace{0.05cm} 
        \begin{subfigure}[b]{0.18\textwidth}
        \includegraphics[width=\textwidth]{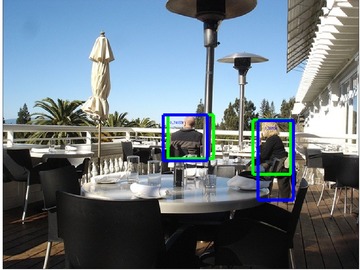}
        \end{subfigure} 
        \end{center}
        \caption{Person detections.}
        \label{fig:person}
        \vspace{5pt}
\end{subfigure}  
\begin{subfigure}[b]{\textwidth}
\center
\renewcommand{\figurename}{Figure}
\renewcommand{\captionlabelfont}{\bf}
\renewcommand{\captionfont}{\footnotesize} 
        \begin{center}
        \begin{subfigure}[b]{0.18\textwidth}
        \includegraphics[width=\textwidth]{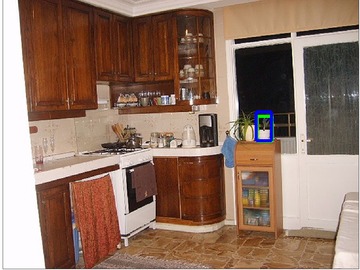}
        \end{subfigure} 
        \hspace{0.05cm} 
        \begin{subfigure}[b]{0.18\textwidth}
        \includegraphics[width=\textwidth]{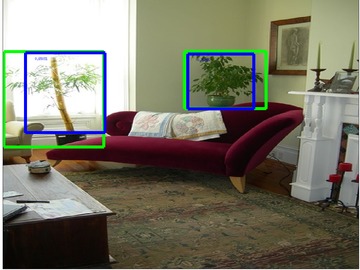}
        \end{subfigure} 
        \hspace{0.05cm} 
        \begin{subfigure}[b]{0.18\textwidth}
        \includegraphics[width=\textwidth]{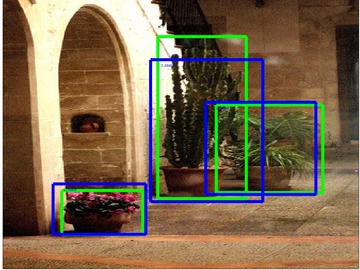}
        \end{subfigure} 
        \hspace{0.05cm} 
        \begin{subfigure}[b]{0.18\textwidth}
        \includegraphics[width=\textwidth]{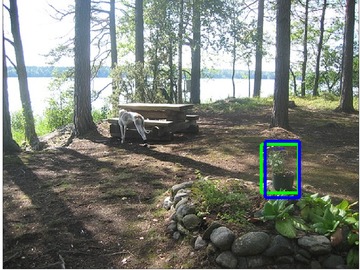}
        \end{subfigure} 
        \hspace{0.05cm} 
        \begin{subfigure}[b]{0.18\textwidth}
        \includegraphics[width=\textwidth]{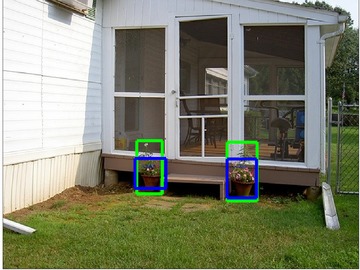}
        \end{subfigure} 
        \end{center}
        \caption{Pottedplant detections.}
        \label{fig:pottedplant}
        \vspace{5pt}
\end{subfigure}       
\begin{subfigure}[b]{\textwidth}
\center
\renewcommand{\figurename}{Figure}
\renewcommand{\captionlabelfont}{\bf}
\renewcommand{\captionfont}{\footnotesize} 
        \begin{center}
        \begin{subfigure}[b]{0.18\textwidth}
        \includegraphics[width=\textwidth]{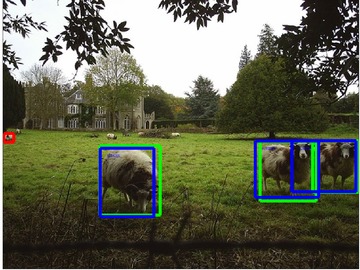}
        \end{subfigure} 
        \hspace{0.05cm} 
        \begin{subfigure}[b]{0.18\textwidth}
        \includegraphics[width=\textwidth]{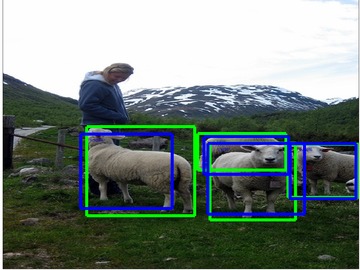}
        \end{subfigure} 
        \hspace{0.05cm} 
        \begin{subfigure}[b]{0.18\textwidth}
        \includegraphics[width=\textwidth]{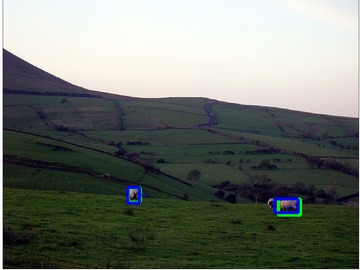}
        \end{subfigure} 
        \hspace{0.05cm} 
        \begin{subfigure}[b]{0.18\textwidth}
        \includegraphics[width=\textwidth]{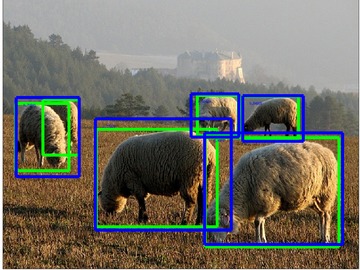}
        \end{subfigure} 
        \hspace{0.05cm} 
        \begin{subfigure}[b]{0.18\textwidth}
        \includegraphics[width=\textwidth]{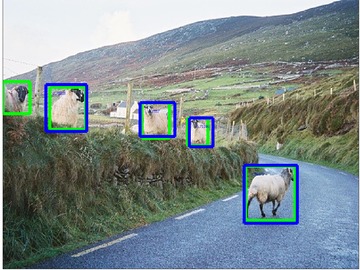}
        \end{subfigure} 
        \end{center}
        \caption{Sheep detection.}
        \label{fig:sheep}
                \vspace{5pt}
\end{subfigure}  
\begin{subfigure}[b]{\textwidth}
\center
\renewcommand{\figurename}{Figure}
\renewcommand{\captionlabelfont}{\bf}
\renewcommand{\captionfont}{\footnotesize} 
        \begin{center}
        \begin{subfigure}[b]{0.18\textwidth}
        \includegraphics[width=\textwidth]{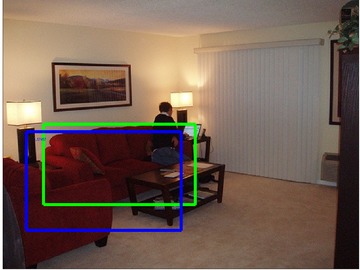}
        \end{subfigure} 
        \hspace{0.05cm} 
        \begin{subfigure}[b]{0.18\textwidth}
        \includegraphics[width=\textwidth]{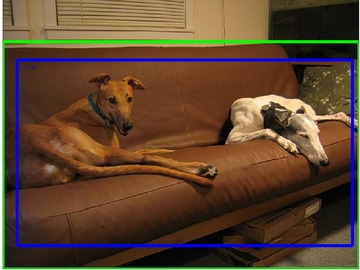}
        \end{subfigure} 
        \hspace{0.05cm} 
        \begin{subfigure}[b]{0.18\textwidth}
        \includegraphics[width=\textwidth]{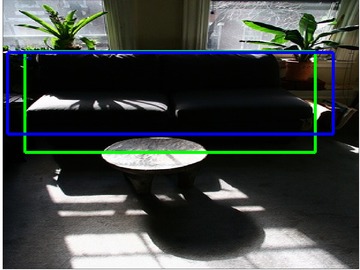}
        \end{subfigure} 
        \hspace{0.05cm} 
        \begin{subfigure}[b]{0.18\textwidth}
        \includegraphics[width=\textwidth]{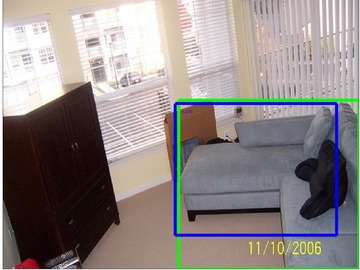}
        \end{subfigure} 
        \hspace{0.05cm} 
        \begin{subfigure}[b]{0.18\textwidth}
        \includegraphics[width=\textwidth]{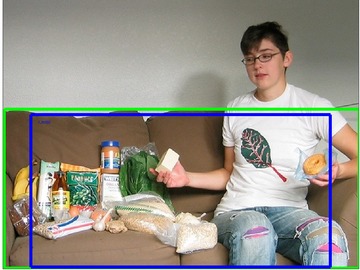}
        \end{subfigure} 
        \end{center}
        \caption{Sofa detections.}
        \label{fig:sofa}
        \vspace{5pt}
\end{subfigure}  
\begin{subfigure}[b]{\textwidth}
\center
\renewcommand{\figurename}{Figure}
\renewcommand{\captionlabelfont}{\bf}
\renewcommand{\captionfont}{\footnotesize} 
        \begin{center}
        \begin{subfigure}[b]{0.18\textwidth}
        \includegraphics[width=\textwidth]{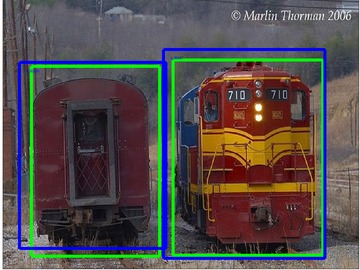}
        \end{subfigure} 
        \hspace{0.05cm} 
        \begin{subfigure}[b]{0.18\textwidth}
        \includegraphics[width=\textwidth]{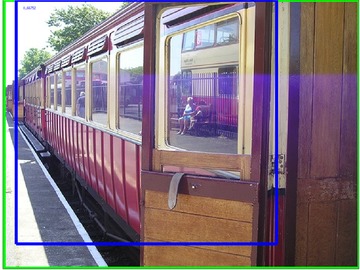}
        \end{subfigure} 
        \hspace{0.05cm} 
        \begin{subfigure}[b]{0.18\textwidth}
        \includegraphics[width=\textwidth]{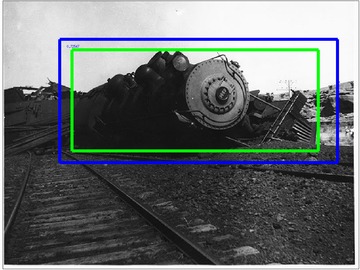}
        \end{subfigure} 
        \hspace{0.05cm} 
        \begin{subfigure}[b]{0.18\textwidth}
        \includegraphics[width=\textwidth]{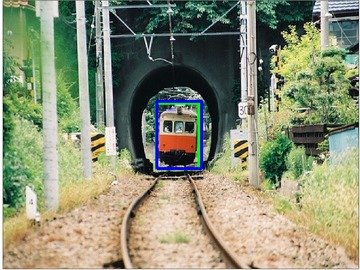}
        \end{subfigure} 
        \hspace{0.05cm} 
        \begin{subfigure}[b]{0.18\textwidth}
        \includegraphics[width=\textwidth]{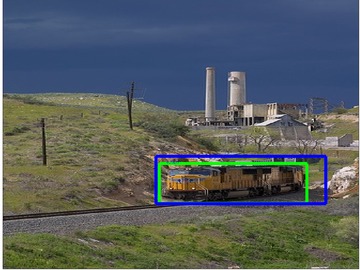}
        \end{subfigure} 
        \end{center}
        \caption{Train detections.}
        \label{fig:train}
        \vspace{5pt}
\end{subfigure}                 
\begin{subfigure}[b]{\textwidth}
\center
\renewcommand{\figurename}{Figure}
\renewcommand{\captionlabelfont}{\bf}
\renewcommand{\captionfont}{\footnotesize} 
        \begin{center}
        \begin{subfigure}[b]{0.18\textwidth}
        \includegraphics[width=\textwidth]{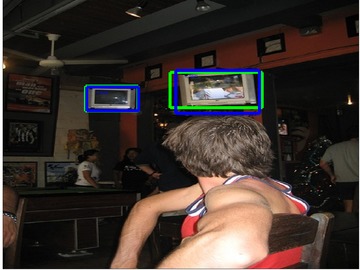}
        \end{subfigure} 
        \hspace{0.05cm} 
        \begin{subfigure}[b]{0.18\textwidth}
        \includegraphics[width=\textwidth]{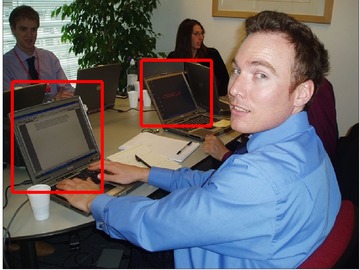}
        \end{subfigure} 
        \hspace{0.05cm} 
        \begin{subfigure}[b]{0.18\textwidth}\
        \includegraphics[width=\textwidth]{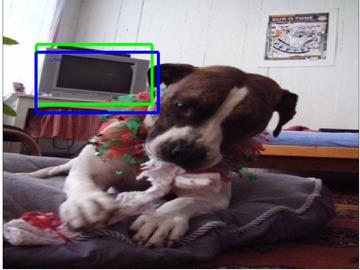}
        \end{subfigure} 
        \hspace{0.05cm} 
        \begin{subfigure}[b]{0.18\textwidth}
        \includegraphics[width=\textwidth]{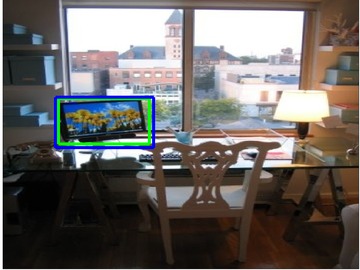}
        \end{subfigure} 
        \hspace{0.05cm} 
        \begin{subfigure}[b]{0.18\textwidth}
        \includegraphics[width=\textwidth]{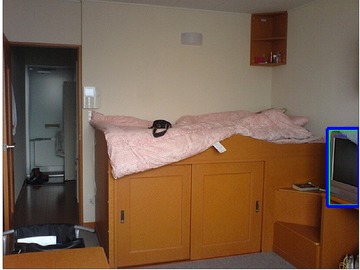}
        \end{subfigure} 
        \end{center}
        \caption{Tvmonitor detections.}
        \label{fig:tvmonitor}
\end{subfigure} 
\vspace{5pt}
\caption{We use blue bounding boxes for the true positive detections and red bounding boxes (if any) for the false positive detections. The ground truth bounding boxes are drawn with green color.}
        \label{fig:Det3}
\end{figure*}

In figures~\ref{fig:Det1} -~\ref{fig:Det3} 
we present some object detections obtained by our approach. 
We use blue bounding boxes to mark the true positive detections and red bounding boxes to mark the false positive detections.
The ground truth bounding boxes are marked with green color.

\begin{figure*}[t!]
\center
\renewcommand{\figurename}{Figure}
\renewcommand{\captionlabelfont}{\bf}
\renewcommand{\captionfont}{\small} 
        \begin{center}
        \begin{subfigure}[b]{0.24\textwidth}
        \includegraphics[width=\textwidth]{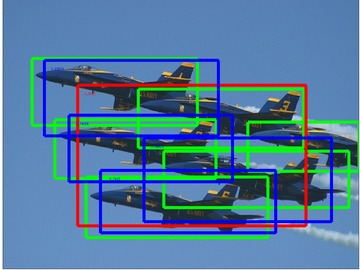}
        \includegraphics[width=\textwidth]{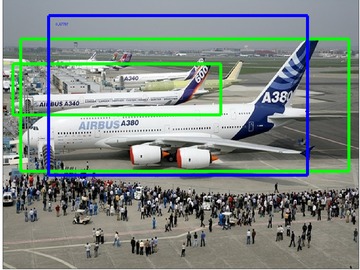}        
        \caption{\footnotesize{Aeroplane}}
        \label{fig:AdjacentAeroplane}
        \end{subfigure} 
        \hspace{0.05cm} 
        \begin{subfigure}[b]{0.24\textwidth}
        \includegraphics[width=\textwidth]{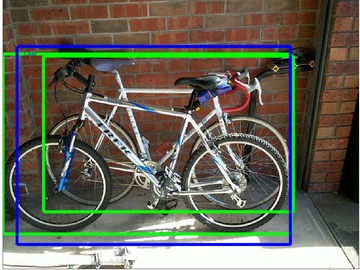}
        \includegraphics[width=\textwidth]{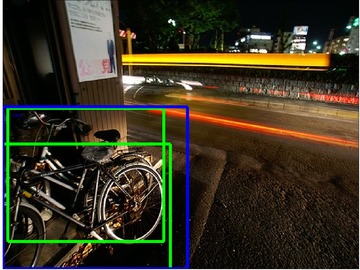}
        \caption{\footnotesize{Bicycle}}
        \label{fig:AdjacentBicycle}
        \end{subfigure} 
        \hspace{0.05cm} 
        \begin{subfigure}[b]{0.24\textwidth}
        \includegraphics[width=\textwidth]{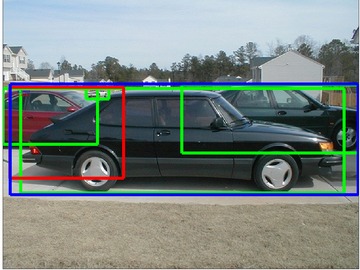}
        \includegraphics[width=\textwidth]{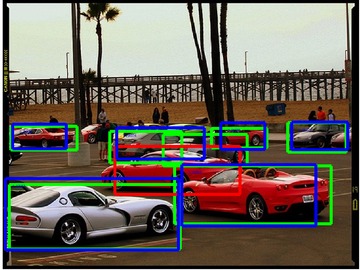}
        \caption{\footnotesize{Car}}
        \label{fig:AdjacentCar}
        \end{subfigure} 
        \hspace{0.05cm} 
        \begin{subfigure}[b]{0.24\textwidth}
        \includegraphics[width=\textwidth]{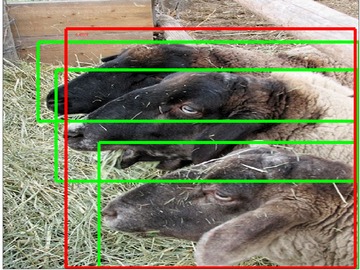}
        \includegraphics[width=\textwidth]{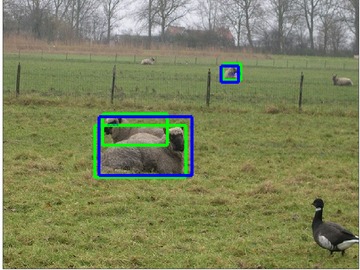}
        \caption{\footnotesize{Sheep}}
        \label{fig:AdjacentSheep}
        \end{subfigure}        
        \end{center}
        \vspace{10pt}
        \caption{Examples of multiple adjacent object instances where our approach fails to detect all of them.
        We use blue bounding boxes to mark the true positive detections and red bounding boxes to mark the false positive detections.
The ground truth bounding boxes are drawn with green color.}
        \label{fig:adjacent}
\end{figure*}

\begin{figure*}[t!]
\center
\renewcommand{\figurename}{Figure}
\renewcommand{\captionlabelfont}{\bf}
\renewcommand{\captionfont}{\small} 
        \begin{center}
        \begin{subfigure}[b]{0.24\textwidth}
        \includegraphics[width=\textwidth]{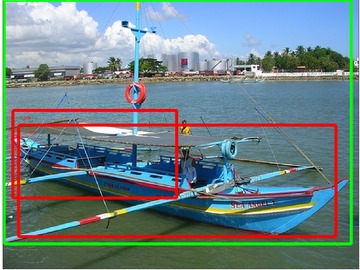}     
        \end{subfigure} 
        \hspace{0.05cm} 
        \begin{subfigure}[b]{0.24\textwidth}
        \includegraphics[width=\textwidth]{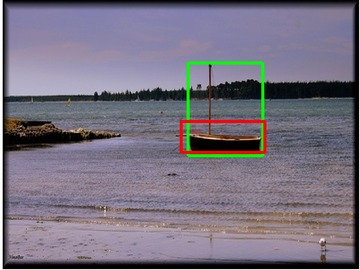}
        \end{subfigure} 
        \hspace{0.05cm} 
        \begin{subfigure}[b]{0.24\textwidth}
        \includegraphics[width=\textwidth]{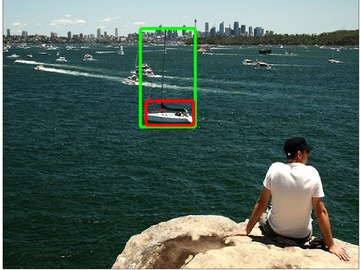}
        \end{subfigure} 
        \hspace{0.05cm} 
        \begin{subfigure}[b]{0.24\textwidth}
        \includegraphics[width=\textwidth]{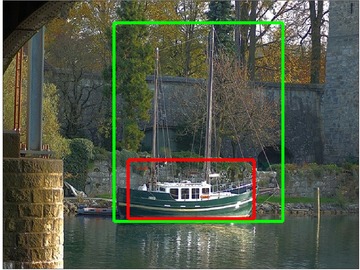}
        \end{subfigure}        
        \end{center}
        \vspace{10pt}
        \caption{Examples of false positive detections for the class boat due to the fact that the detected bounding boxes do not include inside their borders the mast of the boat (it is worth noting that on same cases also the annotation provided from PASCAL neglects to include them on its ground truth bounding boxes). The false positive bounding boxes are drawn with red color and the ground truth bounding boxes are drawn with green color.}
        \label{fig:extended_parts}
\end{figure*}

\emph{\textbf{Failure cases.}}
Accurately detecting multiple adjacent object instances remains in many cases a difficult problem even for our approach. 
In figure~\ref{fig:adjacent} we present a few difficult examples of this type.
In figure~\ref{fig:extended_parts} we show some other failure cases. 

\begin{figure*}[t!]
\center
\renewcommand{\figurename}{Figure}
\renewcommand{\captionlabelfont}{\bf}
\renewcommand{\captionfont}{\small} 
\captionsetup{labelsep=space}
        \begin{center}
        \begin{subfigure}[b]{0.24\textwidth}
        \includegraphics[width=\textwidth]{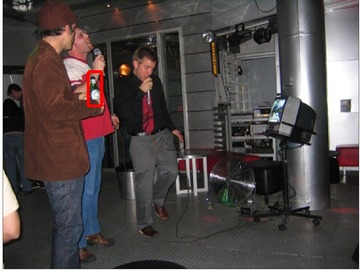}
        \caption{\footnotesize{Bottle}}
        \label{fig:MissedBottle}
        \end{subfigure} 
        \hspace{0.05cm} 
        \begin{subfigure}[b]{0.24\textwidth}
        \includegraphics[width=\textwidth]{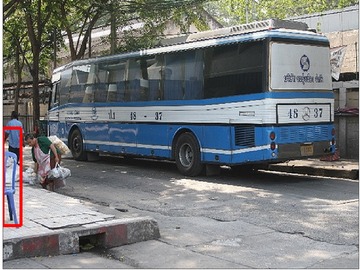}
        \caption{\footnotesize{Chair}}
        \label{fig:MissedChair}
        \end{subfigure} 
        \hspace{0.05cm} 
        \begin{subfigure}[b]{0.24\textwidth}
        \includegraphics[width=\textwidth]{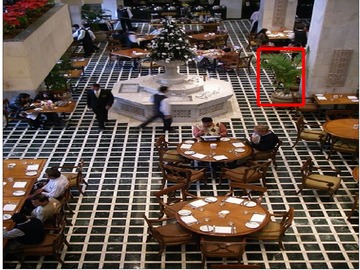}
        \caption{\footnotesize{Pottedplant}}
        \label{fig:MissedPottedPlant}
        \end{subfigure} 
        \hspace{0.05cm}         
        \begin{subfigure}[b]{0.24\textwidth}
        \includegraphics[width=\textwidth]{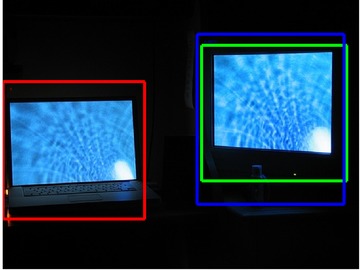}
        \caption{\footnotesize{Tvmonitor}}
        \label{fig:MissedTvmonitor}
        \end{subfigure}        
        \end{center}
        \vspace{10pt}
        \caption{\textbf{-- Missing Annotations: }Examples where our proposed detection system have truly detected an object instance, but because of missed annotations it is considered false positive. For those detections we used red bounding boxes. For any true positive detection on those images we use blue bounding boxes and the corresponding ground truth bounding boxes are drawn with green color.}
        \label{fig:missed_annotation}
\end{figure*}

\emph{\textbf{Missing annotations.}}
There were also cases of object instances that were correctly detected by our approach but which were not in the ground truth annotation of PASCAL VOC2007. 
Figure~\ref{fig:missed_annotation} presents a few such examples of non-annotated object instances. 

\section{Conclusions} \label{sec:conclusions}
We proposed a powerful CNN-based representation for object detection that relies on two key factors: (i)  diversification of the discriminative
appearance factors captured by it through  steering its focus on different regions of the object, and (ii) the encoding of semantic segmentation-aware features. By using it in the context of a CNN-based localization refinement scheme, we show that it
achieves excellent results that surpass the state-of-the art by a significant margin. 

{\small
\bibliographystyle{ieee}
\bibliography{my_paper_bib}

\begin{thebibliography}{10}\itemsep=-1pt

\bibitem{bengio2007greedy}
Y.~Bengio, P.~Lamblin, D.~Popovici, H.~Larochelle, et~al.
\newblock Greedy layer-wise training of deep networks.
\newblock {\em Advances in neural information processing systems}, 19:153,
  2007.

\bibitem{dai2014convolutional}
J.~Dai, K.~He, and J.~Sun.
\newblock Convolutional feature masking for joint object and stuff
  segmentation.
\newblock {\em arXiv preprint arXiv:1412.1283}, 2014.

\bibitem{dalal2005histograms}
N.~Dalal and B.~Triggs.
\newblock Histograms of oriented gradients for human detection.
\newblock In {\em Computer Vision and Pattern Recognition, 2005. CVPR 2005.
  IEEE Computer Society Conference on}, volume~1, pages 886--893. IEEE, 2005.

\bibitem{deng2009imagenet}
J.~Deng, W.~Dong, R.~Socher, L.-J. Li, K.~Li, and L.~Fei-Fei.
\newblock Imagenet: A large-scale hierarchical image database.
\newblock In {\em Computer Vision and Pattern Recognition, 2009. CVPR 2009.
  IEEE Conference on}, pages 248--255. IEEE, 2009.

\bibitem{dong2014towards}
J.~Dong, Q.~Chen, S.~Yan, and A.~Yuille.
\newblock Towards unified object detection and semantic segmentation.
\newblock In {\em Computer Vision--ECCV 2014}, pages 299--314. Springer, 2014.

\bibitem{everingham2008pascal}
M.~Everingham, L.~Van~Gool, C.~Williams, J.~Winn, and A.~Zisserman.
\newblock The pascal visual object classes challenge 2007 (voc 2007) results
  (2007), 2008.

\bibitem{everingham2012pascal}
M.~Everingham, L.~Van~Gool, C.~Williams, J.~Winn, and A.~Zisserman.
\newblock The pascal visual object classes challenge 2012, 2012.

\bibitem{felzenszwalb2010object}
P.~F. Felzenszwalb, R.~B. Girshick, D.~McAllester, and D.~Ramanan.
\newblock Object detection with discriminatively trained part-based models.
\newblock {\em Pattern Analysis and Machine Intelligence, IEEE Transactions
  on}, 32(9):1627--1645, 2010.

\bibitem{girshick2015fast}
R.~Girshick.
\newblock Fast r-cnn.
\newblock {\em arXiv preprint arXiv:1504.08083}, 2015.

\bibitem{girshick2014rich}
R.~Girshick, J.~Donahue, T.~Darrell, and J.~Malik.
\newblock Rich feature hierarchies for accurate object detection and semantic
  segmentation.
\newblock In {\em Computer Vision and Pattern Recognition (CVPR), 2014 IEEE
  Conference on}, pages 580--587. IEEE, 2014.

\bibitem{GuoG15a}
J.~Guo and S.~Gould.
\newblock Deep {CNN} ensemble with data augmentation for object detection.
\newblock {\em CoRR}, abs/1506.07224, 2015.

\bibitem{hariharan2014simultaneous}
B.~Hariharan, P.~Arbel{\'a}ez, R.~Girshick, and J.~Malik.
\newblock Simultaneous detection and segmentation.
\newblock In {\em Computer Vision--ECCV 2014}, pages 297--312. Springer, 2014.

\bibitem{he2014spatial}
K.~He, X.~Zhang, S.~Ren, and J.~Sun.
\newblock Spatial pyramid pooling in deep convolutional networks for visual
  recognition.
\newblock {\em arXiv preprint arXiv:1406.4729}, 2014.

\bibitem{he2015delving}
K.~He, X.~Zhang, S.~Ren, and J.~Sun.
\newblock Delving deep into rectifiers: Surpassing human-level performance on
  imagenet classification.
\newblock {\em arXiv preprint arXiv:1502.01852}, 2015.

\bibitem{hinton2006reducing}
G.~E. Hinton and R.~R. Salakhutdinov.
\newblock Reducing the dimensionality of data with neural networks.
\newblock {\em Science}, 313(5786):504--507, 2006.

\bibitem{hoiem2012diagnosing}
D.~Hoiem, Y.~Chodpathumwan, and Q.~Dai.
\newblock Diagnosing error in object detectors.
\newblock In {\em Computer Vision--ECCV 2012}, pages 340--353. Springer, 2012.

\bibitem{ioffe2015batch}
S.~Ioffe and C.~Szegedy.
\newblock Batch normalization: Accelerating deep network training by reducing
  internal covariate shift.
\newblock {\em arXiv preprint arXiv:1502.03167}, 2015.

\bibitem{krizhevsky2012imagenet}
A.~Krizhevsky, I.~Sutskever, and G.~E. Hinton.
\newblock Imagenet classification with deep convolutional neural networks.
\newblock In {\em Advances in neural information processing systems}, pages
  1097--1105, 2012.

\bibitem{lecun1989backpropagation}
Y.~LeCun, B.~Boser, J.~S. Denker, D.~Henderson, R.~E. Howard, W.~Hubbard, and
  L.~D. Jackel.
\newblock Backpropagation applied to handwritten zip code recognition.
\newblock {\em Neural computation}, 1(4):541--551, 1989.

\bibitem{leordeanu2014features}
M.~Leordeanu, A.~Radu, and R.~Sukthankar.
\newblock Features in concert: Discriminative feature selection meets
  unsupervised clustering.
\newblock {\em arXiv preprint arXiv:1411.7714}, 2014.

\bibitem{LinCY2013NIN}
M.~Lin, Q.~Chen, and S.~Yan.
\newblock Network in network.
\newblock {\em CoRR}, abs/1312.4400, 2013.

\bibitem{lin2014microsoft}
T.-Y. Lin, M.~Maire, S.~Belongie, J.~Hays, P.~Perona, D.~Ramanan,
  P.~Doll{\'a}r, and C.~L. Zitnick.
\newblock Microsoft coco: Common objects in context.
\newblock In {\em Computer Vision--ECCV 2014}, pages 740--755. Springer, 2014.

\bibitem{long2014fully}
J.~Long, E.~Shelhamer, and T.~Darrell.
\newblock Fully convolutional networks for semantic segmentation.
\newblock {\em arXiv preprint arXiv:1411.4038}, 2014.

\bibitem{lowe2004distinctive}
D.~G. Lowe.
\newblock Distinctive image features from scale-invariant keypoints.
\newblock {\em International journal of computer vision}, 60(2):91--110, 2004.

\bibitem{mottaghi2014role}
R.~Mottaghi, X.~Chen, X.~Liu, N.-G. Cho, S.-W. Lee, S.~Fidler, R.~Urtasun, and
  A.~Yuille.
\newblock The role of context for object detection and semantic segmentation in
  the wild.
\newblock In {\em Computer Vision and Pattern Recognition (CVPR), 2014 IEEE
  Conference on}, pages 891--898. IEEE, 2014.

\bibitem{ouyang2014deepid}
W.~Ouyang, P.~Luo, X.~Zeng, S.~Qiu, Y.~Tian, H.~Li, S.~Yang, Z.~Wang, Y.~Xiong,
  C.~Qian, et~al.
\newblock Deepid-net: multi-stage and deformable deep convolutional neural
  networks for object detection.
\newblock {\em arXiv preprint arXiv:1409.3505}, 2014.

\bibitem{redmon2015yolo}
J.~Redmon, S.~Divvala, R.~Girshick, and A.~Farhadi.
\newblock You only look once: Unified, real-time object detection.
\newblock {\em arXiv preprint arXiv:1506.02640}, 2015.

\bibitem{shaoqing2015faster}
S.~Ren, K.~He, R.~Girshick, and J.~Sun.
\newblock Faster r-cnn: Towards real-time object detection with region proposal
  networks.
\newblock {\em arXiv preprint arXiv:1506.01497}, 2015.

\bibitem{ren2015object}
S.~Ren, K.~He, R.~Girshick, X.~Zhang, and J.~Sun.
\newblock Object detection networks on convolutional feature maps.
\newblock {\em arXiv preprint arXiv:1504.06066}, 2015.

\bibitem{sermanet2013overfeat}
P.~Sermanet, D.~Eigen, X.~Zhang, M.~Mathieu, R.~Fergus, and Y.~LeCun.
\newblock Overfeat: Integrated recognition, localization and detection using
  convolutional networks.
\newblock {\em arXiv preprint arXiv:1312.6229}, 2013.

\bibitem{simonyan2014very}
K.~Simonyan and A.~Zisserman.
\newblock Very deep convolutional networks for large-scale image recognition.
\newblock {\em arXiv preprint arXiv:1409.1556}, 2014.

\bibitem{szegedy2014going}
C.~Szegedy, W.~Liu, Y.~Jia, P.~Sermanet, S.~Reed, D.~Anguelov, D.~Erhan,
  V.~Vanhoucke, and A.~Rabinovich.
\newblock Going deeper with convolutions.
\newblock {\em arXiv preprint arXiv:1409.4842}, 2014.

\bibitem{szegedy2014scalable}
C.~Szegedy, S.~Reed, D.~Erhan, and D.~Anguelov.
\newblock Scalable, high-quality object detection.
\newblock {\em arXiv preprint arXiv:1412.1441}, 2014.

\bibitem{van2011segmentation}
K.~E. Van~de Sande, J.~R. Uijlings, T.~Gevers, and A.~W. Smeulders.
\newblock Segmentation as selective search for object recognition.
\newblock In {\em Computer Vision (ICCV), 2011 IEEE International Conference
  on}, pages 1879--1886. IEEE, 2011.

\bibitem{vedaldi2009multiple}
A.~Vedaldi, V.~Gulshan, M.~Varma, and A.~Zisserman.
\newblock Multiple kernels for object detection.
\newblock In {\em Computer Vision, 2009 IEEE 12th International Conference on},
  pages 606--613. IEEE, 2009.

\bibitem{yuting2015improving}
Z.~Yuting, S.~Kihyuk, V.~Ruben, P.~Gang, and H.~Lee.
\newblock Improving object detection with deep convolutional networks via
  bayesian optimization and structured prediction.
\newblock {\em arXiv preprint arXiv:1504.03293}, 2015.

\bibitem{zeiler2014visualizing}
M.~D. Zeiler and R.~Fergus.
\newblock Visualizing and understanding convolutional networks.
\newblock In {\em Computer Vision--ECCV 2014}, pages 818--833. Springer, 2014.

\bibitem{zhu2015segdeepm}
Y.~Zhu, R.~Urtasun, R.~Salakhutdinov, and S.~Fidler.
\newblock segdeepm: Exploiting segmentation and context in deep neural networks
  for object detection.
\newblock {\em arXiv preprint arXiv:1502.04275}, 2015.

\bibitem{zitnick2014edge}
C.~L. Zitnick and P.~Doll{\'a}r.
\newblock Edge boxes: Locating object proposals from edges.
\newblock In {\em Computer Vision--ECCV 2014}, pages 391--405. Springer, 2014.

\end{thebibliography}
}

\newpage


\clearpage

\clearpage
\end{document}